\documentclass[sn-apa,iicol]{sn-jnl}

\jyear{2021}%

\theoremstyle{thmstyleone}%
\theoremstyle{thmstyletwo}%

\theoremstyle{thmstylethree}%

\usepackage{times}
\usepackage{amssymb}
\usepackage{tabu}
\newcolumntype{C}{>{\centering\arraybackslash}X}
\usepackage{subcaption}
\DeclareCaptionType{equ}[][]
\graphicspath{{../}}
\usepackage{doi}
\usepackage{mdframed}

\usepackage{makecell}
\usepackage{soul}
\newcommand{\xh}[1]{#1}

\DeclareGraphicsExtensions{.pdf,.jpeg,.png,.jpg}

\begin{document}

\title[Article Title]{\textcolor{red}{Image and Object Geo-localization}}


\author*[1]{\fnm{Daniel} \sur{Wilson}}\email{daniel.wilson@uvm.edu}

\author[1]{\fnm{Xiaohan} \sur{Zhang}}\email{xiaohan.zhang@uvm.edu}

\author[2]{\fnm{Waqas} \sur{Sultani}}\email{waqas5163@gmail.com}

\author*[1]{\fnm{Safwan} \sur{Wshah}}\email{safwan.wshah@uvm.edu}

\affil*[1]{\orgdiv{Department of Computer Science}, \orgname{University of Vermont}, \city{Burlington}, \country{USA}}

\affil[2]{\orgdiv{Department of Computer Science}, \orgname{Information Technology University},   \city{Lahore},  \country{Pakistan}}


\abstract{The concept of geo-localization broadly refers to the process of determining an entity's geographical location, typically \textcolor{red}{in the form of} Global Positioning System (GPS) coordinates. The entity of interest may be an image, a sequence of images, a video, a satellite image, or even objects visible within the image. Recently, massive datasets of GPS-tagged media have become available due to smartphones and the internet, and deep learning has risen to prominence and enhanced the performance capabilities of machine learning models. These developments have enabled the rise of \textcolor{red}{image and object geo-localization}, which has impacted a wide range of applications such as augmented reality, robotics, self-driving vehicles, road maintenance, and 3D reconstruction. This paper provides a comprehensive survey of \textcolor{red}{visual geo-localization}, which may involve either determining the location at which an image has been captured (image geo-localization) or geolocating objects within an image (object geo-localization). We will provide an in-depth study \textcolor{red}{of visual geo-localization} including a summary of popular algorithms, a description of proposed datasets, and an analysis of performance results to illustrate the current state of the field.}




\maketitle

\footnote{We have obtained the copyright for all figures used in this paper by purchasing all the applicable rights from their publishers.}

\section{Introduction}\label{intro}

\begin{figure*}[t]
    \centering
    \begin{subfigure}[t]{0.5\textwidth}
        \centering
        \includegraphics[height=1.2in]{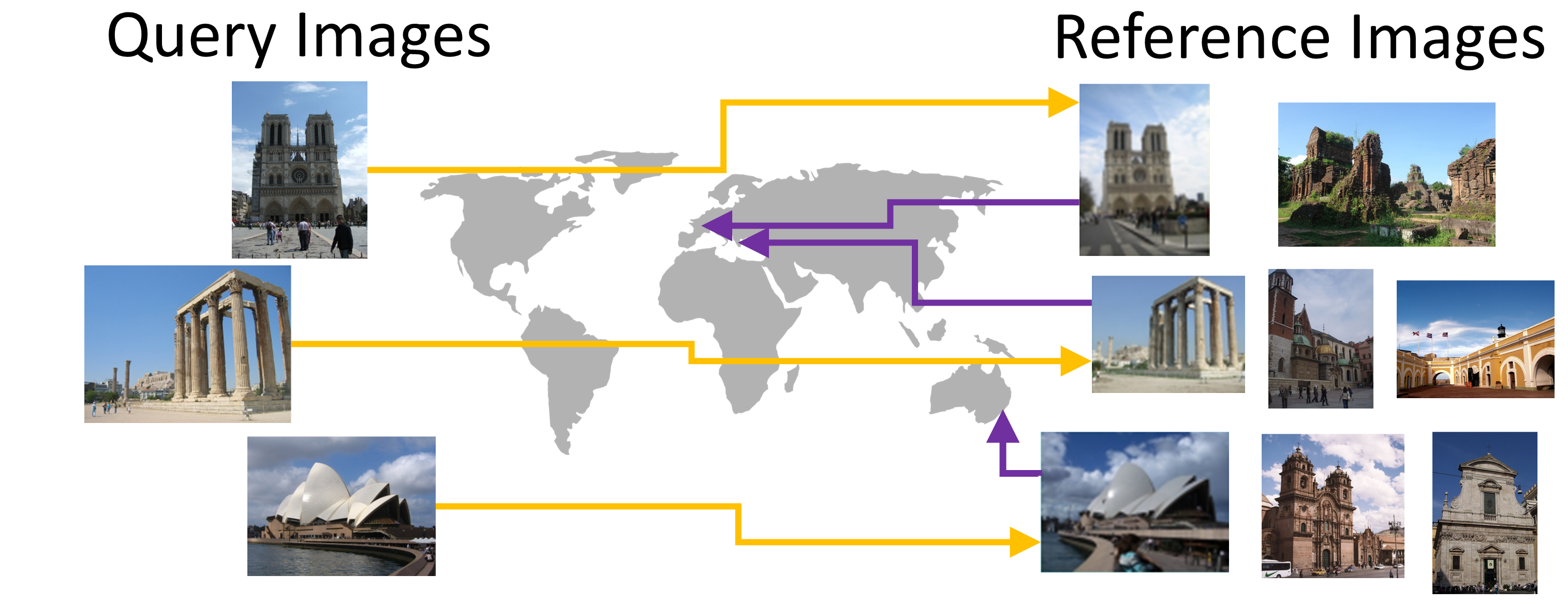}
        \caption{The objective of a single-view geo-localization algorithm is to determine the GPS coordinates of images without additional views. Most commonly, query images (left) are compared against reference images (right) to find a similar image with a known geo-location.}
        \label{fig:single_overview}
    \end{subfigure}%
    ~
    \begin{subfigure}[t]{0.5\textwidth}
        \centering
        \includegraphics[height=1.2in]{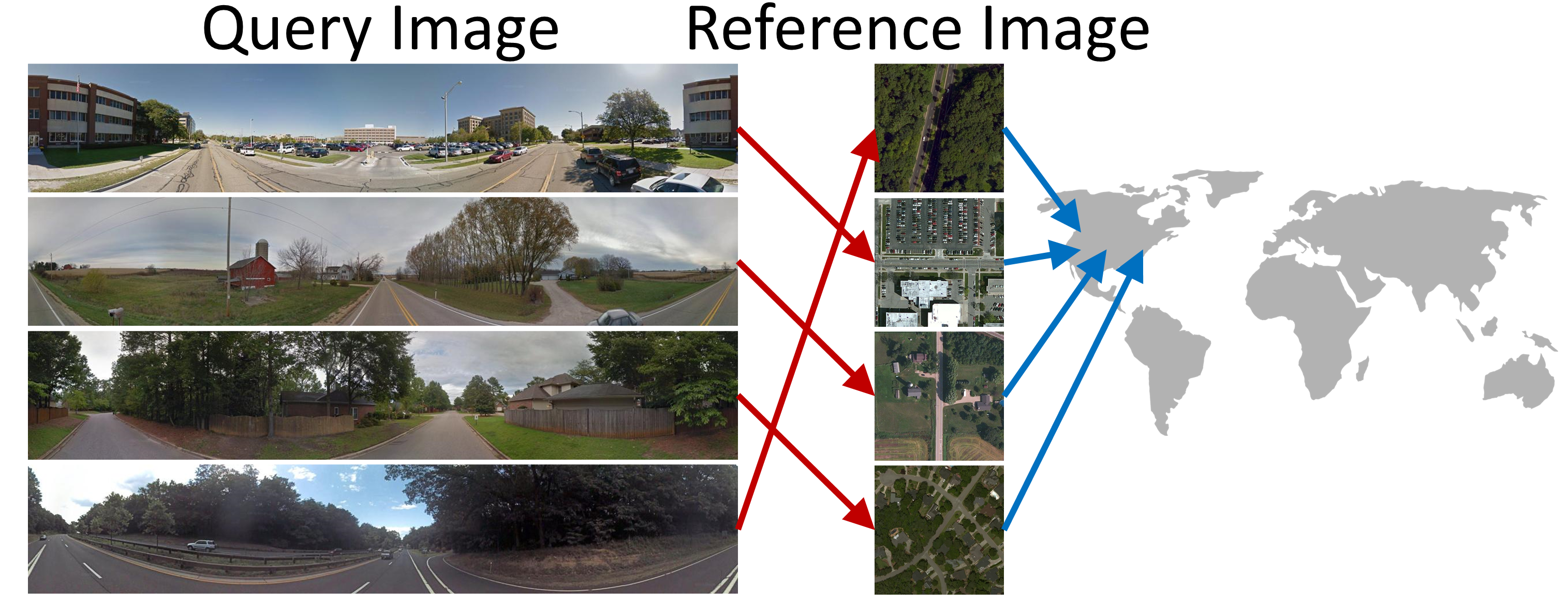}
        \caption{Cross-view geo-localization approaches take advantage of both ground and satellite views. Ground-view query images (left) are matched to geo-tagged reference satellite images (center). The GPS locations of the reference images are used as the geo-spatial prediction (right) for the query images.}
        \label{fig:cross_overview}
    \end{subfigure}
    \caption{Single view (left) and cross-view geo-localization (right) are the two core approaches to image geo-localization.}
\end{figure*}

\begin{figure*}[t]
    \centering
    \includegraphics[width=\textwidth]{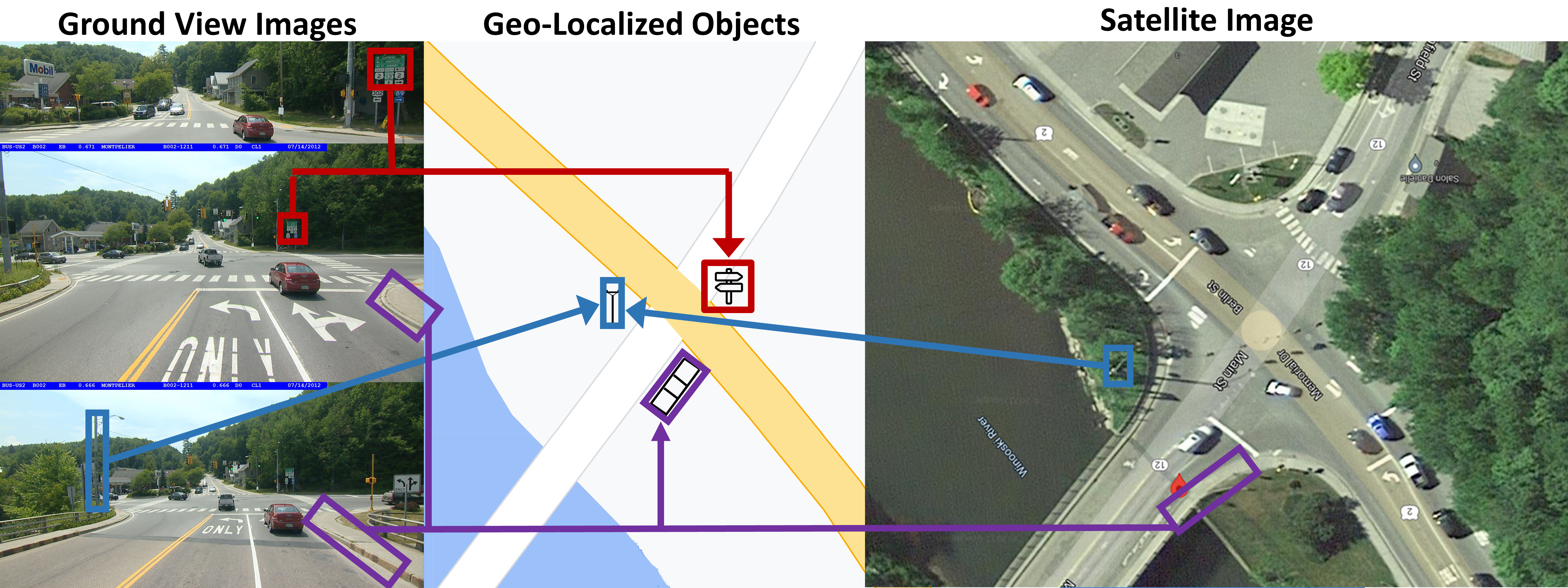}
    \caption{A high level overview of the basic tasks performed by object geo-localization algorithms. Objects are geo-localized either from ground-view images, satellite images, or both. Note that one of the challenges associated with this task is that not all objects may be visible in both ground or satellite images.}
    \label{fig:object_overview}
\end{figure*}

Location is an important piece of context information that humans subconsciously take into account when interpreting the meaning behind a scene. For example, determining that an image is located in a popular tourist city may reveal that the image contains popular attractions, whereas a video geo-localized in a remote region may be useful to people performing land surveying. Recently, \textcolor{red}{visual geo-localization has emerged as a field in computer vision due to the richness of technical challenges it poses and its} wide range of practical applications such as augmented reality~\citep{6-DOF}, robotics~\citep{shady_dealings,robust}, self-driving vehicles~\citep{robust,uber}, road maintenance~\citep{uber} and 3D reconstruction~\citep{building_rome}.

% multiple fields have attempted to interpret meaning of the concept of location within images
% predict positions realtive to other images
% predict or classify scenes
% find simlar images
% map local environment
% directory regress GPS coordinates which is main focus of this paper
% varying camera perspectives ground-view aerial-view
% varying environments city, rural, landmarks

Multiple \textcolor{red}{related fields have emerged} for extracting location information from pixel data within images. Visual geo-localization approaches, \textcolor{red}{as discussed in this paper,} specify the set of global positioning system (GPS) coordinates to which an image or objects within an image is predicted to belong. Other proposed methods involve predicting the scene depicted in an image, constructing a map of the surrounding environment, or matching the image to other similar images from a database. The source images could be taken from a variety of camera perspectives, including ground, aerial, or hybrid perspectives. Furthermore, images may depict varying environments such as city or rural settings. A detailed description of the structure of the field of visual geo-localization and its sub-fields is provided in Section~\ref{sec:related_fields}.

In the field of visual geo-localization, few survey papers have been published. In~\citep{state_art} a comprehensive study was conducted including images captured from large cities and natural environments. The authors of the paper focused mainly on non-deep learning methods as few deep learning methods had been applied to this domain at the time this paper was published. In~\citep{localization_reconstruction} the authors conducted a comprehensive review of the state-of-the-art methods but did not provide extensive details on the existing datasets, evaluation metrics, and generative-based methods. Their survey also did not cover the emerging cross-view image geo-localization sub-field.

Our survey is the first to fully cover the breadth of the field of \emph{visual geo-localization}, including both object and image-based approaches using ground-view, satellite-view, and cross-view images. We also place a greater emphasis on cross-view and deep-learning approaches, due to their rise to prominence and the expectation that they will play a central role in future developments. We will present a comprehensive study of each of these sub-fields by addressing existing methods, current public datasets, and evaluation metrics. We compare each method's pros and cons and discuss their reported performance on recent public datasets. We summarize popular datasets and benchmark results in tables for accessibility and ease of readability.

The remainder of this paper is structured as follows. We will provide a high-level overview of fields related to visual geo-localization in~\ref{sec:related_fields}. Next we will survey the three major visual geo-localization sub-categories including single-view geo-localization, cross-view geo-localization, and object geo-localization in Sections~\ref{sec:single-view},~\ref{sec:cross-view}, and~\ref{sec:stationary}. A visual overview of each each respective geo-localization category is provided in Figure~\ref{fig:single_overview}, Figure~\ref{fig:cross_overview}, and Figure~\ref{fig:object_overview}. For each category, we will introduce the theory behind the technique, survey proposed approaches, discuss popular datasets, and describe evaluation metrics.

\section{Related Fields}\label{sec:related_fields}
\begin{figure*}[t]
% \begin{mdframed}[backgroundcolor=red!50,linecolor=red!50]
    \centering
    \includegraphics[width=\textwidth]{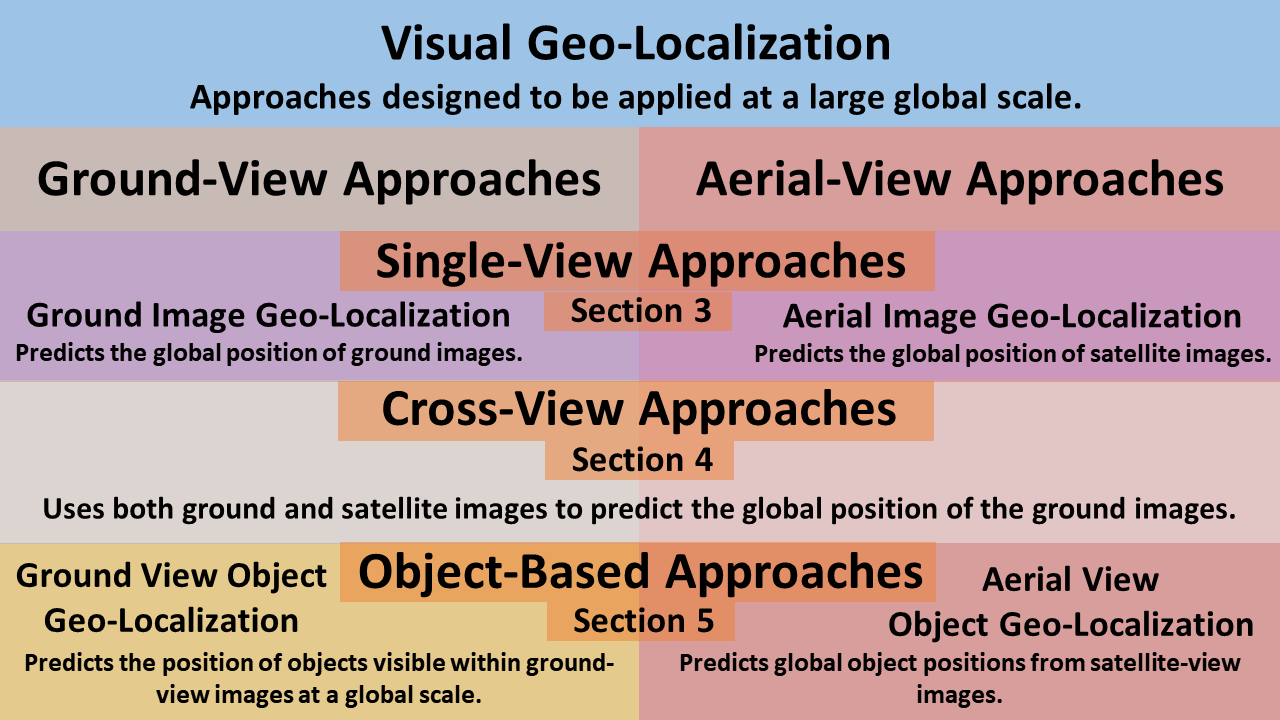}
    \caption{\textcolor{red}{Visual geo-localization, as covered in this survey, can be sub-divided based on whether the viewpoint of their input(s) are ground-view, satellite-view, or cross-view. Both ground and satellite view geo-localization may either geo-localize the position of the images, which is referred to as image geo-localization, or objects within the image, referred to as object geo-localization.}}
    \label{fig:field_overview}
% \end{mdframed}
\end{figure*}

Computer vision-based localization algorithms have branched out into multiple intertwined sub-fields to address the unique challenges associated with different applications (see Figure~\ref{fig:field_overview}). First, these vision-based algorithms can be broadly characterized as either~\emph{visual geo-localization}~\citep{state_art, localization_reconstruction} or~\emph{visual localization}~\citep{slam_survey, visual_based_survey} approaches. These categories are distinguished based on their scale and coordinate system. \emph{Visual localization} algorithms predict the positions of images or objects within a known spatial representation typically covering a restricted area, and are normally applied on a smaller spatial scale~\citep{deep_visual_survey, visual_based_survey}. Specifically, these approaches typically select a nearby object or image to use as a reference point, and predict object offsets relative to it~\citep{visual_based_survey}, as opposed to predicting the global position coordinates of an object. \emph{Visual geo-localization} is concerned with determining the geospatial position of objects or images at a global scale~\cite{state_art}. Input images could originate from anywhere in the world, and thus the goal of the algorithm is to predict each image's or object's GPS coordinates. These approaches are broader in their scope due to not being confined to a local area. The trade-off is that \textcolor{red}{the accuracy of these approaches is reduced due to their larger scale and lack of a nearby reference point to compare features against..}
% The largest distinction between visual localization research problems is whether the goal is to localize images or objects.
% The major difference is that object localization approaches attempt to predict the relative location of objects within an image, whereas image localization approaches predict the relative position of the image itself.

Visual localization algorithms can be further divided into sub-fields. Since this paper focuses on \textcolor{red}{visual} geo-localization, we will briefly note different \textcolor{red}{visual} localization sub-fields and direct the reader to applicable survey papers for additional information. The core goal shared by all visual localization methods is to predict the relative position of objects or images within a relatively small-scale, localized environment. We direct the reader to ~\citep{deep_visual_survey} and ~\citep{visual_based_survey} for surveys broadly focusing on image localization. One key visual localization sub-field is camera pose estimation, in which the objective is to predict the position and orientation of the camera relative to a scene or object depicted in an image. Typically, the scene or object is considered to be fixed and the camera pose variable, and thus the camera pose must be determined from each image relative to the depicted scene. An important distinction is that camera pose differs from camera position, since pose additionally includes camera x, y, and z axis rotation, whereas position exclusively refers to where a camera is located in a local coordinate system. Camera pose approaches predict the full camera pose, whereas purely image localization algorithms only predict camera position. We direct the reader to Section 4 of~\citep{visual_based_survey} for additional information about camera pose estimation. Simultaneous localization and mapping (SLAM) algorithms are designed to construct maps of the surrounding environment. Like all visual localization approaches, these maps are local and not global in scale, and since they are typically applied for robot navigation, they are usually designed to run using live sensor inputs in real time.~\citep{slam_survey} provides a survey of SLAM methods. Finally, image retrieval algorithms attempt to find images containing semantically similar objects or scenes. This task originated as a visual localization approach in which a query image must be matched to other nearby images in a local environment. As we will discuss in Section~\ref{sec:reference_based}, image retrieval is now commonly adopted as one of the steps in \textcolor{red}{cross-view image} geo-localization pipelines. We refer the reader to~\citep{instance_retrieval} and~\citep{deep_image_retrieval} for an in-depth survey of this sub-field. Our survey will cover approaches that adapt image retrieval methods for \textcolor{red}{cross-view image} geo-localization.

Our survey focuses on~\emph{visual geo-localization}. We remind the reader that the distinction between localization and geo-localization is that geo-localization focuses on a broader, geographical scale, where dataset images could originate from anywhere across the earth. The ultimate goal of \textcolor{red}{visual} geo-localization approaches is to predict an explicit set of GPS coordinates~\citep{state_art, localization_reconstruction}, unlike localization approaches which predict positions within a local scene or relative coordinate system~\citep{slam_survey, visual_based_survey}. These approaches can be divided based on whether they predict the GPS coordinates of images, referred to as~\emph{image geo-localization}, or the GPS position of objects visible within an image, referred to as ~\emph{object geo-localization}. We can further divide these approaches into sub-fields based on the view perspective of the images provided to the geo-localization algorithm.~\emph{Ground-View object geo-localization} uses images taken from a ground perspective to predict the GPS coordinates and, in some cases, the class of each object visible with an image.~\emph{Satellite-View object geo-localization} performs the same task of predicting each objects' GPS coordinates, however, the input images are from \textcolor{red}{an aerial (typically satellite)} perspective. Our paper focuses on datasets containing ground based images, so we direct the reader to~\cite{satellite_detection} for an in-depth survey specifically focusing on satellite imagery. The other major sub-category of visual geo-localization approaches is \emph{image geo-localization} algorithms, which predict the GPS coordinates of images as opposed to objects. As shown in Figure~\ref{fig:field_overview}, \emph{single-view geo-localization} algorithms predict the geospatial location of each image, and can be further divided into~\emph{ground-view} and\textcolor{red}{~\emph{aerial-view}} approaches depending on whether the input images are taken from a ground or \textcolor{red}{aerial} perspective. An interesting recent development has been the advent of~\emph{cross-view image geo-localization} approaches, which are a hybrid of ground-view and \textcolor{red}{aerial-view} image geo-localization. These methods typically employ a reference database of satellite images to identify features that can be matched to the query ground images to determine the query's geolocation. The reference image database could reach hundreds of thousands or millions of images, yielding a complex matching problem. Since this survey covers the full field of visual geo-localization, we will cover all the approaches introduced in this paragraph.

% Image or video geo-localization, also referred to as image geo-localization, mainly answers the question of where in a region an image or video was captured when camera metadata indicating its location is not available~\citep{large_scale,state_art,localization_reconstruction, lin2013cross}. This task is most commonly defined as a query to reference matching problem~\citep{tag_refinemnt,aerial_database, CVUSA}. Typically, the query is an image or sequence of images with an unknown GPS location, and the references are a large set of images with known GPS locations against which the query is matched. The reference image database could reach hundreds of thousands or millions of images, yielding a complex matching problem.

% \begin{figure*}[t]
%     \centering
%     \includegraphics[width=\textwidth]{intro-fig-0-field_structure.png}
%     \caption{A visual representation of the high-level organizational structure between the fields involving vision-based localization algorithms. These algorithms can be divided into two major sub-categories, visual geo-localization and visual localization. As displayed on the right, visual localization is composed of many smaller fields including image localization, camera pose prediction, SLAM, and image retrieval. Visual geo-localization (left) contains both image-based approaches (top left) and object based approaches (bottom left). Image based geo-localization approaches can be divided even further based on whether they are ground-view, satellite-view, or cross-view (center left).}
%     \label{fig:field_overview}
% \end{figure*}

\section{Single View Image Geo-localization}\label{sec:single-view}

Estimating the location of an image using only visual information is a uniquely challenging task as the images could be taken from anywhere in the world. The scenes depicted in images can show huge variations in time of day, weather, type of visible objects, background, and camera settings. However, images also contain useful \textcolor{red}{context} information which can help in geo-localization. This information includes landmarks, architectural details, building shapes, and surrounding environments. In single-view image geo-localization, the objective is to determine the geographic location of an image from a single camera or view. A database of reference images (also captured from the same view) may be available to assist in the geo-localization process.

This section is structured as follows. First, we introduce the main categories of techniques used to perform single-view geo-localization, followed by a description of key proposed approaches within each of these categories. Second, we describe the popular datasets commonly used for training and testing models. Finally, we discuss popular adopted benchmarks and provide a comparison of results reported by authors.

\begin{table*}[t]
    % \begin{mdframed}[backgroundcolor=red!50,linecolor=red!50]
    \footnotesize
    \centering
    \begin{tabular}{cccc} \toprule \toprule
    Method & Type & Deep Learning vs. Traditional & Year  \\ \midrule
    Im2GPS~\citep{im2gps} & Reference-Based & Traditional & 2008 \\
    Google Maps Localization~\citep{accurate_local} & Reference-Based & Traditional & 2010 \\
    Alps Skyline Segmentation~\citep{large_mountains} & Rural Geo-localization & Traditional & 2012 \\
    Topographic Maps~\citep{topographic_maps} & Rural Geo-localization & Traditional & 2012 \\
    City Identity~\citep{city_identity} & Cell-Based & Traditional & 2014 \\
    Tag Refinement~\citep{tag_refinemnt} & GPS Refinement & Traditional & 2014 \\
    Generalized Clique Graphs~\citep{nearest_neighbor} & Reference-Based & Traditional & 2014 \\
    Skyline and Ridges~\citep{mountain_images} & Rural Geo-localization & Traditional & 2015 \\
    GPS Image Fusion~\citep{fusion} & GPS Refinement & Traditional & 2015 \\
    Im2GPSv2~\citep{large_scale} & Reference-Based & Traditional & 2015 \\
    View Synthesis~\citep{place_recognition} & Reference-Based & Traditional & 2015 \\
    Feature Prediction~\citep{predicting_features} & Reference-Based & Traditional & 2015 \\
    PlaNet~\citep{planet} & Cell-Based & Deep Learning & 2016 \\
    Aerial Intersections~\citep{matching_roads} & Aerial & Traditional & 2016 \\
    Feature Re-Weighting~\citep{feature_reweighting} & Reference-Based & Traditional & 2017 \\
    Deep Learning Era~\citep{deep_era} & Reference-Based & Deep Learning & 2017 \\
    NetVLAD~\citep{netvlad} & Reference-Based & Deep Learning & 2018 \\
    CPlaNet~\citep{cplanet} & Cell-Based & Deep Learning & 2018 \\
    Hierarchical Cells~\citep{hierarchical} & Cell-Based & Deep Learning & 2018 \\
    Trip Reporting~\citep{immersive_trip} & Rural Geo-localization & Traditional & 2018 \\
    Semantic Edges~\citep{bucolic_environment} & Rural Geo-localization & Deep Learning & 2020 \\
    LandscapeAR~\citep{landscape_ar} & Rural Geo-localization & Traditional & 2020 \\
    CrossLocate~\citep{cross_locate} & Rural Geo-localization & Deep Learning & 2022 \\
    \textcolor{red}{Translocator}~\citep{pigeon} & \textcolor{red}{Cell-Based} & \textcolor{red}{Deep Learning} & \textcolor{red}{2023} \\
    \textcolor{red}{PIGEON}~\citep{pigeon} & \textcolor{red}{Cell-Based} & \textcolor{red}{Deep Learning} & \textcolor{red}{2023} \\
    \textcolor{red}{Hierarchies and Scenes}~\citep{hierarchies_and_scenes} & \textcolor{red}{Cell-Based} & \textcolor{red}{Deep Learning} & \textcolor{red}{2023} \\
    \end{tabular}
    \caption{Summary of single-view geo-localization methods. \textcolor{red}{Each method can be characterized as either a reference-based, cell-based, rural, or GPS refinement approach. Methods can also be distinguished based or whether they apply deep learning or rely on traditional computer vision techniques.}}
    \label{tab:summary_single_view_models}
    % \end{mdframed}
\end{table*}

% We note that unlike single object geolocalizaiton, performance benchmarks are reasonably standardized in this field, and results can often be compared between different approaches. Papers using private datasets may still not be comparable, since the level of geographic descriptiveness present in a dataset can dramatically effect the performance of a model. For example, a dataset containing mostly tourist images containing popular landmarks will be far easier to geolocalize than a dataset containing indoor photos lacking geographically descriptive features.
\subsection{Techniques}\label{Techniquessingle view image geo-localization}
Single view image geo-localization approaches can be broadly categorized into four main sub-categories: geographical cell-based approaches, GPS reference-based approaches, GPS refinement approaches, and \textcolor{red}{aerial-view} approaches. \textbf{Geographical cell-based geo-localization} approaches \textcolor{red}{geolocalize} images by dividing the earth into discrete geographical cells and then training a model to predict which cell an image belongs to. Since the cells could be thought of as discrete classes, these approaches are also often denoted as classification approaches.
\textbf{GPS \textcolor{red}{reference-based}} approaches structure geo-localization as an image retrieval problem. Specifically, a large GPS-tagged image database containing images with known GPS locations is constructed. Given a query image, the GPS coordinates of the matched reference images are used to determine the GPS coordinates of the query image. \textbf{GPS refinement} approaches are designed to refine noisy GPS coordinates (commonly found in photo albums) with the goal of increasing their accuracy. Instead of geo-localizing ground-based images, \textcolor{red}{\textbf{aerial-view}} approaches focus on geo-localizing aerial (such as satellite) images. Note that in contrast to cross-view geo-localization which involves jointly reasoning from both ground and aerial images, \textcolor{red}{aerial-view} approaches use aerial images only. Finally, \textbf{rural geo-localization} approaches specifically focus on predicting geographical locations of images depicting rural environments. City images are omitted from their datasets.

\subsubsection{Cell-Based Image Geo-localization}
A conceptually simple method of formulating image geo-localization is to structure it as a classification problem. Geographical cell-based geo-localization techniques divide Earth's geography into cells and label each GPS-tagged image with its geographical cell. A machine learning model is then trained to learn a function that maps each input image to a cell prediction. This approach has the advantage of framing the challenging geo-localization problem as a conceptually simple image classification problem, mitigating the need for an elaborate algorithm that \textcolor{red}{regresses numerical} GPS predictions. Also, since deep learning models output probabilities for each class, a model's output may still be useful to a human even if the correct cell is not assigned the greatest probability. For example, a human attempting to determine where an image was taken would still benefit if a model could narrow down its possible locations to a list of five high-probability geographical cells.

An early approach was proposed by~\citep{city_identity}, who built a classifier capable of predicting the city in which images were captured. Since each city is a restricted geographic location, \textcolor{red}{the cities could be loosely interpreted as representing geographical cells.} The authors employed high-level labels from the SUN database~\citep{SUN} to serve as scene attributes. These attributes include architecture type, water coverage, and green space. After extracting deep features for each attribute, an ensemble of Support Vector Machine (SVM) classifiers were trained to predict the attributes associated with an image and its corresponding city. At the time, this approach provided a novel idea for how to formulate geo-localization as a classification problem. Compared to new approaches, however, its performance \textcolor{red}{is} limited by the capabilities of support vector machines and hand-crafted features, the number of cities available, and the size of the dataset.

~\citet{planet} improved upon~\citep{city_identity}, by incorporating a deep learning model instead of relying on support vector machines. The enhanced capabilities of deep learning models enabled the authors to perform a much \textcolor{red}{larger-scale} experiment in which the surface of the earth was divided into cells. Their model was trained via standard backpropagation to predict which cell each image belonged to. The authors also studied the impact of different sized cell partitions. Since the predicted geolocation is bounded by the size of the cell, smaller sized cells are desirable to produce more accurate GPS predictions. However, reducing the cell size also results in fewer training samples per cell, \textcolor{red}{which harms performance and can cause overfitting.}~\citep{planet} proposed to satisfy both these restrictions by utilizing an adaptive partitioning in which areas more populated with images could be divided into more fine-grained cells, and areas containing fewer images would have larger cells to compensate for the lack of samples.

Later research has continued to improve cell-based geo-localization by applying improved models, developing better cell partitioning techniques, and applying larger datasets.~\citep{cplanet} proposed to partition cells using a combinatorial approach which produces a map containing intersecting fine and coarse-grained partitions. By training separate classifiers on different cell partitionings and then aggregating the results to produce a final prediction, the authors mitigated the limitations associated with the fixed-sized cell partitioning reported in~\citep{planet}.

\textcolor{red}{~\citep{hierarchical} achieved further performance improvements by constructing} a larger dataset and using a more sophisticated deep learning model specifically adapted to the geo-localization task. Specifically, they mined their data from Yahoo Flickr Creative Commons~\citep{yahoo_flickr}. Each image had associated scene information indicating if its environment was indoor, natural, or urban. Their main architectural innovation was the finding that training separate classifiers for different types of scenes improved performance.

\textcolor{red}{Some recent methods have been proposed to take advantage of the performance capabilities of recently popularized transformer architectures.For example, ~\cite{translocator} proposed Translocator, a transformer architecture which simultaneously predicts a course, middle, and fine cell partition to handle the trade-off between samples per cell and cell size.~\cite{pigeon} leveraged transformers using an agnostic geocell division approach which constructs semantically meaningfull cells by applying an OPTICS~\cite{OPTICS} clustering algorithm. They train their transformer model, PIGEON, using a custom loss which penalizes the Haversine distance between the predicted and actual geocell.~\cite{hierarchies_and_scenes} built a transformer architectuer which used a hierarchical cross-attention method to model the relationship between different geographic levels and learn a separate representation for different environmental scenes, similar to~\cite{hierarchical}. While these transformer architectures are flexible computational models with the potential to yield high performance, the trade off is that they require large amounts of data, pre-training, and additional pretext tasks to reach their full potential.~\cite{translocator} attempted to address this weakness by adding a second branch to their model which predicts a segmentation mask, resulting in a more robust feature representation.~\cite{pigeon} proposed a pretraining procedure in which they develop a rule-based system to create captions for their images, and then train an architecture based on the CLIP~\cite{CLIP} model to predict the captions. These pretraining tasks are intended to fine tune transformer architectures to learn domain specific features without overfitting and offset the large amounts of data these architectures require.}

While cell-based approaches offer a simple framework with which to study geo-localization, these approaches suffer from an inherent limitation imposed by static cells, which is that the coordinate prediction is only as specific as the cell size. While some research has proposed improved partitioning methods, there remains an inherent trade-off between the cell size/specificity and the number of training samples. Additionally, since these methods do not employ a reference database, it is unclear if neural networks can memorize enough features to map an image from any location in the entire world to its corresponding cell.

\subsubsection{Reference Based Geo-localization}\label{sec:reference_based}
The most popular approach to image geo-localization is reference-based geo-localization. These techniques localize a query image by retrieving one or more nearby reference images with known GPS coordinates using global image features. These approaches have the advantages of being able to regress a set of GPS coordinates as opposed to restricting predictions to a single cell, and the capability to generate predictions for any location from which reference images are available.

% basic approaches rely on hand crafted features
An early reference base approach was proposed by~\citep{im2gps}. The authors demonstrated a baseline algorithm in which they extracted hand-crafted visual features including textures and lines, gist descriptors, and geometric context from each image. The authors computed the distance in the feature space between each query and feature image. They geo-located each query by calculating its K nearest feature spaced images and performing mean shift clustering on the geolocations of the matching images.~\citep{large_scale} proposed two major improvements compared to~\citep{im2gps}. First, they improved the quality of extracted features by incorporating additional SIFT~\citep{sift} interest points using Hessian-affine and maximally stable extremely region (MSER)~\citep{mser} detectors. Second, the authors proposed an improved approach to image matching. Instead of matching images nearby in the feature space, they used a lazy learning approach inspired by SVM-KNN~\citep{svm_knn}. This new matching algorithm yielded better performance without significantly increasing run time.

% main contribution is tree based approaches
% disadvantage is tree based approaches rely on many nearby images and images can inherit error fro mother images
In practice, a large reference database with densely packed images will contain multiple reference images near each query, which could be further exploited for more accurate GPS predictions. Instead of matching queries to a single reference image, some works have proposed methods to match queries to multiple reference images for greater performance.~\citep{accurate_local} computed SIFT descriptors and arranged them into a tree using fast approximate nearest neighbors (FLANN)~\citep{FLANN}. They proposed a GPS pruning method to remove unreliable features, and they implemented a smoothing step to handle objects which appear in multiple query images, resulting in features that were more descriptive and less redundant compared to~\citep{im2gps} and~\citep{large_scale}. Instead of matching the query to a single reference, the authors proposed a method identifying multiple nearby reference image subsets and predicting the queries coordinates from the highest quality subsets. This \textcolor{red}{produced} more accurate GPS predictions than the previously proposed methods that matched to a single reference.~\citet{nearest_neighbor} improved this approach by incorporating both local features (SIFT) and global features (GIST). Instead of matching to the first nearest neighbors, they selected the matching reference images using the NP-hard Generalized Minimum Clique Graphs (GMCP) problem. The authors benchmarked their dataset on 102,000 images that were mined from Google Street View. While approaches matching to multiple reference images make better use of the entire database by exploiting more than one matching reference image, the tradeoff is that they require a larger reference database, resulting in more computationally expensive approximate matching algorithms.

\begin{figure*}
    \centering
    \includegraphics[width=1\textwidth]{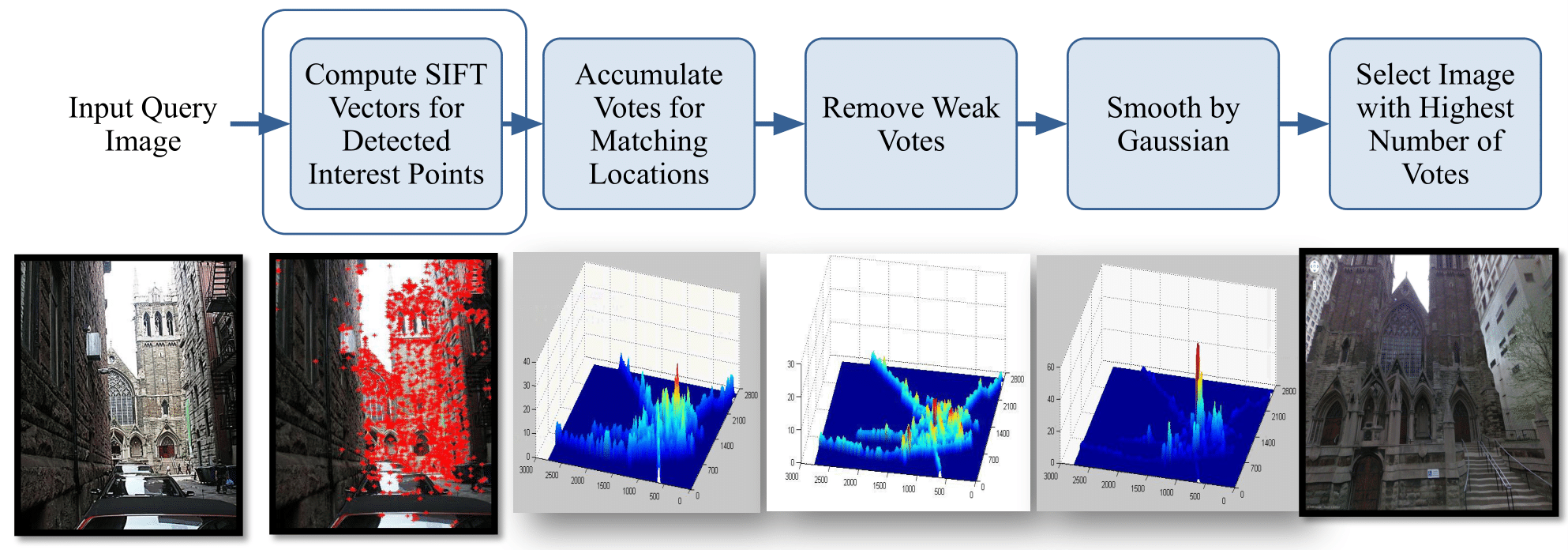}
    \caption{A block diagram of the localization process proposed by~\citep{accurate_local}. SIFT vectors are calculated and used to count votes. After removing weak votes and applying a smoothing function, the reference image with the greatest number of votes is selected to predict the GPS location. Figure is taken from~\citep{accurate_local}.}
    \label{fig:accurate_local}
\end{figure*}

\begin{figure*}
    \centering
    \includegraphics[width=\textwidth]{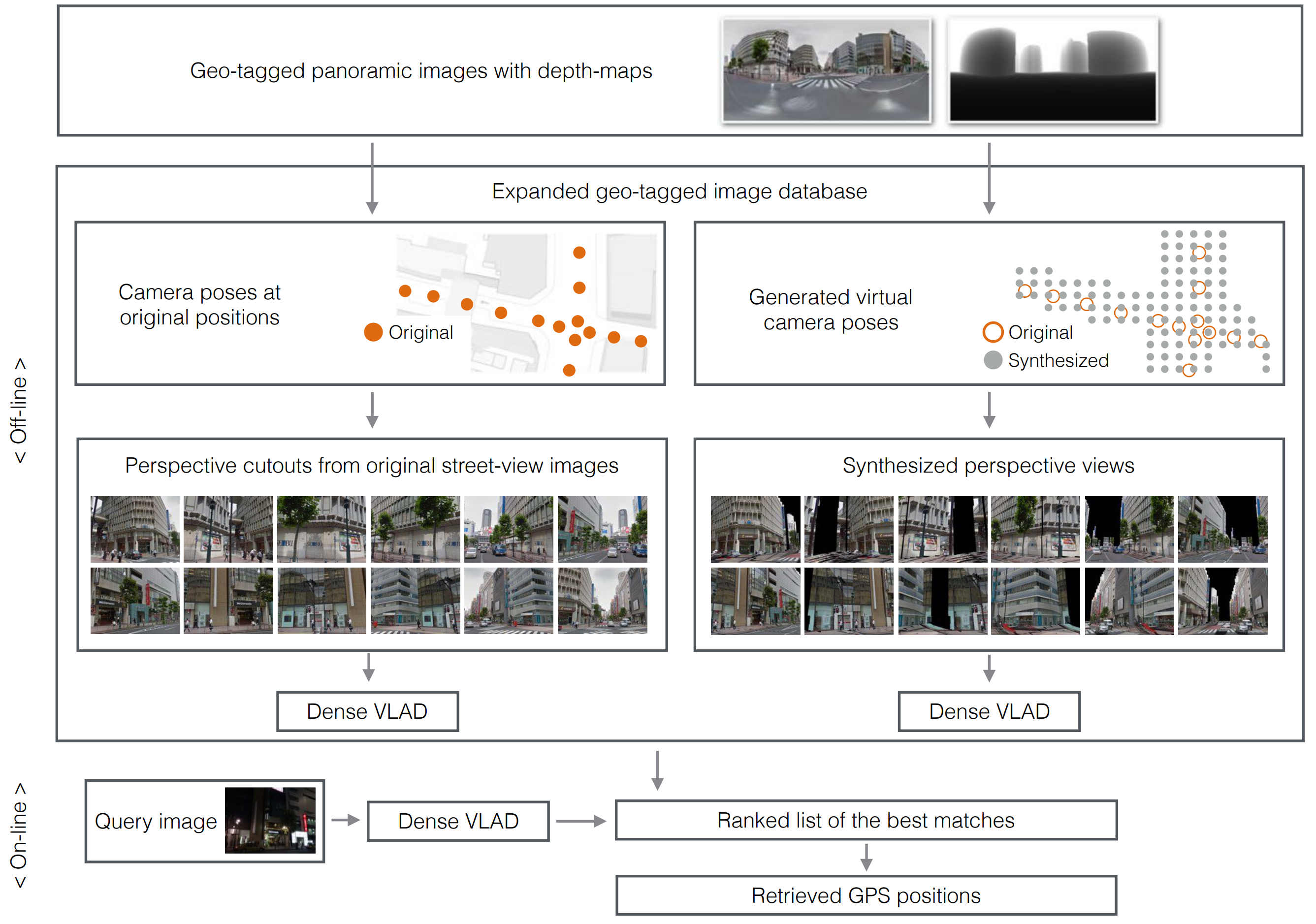}
    \caption{\textcolor{red}{An overview of the approach for place recognition proposed by~\citep{place_recognition}. Their system is composed of two stages, the first of which computes features from both the original and synthesized perspective, and the second matches the query image to a similar image in the database using those features. Figure is taken from~\citep{place_recognition}.}}
    \label{fig:place_recognition}
\end{figure*}

\begin{figure}
    \centering
    \includegraphics[width=.45\textwidth]{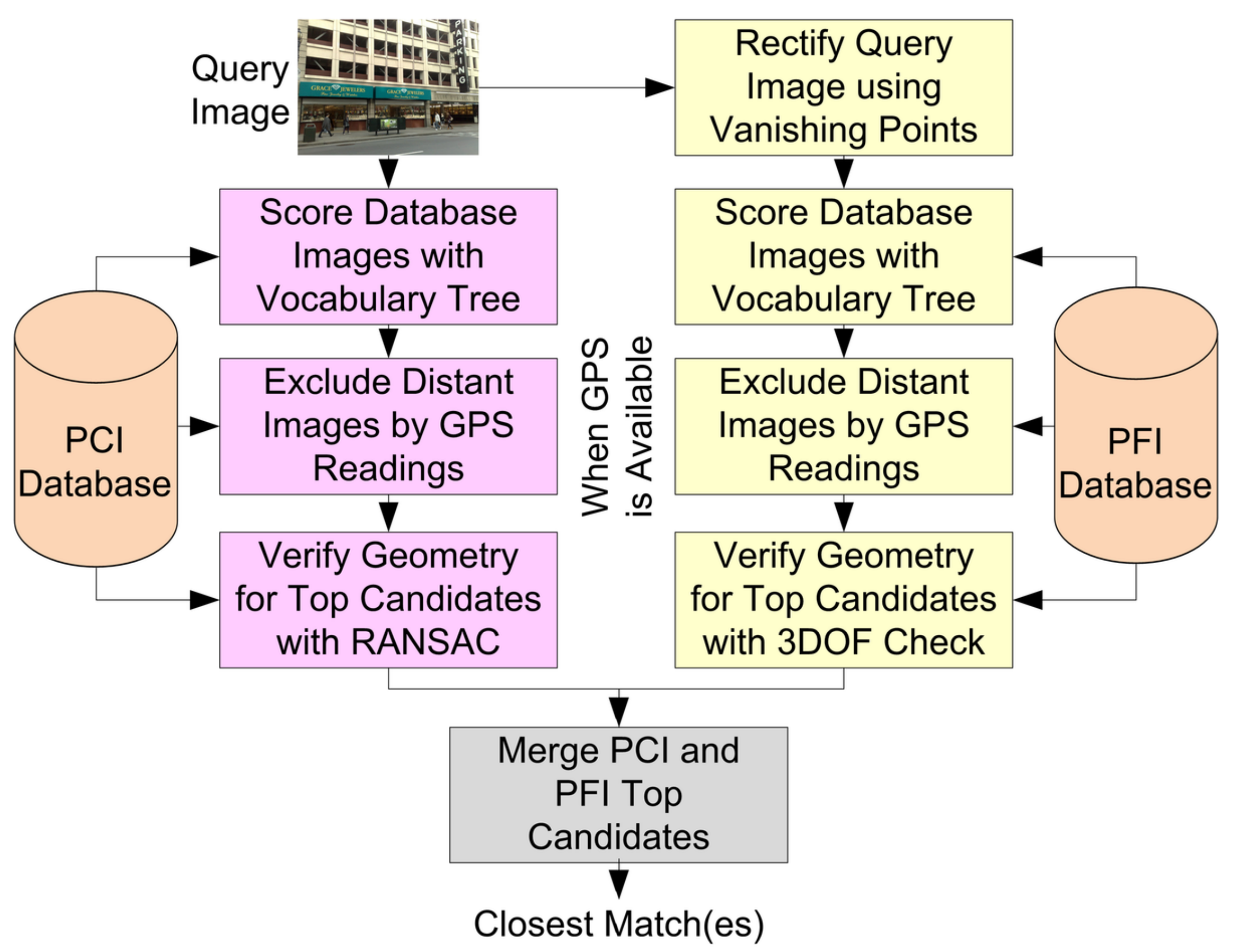}
    \caption{The pipeline proposed by~\citep{city_landmark}. The perspective central images (left) and perspective frontal images (right) are compared against separate databases in parallel. The predictions from the separate perspectives are merged to generate the candidate list to perform city-scale location recognition. Figure is taken from~\citep{city_landmark}.}
    \label{fig:city_landmark}
\end{figure}

% used synthetic perspectives, attempting to harvest features invariant of camera angle, also used better descriptor vlad as opposed to sift
% disadvantage is synthetic never as good as real images
A crucial challenge faced by reference-based approaches is that the camera images will have imperfect alignments with images from the reference database. To address this fundamental challenge, some work has been proposed attempting to produce features invariant of camera perspective to make matching more robust.~\citep{place_recognition} used depth map panorama images \textcolor{red}{to construct synthetic images} from multiple 'virtual' camera locations using ray tracing and bi-linear interpolation. To match the queries to their reference database, they extracted multi-scale SIFT descriptors which were aggregated using a VLAD~\citep{vlad} descriptor followed by principal component analysis (PCA)~\citep{pca} and $L_2$ normalization. This two-stage approach is shown in Figure~\ref{fig:place_recognition}.~\citep{city_landmark} proposed a similar method using facade-aligned and viewpoint-aligned images for improved geo-localization. As shown in Figure~\ref{fig:city_landmark}, they used two parallel pipelines; one of which processed images generated from the center of the panorama (PCI), and the other used projective geometry to transform the panorama to a frontal view (PFI). A fused vocabulary representation was matched to the query image to perform geo-localization. Compared to other methods, both these approaches attempt to overcome the challenge associated with camera alignment by mining additional features from synthetic images. While this results in features that are more robust to the camera perspective, the disadvantage is that synthetic image features are less accurate than `real' features from actual reference images.
% resulting in a fixed size feature descriptor
% match across scene images in which query and database image depict a scene from the same viewpoint
% build compact indexable image representation using VLAD encoding to compress images
% represent images using SIFT features from multiple scales
% robust representation that does not rely on location-specific features and perspective changes
% use depth map associated each panorama
% synthesize virtual camera locations using google maps panorama images and depth maps using ray tracing and bilinear interpolation
% and then synthesize individual views
% combine real and virtual views into a single database
% to perform geo-localization, they extract SIFT descriptors at 4 scales to extract a visual vocabulary, aggregated using VLAD descriptor followed by PCA and then L2 norm
% results in a consistent size, measure similarity between a query and other images using the dot product

\begin{figure}
    \centering
    \includegraphics[width=0.5\textwidth]{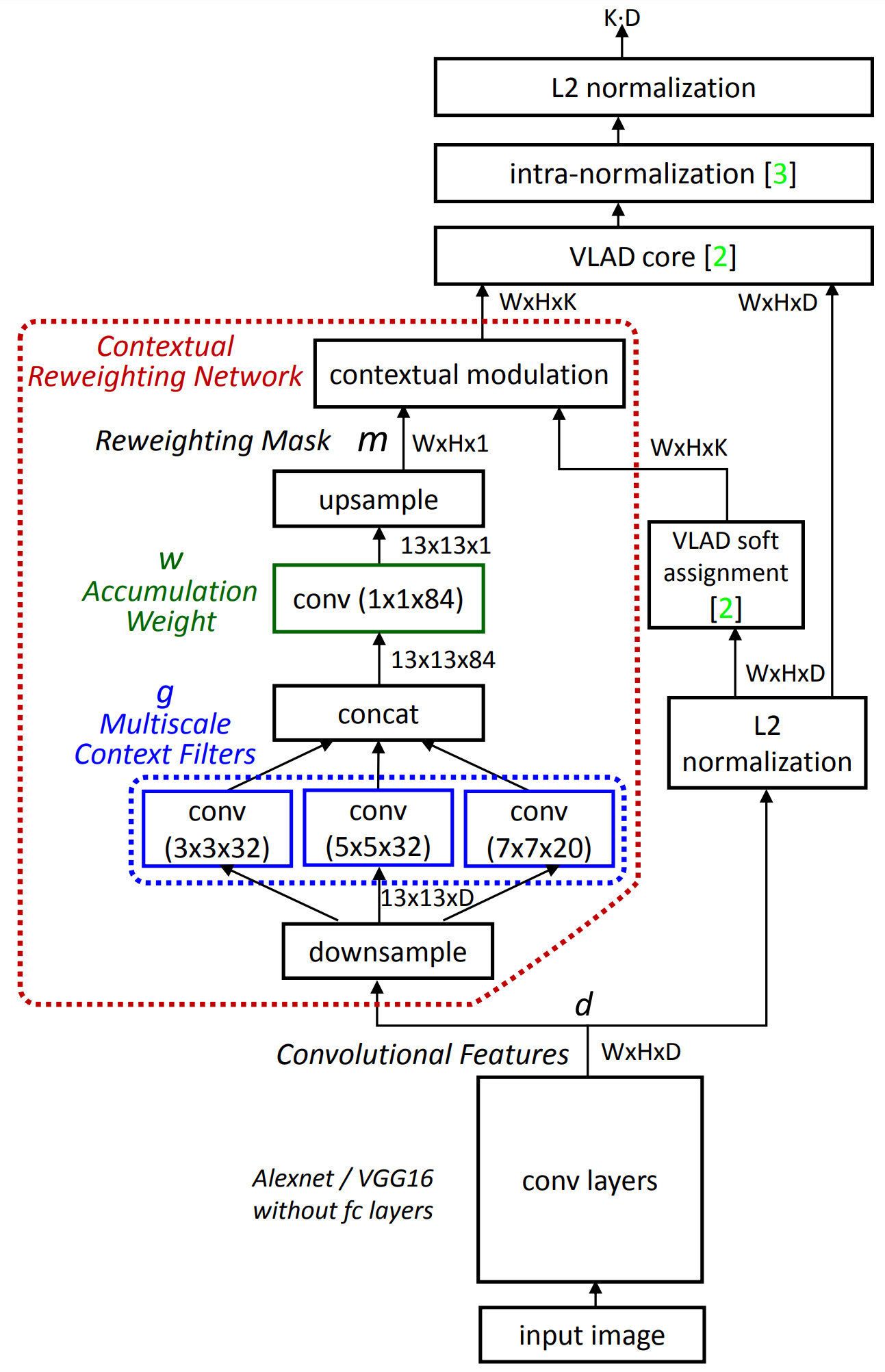}
    \caption{The neural network architecture proposed by~\citep{feature_reweighting}. The authors add a "contextual re-weighting network" (inside red outline) which learns to weight the relative importance of certain features based on the context of the image. Figure is taken from~\citep{feature_reweighting}.}
    \label{fig:feature_reweighting}
\end{figure}

%  Contribution is explicitly predicting usefulness of features
% disadvantages of this concept are they are more complex and only learn to ignore bad features not fundamentally improve them
To address the challenge of finding the most effective matching features, other approaches have attempted to explicitly predict how `useful' a feature is, and then re-weight or remove the features based on their usefulness.~\citep{predicting_features} \textcolor{red}{extracted} MSER (maximally stable extremal regions) and SIFT keypoint descriptors and proposed a per-bundle vector of locally aggregated descriptors (PBVLAD) to convert the features into a vector of fixed size. They trained an SVM to make a binary prediction indicating the `usefulness' of each feature. Only features with high predicted usefulness were used to perform image matching. An important limitation of their approach was \textcolor{red}{its} capabilities were limited to predicting locations depicted within cities.~\citep{feature_reweighting} designed an end-to-end trainable convolutional network displayed in Figure~\ref{fig:feature_reweighting}, to compute a spatial re-weighting of features in an unsupervised manner. The intuition behind this approach is to enable the model to focus on only features relevant to the particular task. Compared to other reference-based approaches, these methods benefit from the capability to ignore less informative features, which are arguably common in large-scale geo-localization applications. The main shortcoming of these approaches is that while re-weighting reduces the impact of less useful features, it does not inherently improve the features themselves.

% The authors built their dataset using Google Street View images from the Pittsburgh area.

%the authors geolocate images by matching them to a reference image depicting the same place with a known GPS location. They use a convolutional model containing a contextual re-weighting network, which learns to compute a spatial re-weighting of features in an unsupervised manner. 

%The intuition behind this approach is it enables the model to focus on only features relevant to the particular task. Their model has the additional benefit of being end-to-end trainable. Because they are matching unlabeled images with reference images containing known GPS locations, no additional supervision is required.
% unique approach to image geo-localization, they look for geotagged images identifying the same place
% they learn context-aware features to focus on interesting and relevant parts of an image
% learn to re-weight features in an unsupervised manner using Contextual Reweighting Network (CRN) which takes original features as input and outputs spatial weighting of those features
% model is learned end-to-end and requires no additional annotations
% problem is framed as an image retrieval task and the model is trained using triplet loss
% use VLAD as aggregated descriptors to perform image matching

\begin{figure}
    \centering
    \includegraphics[width=0.5\textwidth]{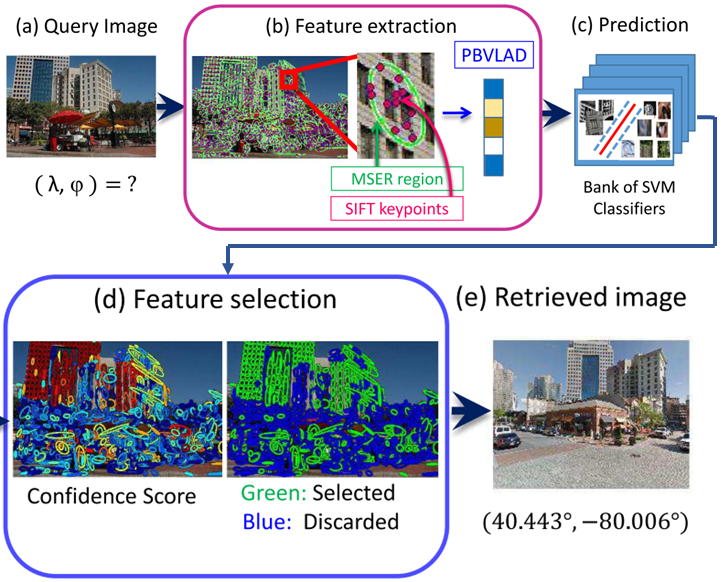}
    \caption{An overview of the approach proposed by~\citep{predicting_features}. Their pipeline extracts features from a query image, and then sends those features through SVMs which are trained to predict how useful each feature is. Only the features predicted to be useful are used for geo-localization. Figure is taken from~\citep{predicting_features}.}
    \label{fig:predicing_features}
\end{figure}

% goal is to recognize place depicted in city image
% dataset contains geo-tagged images at city scale
% focus on extracting useful features with high matching scores using learned classifier (SVM)
% feature score prediction as classification problem
% propose per bundle vector of locally aggregated descriptors they call PBVLAD
% vector of locally aggregated descriptors (VLAD) used for fixed vector size to use SVM
% extract MSER regions and SIFT keypoints as features, PBVLAD gives fixed vector, SVM makes binary prediction of how good each feature is for gelocalization
% features with high predicted confidence scores are used for geo-localization

% A multiple nearest neighbor feature matching
% method based on Generalized Minimum Clique
% Graphs.
%  A novel framework for incorporating both local and
% global features in image geo-localization.
%  A new data set of high resolution street view images.
% use dataset containing 102k google street view images covering pittsburg, orlando, and manhattan

% \begin{figure}
%     \centering
%     \includegraphics[width=.5\textwidth]{single-fig-11-nearest_neighbor.png}
%     \caption{A block diagram of the approach proposed by~\citep{nearest_neighbor}. Their model retrieves query features from a k-means tree to perform a GMCP-based feature matching. Figure is taken from~\citep{nearest_neighbor}.}
%     \label{fig:nearest_neighbor}
% \end{figure}

\begin{figure*}
    % \begin{mdframed}[backgroundcolor=red!50,linecolor=red!50]
    \centering
    \includegraphics[width=1.0\textwidth]{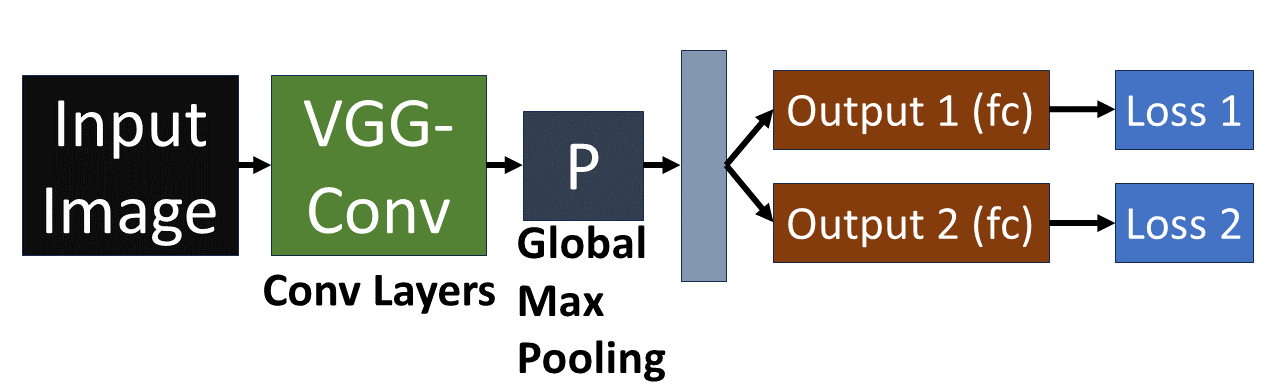}
    \caption{\textcolor{red}{The deep learning model used by~\cite{deep_era} to perform image geo-localization. The first loss penalizes the network for incorrectly classifying the cell an image belongs to, and the second loss is a `distance metric' by which images less distant to each other should have more similar features. Figure is taken from~\citep{deep_era}.}}
    \label{fig:deep_era}
    % \end{mdframed}
\end{figure*}

\begin{figure*}
    \centering
    \includegraphics[width=\textwidth]{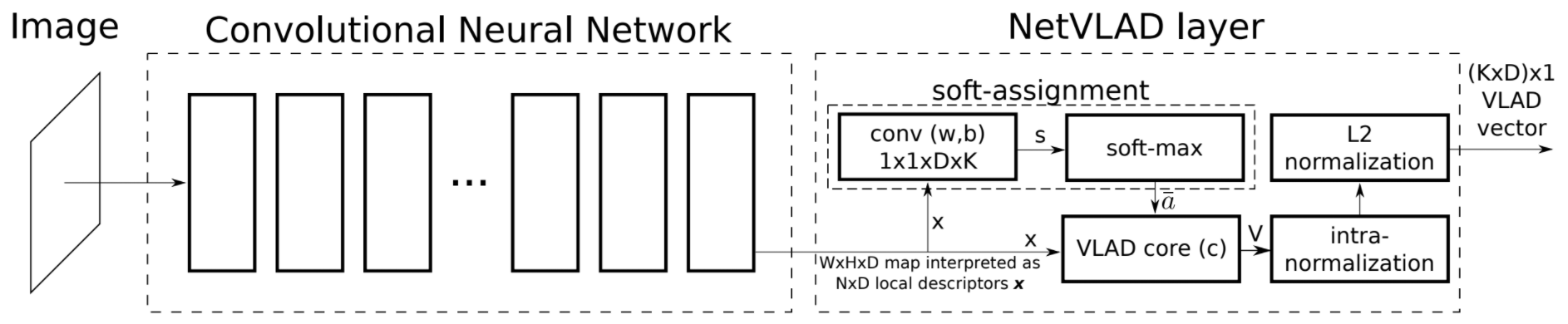}
    \caption{A diagram showing how~\citep{netvlad} modified a traditional CNN with their proposed VLAD layer. The layer contains 1x1 convolutions followed by a softmax function which are provided as inputs to aggregate features in the "VLAD core". Figure is taken from~\citep{netvlad}.}
    \label{fig:netvlad}
\end{figure*}

% First deep learning approach
% Improved vlad with netvlad for deep learning
% deep learning very powerful but requires strong computational hardware and large dataset
The rise of deep learning algorithms and their capability to automate the extraction of descriptive features posed a clear opportunity to further enhance performance.~\citep{deep_era} extended the work of~\citep{im2gps} using a modern deep learning architecture displayed in Figure~\ref{fig:deep_era}. The authors proposed a cell-based approach in which the network simply learned to classify the query image's cell. Second, they proposed a distance metric learning approach in which a model was trained to recognize a pair of images from nearby locations as having greater similarity. They matched the query image to similar reference images using a K-nearest neighbors.~\citep{netvlad} proposed netVLAD, a fully convolutional modification of the vector of locally aggregated descriptors (VLAD) commonly used by hand-crafted approaches. NetVLAD mimics VLAD's pooling layer using an end-to-end trainable differentiable model optimized with a weakly supervised ranking loss.  %Unlike previously proposed solutions, a key improvement was their differentiable VLAD implementation enabled the VLAD parameters to be learned as opposed to fixed. 
The NetVLAD architecture is displayed in Figure~\ref{fig:netvlad}. 
Compared to other approaches, deep learning models have the obvious advantage of being high-performance and automating the construction of powerful feature representations. The main limitations of these approaches are the need for powerful computational hardware and large datasets required to train these CNN models.

% \begin{figure}
%     \centering
%     \includegraphics[width=0.5\textwidth]{single-fig-16-mountain_images.png}
%     \caption{An overview of the approach proposed by~\citep{mountain_images}. Figure is taken from~\citep{mountain_images}.}
%     \label{fig:mountain_images}
% \end{figure}

\subsubsection{GPS Refinement Approaches}

There exist many images containing inaccurate GPS coordinates due to inadequate hardware, low-quality cameras, or other sources of noise. Similarly, due to compression or storage restraints especially in photo albums, GPS information may only be available intermittently. Some researchers have proposed approaches to refine the noisy GPS locations using additional information including motion vectors between frames and extra metadata available from images.

%Sometimes images of interest contain inaccurate GPS coordinates due to inadequate hardware, low quality, small size or noise. Similarly, In other scenarios, GPS information may only be available intermittently due to compression or storage restraints. The approaches discussed in this section attempt to refine these limitations labels using additional information including motion vectors between frames and extra metadata available for each image.

\citet{tag_refinemnt} proposed an algorithm that refines GPS predictions using other images at nearby coordinates. The authors construct triplets of images with similar SIFT features containing a query image and two reference images. Structure from motion is used to compute a GPS prediction for each query image in the triplet. To reduce the noise present in the GPS tags, the authors perform random walks using an adaptive dampening factor to find subsets of data with maximal agreement. The authors constructed their dataset from user-shared images of American cities and images were downloaded from Panoramio, Flickr, and Picasa. Inspired by~\citep{tag_refinemnt},~\citep{fusion} proposed a similar approach that also extracted SIFT features from images and performed random walks with an adaptive dampening factor to denoise GPS coordinates. They further expanded the system's capabilities to geo-localize videos in order to achieve more accurate and consistent GPS signals across video frames.

Compared to other geo-localization methods, these approaches achieve superior accuracy due to refining existing GPS coordinates as opposed to predicting them from scratch. The obvious limitation of these approaches is that these techniques require noisy GPS coordinates are already present in the dataset, making their application somewhat niche.

\subsubsection{\textcolor{red}{Aerial-View} Geo-localization Approaches}

\textcolor{red}{Aerial-View geo-localization} algorithms attempt to regress a set of GPS coordinates for aerial images. This is in contrast to traditional reference-based image geo-localization, where the images are taken from the ground perspective. We also note that this is different from cross-view geo-localization approaches, which jointly \textcolor{red}{learn} from both ground and aerial images to geo-localize the ground images~\citep{tian2017cross, rodrigues2021these, liu2019lending, bridging}.

In \textcolor{red}{aerial-view} images, roads and intersections contain 
discriminative features, which prompted~\citep{matching_roads} to propose a pipeline to perform geo-localization by matching roads and intersections against their \textcolor{red}{known locations.} The first stage of the algorithm performed image segmentation to determine road pixels in \textcolor{red}{aerial-view} images, and the second stage used the detected road pixels to identify intersections. The detected intersections were matched to known intersection locations to find the relative offset of the \textcolor{red}{aerial-view} image, from which its GPS coordinates could be calculated. The authors also constructed a geo-localization dataset containing \textcolor{red}{aerial-view} images from two European cities by collecting ground truth information from Openstreetmap\footnote{\label{foot:open_street_map}www.openstreetmap.org}. The advantage of this approach is its capability to use intersections as a reliable feature for consistently high performance. However, an important limitation is that \textcolor{red}{aerial-view} images are less accessible than the ground-view images used by other approaches. Additionally, this method can not geo-localize images in which roads and intersections are not visible.
% pipeline that performs geo-localization from finding roads and intersections in aerial images
% match detected intersections to labeled ones
% new dataset with images from two aerial European cities
% collect ground truth from OpenStreetMap which is what algorithm matches to since they are labeled
% algorithm is divided into stages, 1st stage classifies pixels part of the road, 2nd the detected roads are used to detect intersections, intersections are matched to OpenStreetMap, this is used to find geometric alignment

% \begin{figure*}
%     \centering
%     \includegraphics[width=0.75\textwidth]{single-fig-6-matching_roads.png}
%     \caption{An overview of the geo-localization framework proposed by~\citep{matching_roads}. Figure is taken from~\citep{matching_roads}.}
%     \label{fig:matching_roads}
% \end{figure*}

\subsubsection{Rural Geo-Localization}\label{sec:rural_geolocalization}
% \textcolor{red}{
% These approaches use a digital elevation model to regress and object's pose relative to terrain.
% Approach:
% Extract features, typically skyline horizon features, from image
% Used to align image with digital elevation model to find camera pose
% camera pose used to locate image within local terrain
% Discuss papers
% Advantages:
% less hardware intensive
% Limitations:
% very accurate within local terrain
% limited to local area, not scalable for larger areas
% require visible horizon, not suitable for cities
% struggle with less identifiable skylines, most approaches use mountain areas
% methods are designed with need for view of distant terrain
% most methods assume reasonable camera orientation
% Recommendations moving forward:
% is not sufficient for full scale geo-localization
% should be combined with existing approaches
% }

The most fundamental challenge in global-scale geo-localization is that the query image could be taken from literally anywhere in the world. At many locations, photos may not contain sufficiently descriptive features for geo-localization to even be possible, such as indoor photos that don't contain any geography-specific information. Therefore, many approaches have chosen to restrict their methodology to rural images taken from a limited area~\citep{geopose3k, large_mountains, topographic_maps}. 

Most early approaches were designed to exploit the fact that in contrast to city images, rural images almost always have their skyline features (mountain ranges, etc.,) visible.~\citep{large_mountains, alps} used a dynamic programming algorithm to segment the skyline from images near the Alps. From the segmented skyline, they extracted contour words encoding characteristics about both the shape and order of skyline features. They employed a voting scheme to match these contourlets to the query image's location and direction in a digital elevation model (DEM). This approach was expanded upon in~\citep{topographic_maps}, which used segmentation to classify each pixel from an image as being part of the sky, water, a settlement, or vegetation. The segmented images could be used to construct a textured 3D model of the surrounding terrain to assist in geo-localizing the images.~\citep{mountain_images} expanded the extracted features to include ridges of the tops of hills and mountains, which provided additional features to further improve performance.

% To geolocalize images that contain mountains,~\citep{mountain_images} proposed to employ features representing the skyline. They built a database containing ridge and skyline features extracted from their reference images. To localize a query image, coarse skyline features were extracted. The authors employed a voting method to vote for which portion of a geospatial grid \textcolor{red}{the image belonged} to. They further refined this prediction at a fine-grained pixel level by projecting the query image's pixels to real-world coordinates, and then minimized the weighted sum of the projected \textcolor{red}{coordinate's} error. Furthermore, their method was capable of predicting camera roll angle, which they used to rotate the skyline to make the rotated image consistent with other images from the dataset. %The authors claim this enhances results due to their finding that even a couple of degrees of camera rotation can substantially effect performance

Recently proposed approaches have incorporated deep learning techniques.~\citep{bucolic_environment} utilized a CNN to extract semantic labels from each pixel in an image. They performed Canny edge detection to build an edge-based wavelet-transformed image. Similar to the approaches discussed in Section~\ref{sec:reference_based}, the images are geo-localized by comparing the Euclidean distance between features from the query image and a database of reference images. The main advantage of this approach over other proposed methods is that the extracted features are less susceptible to changes in vegetation, weather, and illumination.~\citep{cross_locate} employ a CNN designed to encode a feature representation for both the query and reference image. However, instead of constructing a \textcolor{red}{database of} reference images, the authors used a digital elevation model to construct synthetic reference images. This approach has the advantage of not relying on the availability of a database of reference images like many other approaches. The trade-off, however, is that synthetic reference images are unlikely to be as accurate as real reference images.

Compared to other geo-localization approaches, rural geo-localization algorithms are typically designed to operate within a more limited area, which means they do not require sophisticated hardware like other methods. For example,~\citep{landscape_ar} uses structure-from-motion to build a 3D model aligned to terrain from which artificial images can be rendered. The authors geo-locate the image by modeling the cross-domain feature correspondences between the rendered and query images. The full system is lightweight and can be run on a mobile device in real-time for augmented reality applications.~\citep{immersive_trip} also uses structure from motion techniques to build a virtual digital elevation model of terrain from a sequence of images. They use embedded GPS metadata to provide a rough location, and then further align the query images using a DEM. Similarly to~\citep{landscape_ar}, their method is designed to be lightweight and can be run on weak hardware.

Rural geo-localization algorithms directly address the shortcomings of other geo-localization approaches in rural environments where there are fewer descriptive features, primarily through the use of digital elevation models. Many of these approaches can also be run on more accessible hardware. These approaches do have many inherent limitations, however. Since all these methods rely on skyline features to some extent, they can only be applied to rural environments where these features are not blocked by man-made objects. Even in natural environments, these methods require image locations with distinguishable skylines, such as mountain ranges, to ensure the extracted features are descriptive enough to locate the images. These methods also assume the camera is positioned at a reasonably horizontal angle when calculating camera pose, however, it is generally agreed that this is a valid assumption in practice~\citep{large_mountains, alps, landscape_ar}. Perhaps the largest limitation of these approaches is that their scalability with larger geographical areas is unclear. As the locations from which these images are harvested grow, terrain and skyline features will become increasingly repeated across new locations. Future research may need to explore hybrid approaches combining multiple geo-localization methods for the best results.

\subsubsection{Other Approaches}
Since image geo-localization is a broad field encompassing a variety of approaches, there are many methods that are loosely related to single image geo-localization but do not decisively fit into a single category. Some of these approaches propose algorithms that involve matching queries to reference images, so a similar method could be re-used to perform reference-based geo-localization. Other approaches only perform visual localization on a smaller scale, but could potentially be scaled up to perform global localization. In this section, we briefly review some approaches that are loosely related to the field of single-image geo-localization.

~\citep{matching_roads} proposed a pipeline to perform geo-localization by finding roads and intersections in aerial images, and then match them against their known locations. The first stage of the algorithm performs image segmentation to determine road pixels in aerial images, and the second stage uses the detected road pixels to identify intersections. The intersections were matched to known intersection locations to align the satellite image and predict its GPS location. The authors also contributed an aerial geo-localization dataset mined from Openstreetmap\footnote{\label{foot:open_street_map}www.openstreetmap.org}. The advantage of this approach is its high performance and interpretability compared to other methods, mainly due to the use of intersections as a reliable feature for consistent matching. %However, satellite images \textcolor{red}{require specialized hardware and are thus less accessible than ground-view images. 
 However, this method cannot geo-localize images in which roads and intersections are not visible.

\citep{visual_similarity} proposed a method of representing the semantic similarity between images despite how different they may be on a pixel level. This approach could be extended to perform geo-localization by applying their matching algorithm to find other nearby images with known GPS coordinates.~\citep{uncertainity_relocalization} proposed a CNN architecture and Bayesian network capable of regressing the 6-DOF (degree of freedom) camera poses of RGB images. Since their method determines pose relative to a reference point, it is comparable to geo-localization approaches except on a smaller scale.
%Similarly,  Kendall et al., \citep{uncertainity_relocalization} proposed a Bayesian network which is also capable of localization 6-DOF poses.

\citep{moving_camera} proposed a method of predicting the geo-spatial trajectory of a moving camera from a sequence of images extracted from a video. Part of their approach involves geo-localizing individual images similar to traditional single-image geo-localization approaches.~\citep{image_sequence} proposed a related method in which the trajectory and sequence of timestamps are used to predict the geolocation of the next image.~\citep{building_rome} built a system that constructed 3D models of cities from images. Their approach extracted features for each available image within a city, detected similar features to find matching interest points, and then merged these features with a matching graph. Since their algorithm built the 3D geometry of the area, it could be extended using techniques such as~\citep{tag_refinemnt} to predict GPS coordinates for images.

% image geo-localization is closely related to some other fields of research, which should be at least briefly discussed: (1) Landmark recognition or place recognition, as surveyed in [1,2][b], (2) camera pose estimation, as surveyed in section 4 of [2], SLAM, as surveyed in [3] and Image Retrieval and image descriptors[c], as surveyed in [4,5]. The survey also overlaps in scope with other surveys [1,2], which should be made explicit.

%Before providing an in-depth survey of these four categories of single image geolocalization, we briefly note that some authors have proposed algorithms designed to detect objects from satellite images. 

\subsection{Datasets}
In this section, we will discuss popular datasets that have been collected for training and testing single-view geo-localization models. We will cover the size of \textcolor{red}{each} dataset, the area from which the images were taken, and any other noteworthy characteristics unique to each dataset. A summary of all the datasets is provided in Table~\ref{tab:summary_single_view_datasets}. %We will organize these datasets into two subsections; the first will describe datasets that have been made pulicly available and the second section will describe private datasets.

%\subsubsection{Public Datasets}
\noindent\textbf{Im2GPS:}\footnote{http://graphics.cs.cmu.edu/projects/im2gps/} ~\cite{im2gps} built a dataset that has become a standard benchmark in the field. This dataset contains approximately 20 million images. Each image has metadata indicating the GPS coordinates at which the image was taken and geographic keywords mined from websites such as Flickr\footnote{https://www.flickr.com/}. The geographic keywords are composed of terms indicating the location of the image, such as a city, state, or tourist location. The authors excluded images with tags unrelated to geolocation (concert, birthday, pets, etc.,), to avoid filling the dataset with images that do not contain descriptive features for geo-localization.

\noindent\textbf{Im2GPS Test:}\footnote{http://graphics.cs.cmu.edu/projects/im2gps/} ~\cite{im2gps} also provided a small test dataset with the same structure as the training set. It contains 237 images.

\noindent\textbf{Img2GPS3k:}\footnote{http://www.mediafire.com/file/7ht7sn78q27o9we/im2gps3ktest.zip/file} \cite{deep_era} constructed a test set composed of 3000 images taken from the Im2GPS dataset, and hence it contains the same characteristics as described above. Due to its large size and high-quality images compared to other proposed datasets, this dataset has become accepted as a commonly used benchmark amongst researchers in the field.

\noindent\textbf{City-Landmark:}\footnote{http://www.nn4d.com/sanfranciscolandmark} \cite{city_landmark} gathered data using a LIDAR (Light Detection and Ranging) system which captured 8-megapixel panoramas at 15 frames per second. The authors built a dataset containing over 150,000 GPS-tagged panoramic images in San Francisco. They aligned their panoramas with previously available 3D models of over 14,000 buildings in the city using projective geometry and LIDAR. The query images consist of GPS-tagged camera images captured by people using commonly available mobile devices, which makes this dataset especially representative of real-world applications.

\noindent\textbf{KITTI:}\footnote{http://www.cvlibs.net/datasets/kitti/} Another popular dataset in the field is contributed by~\cite{kitti}. Data was captured using a road vehicle with a high-resolution camera, a GPS sensor, and a laser scanner to create ground truth labels with a high accuracy compared to other geo-localization datasets. The dataset includes varying environments including roads and highways. Unlike other geo-localization datasets, its applications are not limited to object geo-localization. Other applications include studying stereo and optical flow, visual optometry, and 3D object detection. The authors also provided an online site for submitting and comparing performance results.

\noindent\textbf{YFCC100M:}\footnote{http://www.multimediacommons.org/} \cite{yahoo_flickr} proposed the Yahoo Flickr Creative Commons (YFCC100M) dataset which contains 99.2 million images and 0.8 million videos. Each media object in the dataset contains metadata identifying the user that created it, the camera with which it was taken, and its GPS coordinates. In addition, each media object contains user-annotated tags describing what type of content it contains, such as a baby, park, etc. This dataset was not exclusively intended for image geo-localization, but due to the availability of GPS metadata, it is commonly applied to this domain.

\noindent\textbf{GeoPose3k:}\footnote{http://cphoto.fit.vutbr.cz/geoPose3K/}~\citep{geopose3k} constructed a dataset designed for the rural geo-localization techniques discussed in Section~\ref{sec:rural_geolocalization}. They collected over 3,000 images of the Alps from Flickr. Each image has an annotated GPS position, camera field of view, and image orientation. The dataset also provides synthetic depth maps, normal maps, illumination maps, and semantic labels for each image.

\begin{table*}[t]
% \begin{mdframed}[backgroundcolor=red!50,linecolor=red!50]
    \tiny
    \centering
    \begin{tabular}{c c c c}
    Name & Approximate Number of Images & Locations & Special Characteristics \\ \toprule
    Im2GPS & 20,000,000 & Global & Images contain associated keywords \\
    Im2GPS Test & 237 & Global & Im2GPS test set \\
    Im2GPS3k & 3000 & Global & subset of Im2GPS dataset \\
    City-Landmark & 150,000 & San Francisco & images are panoramic \\
    KITTI & 12919 & Global & Designed for many other tasks such as object detection. \\
    YFCC100M & 99,200,200 & Global & Images contain user annotated tags \\
    GeoPose3k & 3000 & Alps & Each image has depth map, normal map, illumination map, and semantic labels. \\
    \end{tabular}
    \caption{Summary of single view geo-localization datasets.}
    \label{tab:summary_single_view_datasets}
% \end{mdframed}
\end{table*}
\subsection{Evaluation Metrics}\label{sec:single_metrics}
To quantify the performance of proposed models, consistent metrics must be adopted for measuring how accurately images are geolocalized relative to their ground truth. Unfortunately, many authors have chosen to construct their own unique metrics for their specific methods and datasets, making comparisons challenging. There are, however, two standard performance metrics which have been commonly adopted and are described in this section.\newline

\noindent\textbf{Threshold Accuracy:} The most universally accepted evaluation metric in this field involves defining distance thresholds and computing the percentage of images whose predicted coordinates lie within that threshold of the ground truth coordinates. Specifically, evaluation is performed using the following formula:
\begin{equ}[!h]\vspace{-0.23in}
  \begin{equation}
    a_r = \frac{1}{N} \times \sum_{i=1}^{N} u[geodist(d_{gt}^{i}, d_{pred}^{i}) < r].\
  \end{equation} 
\end{equ}\label{equ:threshold_benchmark}\vspace{-0.23in}
In this equation, $N$ represents the number of images during testing, $r$ indicates the distance threshold within which an image is considered to be geolocalized correctly, and u[.] is an indicator function that returns $1$ if the geo-localization threshold is less than $r$ and $0$ otherwise. Typically, this metric is reported using multiple thresholds to quantify geo-localization performance at different scales such as street, city, and country.\newline

\noindent\textbf{Reference Ranking:} Another way of benchmarking performance is to list out the top `N' reference images matched to the query by the geo-localization algorithm. If one or more of the top `N' reference images lie within a selected threshold of the ground truth coordinates, the image is considered to be correctly geolocalized. Since this evaluation method requires comparing the query to the top N reference images, this evaluation method is only applicable to reference-based approaches. In practice, benchmarks using this metric use a threshold of 25 meters unless otherwise specified.
\subsection{Benchmarks}\label{Single_view: sec:Benchmarks}
In this section, we briefly note the experimental results of the methods discussed in Section~\ref{Techniquessingle view image geo-localization}. Benchmarks on the Im2GPS and IM2GPS3k datasets are provided in Tables~\ref{tab:single_view_performance_1} and~\ref{tab:single_view_performance_2} respectively and use the threshold accuracy evaluation metric discussed in~\ref{sec:single_metrics}. Benchmarks on the Tokyo 24/7 dataset are shown in~\ref{tab:tokyo_performance} and use the reference ranking metric discussed in~\ref{sec:single_metrics}.

%\subsubsection{Public Benchmarks}
% \noindent\textbf{Experimental results on Img2GPS:} 
% Many proposed approaches report results on the Im2GPS dataset using the threshold accuracy metric discussed in Section~\ref{sec:single_metrics}. We have aggregated the reported results in Table~\ref{tab:single_view_performance_1}. We observe that the original Im2GPS approach proposed by~\cite{im2gps} has the weakest performance. This is to be expected, as this was the first method proposed and was only intended to serve as a baseline. These results were later improved by~\cite{large_scale}, which provided a large performance improvement at every distance threshold measured. The new method proposed by~\cite{planet} improved performance at every distance except 2500km.~\cite{cplanet} further improved geo-localization performance. In particular, they increased the percentage of images geo-localized within 1 km and 25 km thresholds. Finally,~\cite{hierarchical} provided a final improvement to performance results at all measured distances.

\begin{table}[t]
     \footnotesize
    \centering
    \tiny
    % \caption{Im2GPS Dataset Benchmarks}
    \begin{tabular}{c r r r r r}
    \multicolumn{6}{c}{\textbf{Im2GPS Dataset}} \\
    \toprule \toprule
         Method & 1 km & 25 km & 200 km & 750 km & 2500 km \\
         \midrule
        IM2GPS~\cite{im2gps} & N/A & 12.0 & 15.0 & 23.0 & 47.0 \\
        %  Hays et al., 
        SVM-KNN~\cite{large_scale} & 2.5 & 21.9 & 32.1 & 35.4 & 71.3 \\
        PlaNet~\cite{planet} & 8.4 & 24.5 & 37.6 & 53.6 & 71.3 \\
        CPlaNet~\cite{cplanet} & 16.5 & 37.1 & 46.4 & 62.0 & 78.5 \\
        %  Budack et al., 
        ISNs~\cite{hierarchical} & 16.9 & 43.0 & 51.9 & 66.7 & 80.2 \\
        \textcolor{red}{Translocator~\cite{translocator}} & \textcolor{red}{19.9} & \textcolor{red}{48.1} & \textcolor{red}{64.6} & \textcolor{red}{75.6} & \textcolor{red}{86.7} \\
        \textcolor{red}{Hierarchies and Scenes~\cite{hierarchies_and_scenes}} & \textcolor{red}{\textbf{22.1}} & \textcolor{red}{\textbf{50.2}} & \textcolor{red}{\textbf{69.0}} & \textcolor{red}{\textbf{80.0}} & \textcolor{red}{\textbf{89.1}} \\
    \end{tabular}
    \caption{The percentage of images correctly geo-localized within the specified distance threshold on the Im2GPS dataset.}
    \label{tab:single_view_performance_1}
\end{table}

\begin{table}[t]
     \tiny
    \centering
    \begin{tabular}{c r r r r r}
    \multicolumn{6}{c}{\textbf{Im2GPS3k Dataset}} \\
    \toprule \toprule
         Method & 1 km & 25 km & 200 km & 750 km & 2500 km \\
         \midrule
        7011C~\cite{deep_era} & 6.8 & 21.9 & 34.6 & 49.4 & 63.7 \\
        ISNs~\cite{hierarchical} & 10.5 & 28.0 & 36.6 & 49.7 & 66.0 \\
        kNN~\cite{deep_era} & 12.2 & 33.3 & 44.3 & 57.4 & 71.3 \\
        \textcolor{red}{Translocator~\cite{translocator}} & \textcolor{red}{11.8} & \textcolor{red}{31.1} & \textcolor{red}{46.7} & \textcolor{red}{58.9} & \textcolor{red}{80.1} \\
        \textcolor{red}{Hierarchies and Scenes~\cite{hierarchies_and_scenes}} & \textcolor{red}{\textbf{12.8}} & \textcolor{red}{\textbf{33.5}} & \textcolor{red}{\textbf{45.9}} & \textcolor{red}{\textbf{61.0}} & \textcolor{red}{\textbf{76.1}} \\
    \end{tabular}
    \caption{The percentage of images correctly geo-localized within the specified distance threshold on the Im2GPS3k.}
    \label{tab:single_view_performance_2}
\end{table}

\begin{table}[]
    \tiny
    \centering
    \begin{tabular}{c r r r r r}
    \multicolumn{6}{c}{\textbf{Tokyo 24/7 Dataset}} \\
    \toprule \toprule
         Method & 1 & 5 & 10 & 20 & 50 \\
         \midrule
         Dense VLAD SYNTH~\cite{place_recognition} & 66.03 & N/A & 75.87 & 80.32 & 85.08 \\
         NetVLAD~cite{netvlad} & ~68 & ~82 & \textbf{~87} & \textbf{~90} & N/A \\
         CRN~\cite{feature_reweighting} & \textbf{75.2} & \textbf{83.8} & \textbf{87.3} & N/A & N/A \\
    \end{tabular}
    \caption{The performance of different methods of the Tokyo 24/7 dataset. Images are considered correctly recalled if one of the top `N' listed images is within 25 meters of the ground truth. Values are expressed as percentages.}
    \label{tab:tokyo_performance}
\end{table}

\subsection{Discussion and Future Work}
% 1. Brief summary of the task. 2. tradeoffs of different approaches, 4. which ideas caused the main improvements, 5. ideas for future work

The fundamental goal of single-view geo-localization is to construct an algorithm that receives a single image perspective as input and predicts the image's geographical location as output. Features are typically extracted from each image, using either a traditional hand-crafted approach or by leveraging modern deep learning architectures. These features can be either used to match the query image against a reference image, classify the image as a part of a cell, or align the image to known terrain features to perform geo-localization.

% reference based is very scalable, but lots of noise, many images required, computationally expensive
% cell based less scalable, tradeoff with cell resolution between training samples and accuracy, location prediction inherently limited in accuracy, advantage outputs propabaility over multiple locations, conceptually more simplistic, single end to end deep learning architecture can be constructed
% rural approaches limited in area and require specific distinctive skyline features in images, however better more consistent performance in their are
% MISSING AERIAL APPROACHES
Each of the discussed methods inherits a set of trade-offs associated with its methodology. Cell-based approaches are conceptually simple because they convert the task into a simple classification problem. With modern deep learning architectures, these approaches enable the construction of a single end-to-end model which directly outputs the cell to which the query image belongs. Models constructed following this paradigm output a probability distribution of the image corresponding to each cell, which may be useful to researchers even if the highest probability cell is not the cell the image belongs to. This simplicity comes at the expense of scalability since a suitable dataset must provide many training samples per each cell. Since training samples per cell decreases as cell size decreases, these methods are limited in their capability to perform accurate geolocalization. Since reference-based approaches match query images to a reference dataset, they can numerically regress the GPS coordinates of an image as opposed to being restricted to predicting a cell. The main disadvantage of these approaches is comparing extracted features from a query to the reference database can be computationally intensive, especially as the approach is scaled to larger databases. GPS refinement approaches produce accurate predictions with minimal effort by taking advantage of preexisting noisy GPS tags. However, they have the obvious disadvantage of requiring the image already contains geospatial information. Finally, geo-localization approaches specialized for rural environments typically predict image geo-locations by aligning them to digital elevation models of the surrounding region. This formulation addresses the reduced set of features present in rural environments but comes at the expense of increased difficulty. Specifically, rural environments tend to contain fewer descriptive features, and a reference image database may be limited or completely unavailable in certain rural regions.

% performance improvements caused by extracting better features
% deep learning was big
% larger datasets, more refining of noise, more attributes data, better filtering of images
Previous performance improvements in this field were achieved by the extraction of more descriptive features that are better at distinguishing an image's geo-location, and are less susceptible to variations in camera perspective, vegetation, and illumination. Further progress was achieved by implementing models designed to predict the `usefulness' of each feature to help in filtering out noise or less useful information. The largest performance improvement was achieved when deep learning models were applied to automatically extract descriptive and powerful features. The influence of larger datasets on performance should also not be overlooked. Due to the size and variety of scenery across the earth, a massive number of images is required to construct models capable of effective performance. Newer datasets have also taken careful steps to filter images that do not contain descriptive geospatial features, such as images taken indoors. Faster matching algorithms have made existing geolocalization approaches more computationally efficient.

% can be improved using "extra" information
% such as scene attributes
% context from internet users
% satellite images
% led to development of cross-view approaches
Future performance improvements can still be achieved by incorporating additional information from datasets. For example, adding scene attributes to images that specify labels such as `outdoor', `concert', etc., may be beneficial to training models. This idea has received some minor exploration by~\citep{hierarchical}. Certain context information available when mining data from the internet, such as the type of website they were taken from, or other images uploaded by the same user could also be used to provide additional context clues. Finally, \textcolor{red}{aerial images} of the surrounding region could serve as another reference to provide an additional perspective to a geo-localization model. Since satellite images are becoming increasingly available, their inclusion is a natural next step to further progress the state of the field. We will discuss this idea in detail in the next section.

\section{Cross View Image Geo-localization}\label{sec:cross-view}

Cross-view image geo-localization approaches were developed to address the limitation of single-view approaches that at least one reference image must be near the query~\citep{lin2013cross}. To tackle cross-view image geo-localization, the problem is usually converted into a retrieval task by matching query ground images to reference aerial images. Note that it has become quite easy to collect dense, high-resolution geo-referenced satellite image datasets, thanks to free public releases of geo-referenced satellite images from large companies e.g., Google Earth. Although appealing, matching ground and satellite images is an extremely challenging task for several reasons: 1) satellite and ground images are most likely captured at different times resulting in different illumination conditions, weather, and objects e.g., cars and people, 2) ground and satellite views contain very different visual contents, i.e., building facades, trees, and cars occupy the majority of ground image scenes, while satellite images mainly contain building-tops, tree-tops, and road structures, 3) ground and satellite images are captured at different resolutions causing ground images to capture finer details while satellite images mainly capture coarse level information. 
% Due to the different image perspectives, objects such as houses, cars, and trees will have different appearances.

Applications of cross-view image geo-localization are diverse. For example, with \textit{weak} GPS signals in metropolitan downtown areas, cross-view image geo-localization can be a supplementary source to estimate the location of a camera~\citep{ChenChen}. With the development of autonomous vehicles, cross-view geo-localization can work with inertial measurement units to provide accurate positioning measurements for autonomous vehicles~\citep{kim2017satellite}. Moreover, with the recent development of Augmented Reality (AR) navigation~\citep{AR1, AR2, AR3}, cross-view image geo-localization can boost the performance of AR navigation in outdoor environments. 

In this section, we categorize existing cross-view image geo-localization methods into four classes: hand-crafted feature representations, graph-based feature matching, deep siamese-like methods, and generative methods. We then introduce evaluation protocols, existing datasets, and benchmark results in the cross-view image geo-localization domain.

%Image In these methods, the geo-location of a query image is obtained
%by finding its matching reference images from the
%same view (e.g. ground-level Google Street View images),
%based on the assumption that a reference dataset consisting
%of geo-tagged images is available. However, such geotagged
%reference data may not be available.}
%\ws{Predicting ground-level scene layout from aerial imagery (CVPR, 2017)}
%\ws { Understanding and mapping natural beauty (ICCV, 2017)}
\subsection{Hand-Crafted Feature Representations}
\label{hand_crafted}
These approaches frame the cross-view geo-localization problem as a retrieval task. They extract features from aerial and ground imagery such that corresponding images from the same location should have similar features. Unlike Siamese-like CNN methods which will be discussed in later sections, these approaches rely on hand-crafted feature extractors to extract representative features from both satellite and ground images. Typically, the extracted features can be directly used for evaluating the similarity by calculating the Euclidean or Cosine distance in feature space. It can also be fed into a machine learning model, such as a support vector machine, to predict the similarity.
\newline\newline
\noindent\textbf{Feature Averaging and Discriminative Translation:}~\citet{lin2013cross} proposed two data-driven cross-view geo-localization approaches: Data-driven Feature Averaging (AVG) and Discriminative Translation (DT). Their dataset is composed of \textcolor{red}{ground-view images, aerial-view images}, and land cover attribute images. Features are extracted from both ground and aerial images using four feature descriptors: HoG~\citep{HoG}, self-similarity~\citep{selfSim}, GIST~\citep{gist}, and color histograms. AVG first matches the query ground view image to the top $k$ matches in the database using the im2gps~\citep{im2gps} algorithm. Then, corresponding aerial images and land cover attribute images from the top $k$ matches are averaged separately to build the prediction features. Finally, by matching the prediction features to the reference database, the place with the closest features is selected as the prediction result. In contrast to the AVG method which only uses the best scene matches to make predictions, the DT method takes advantage of dissimilar ground scenes. In detail, the DT method utilizes the same positive set as the AVG method and adds a negative set from the lowest match samples. A Support Vector Machine (SVM) is trained on these samples as well as on the aerial imagery. The trained SVM can be applied to predict the locations of the query ground images.
\\
\begin{figure*}
    \centering
    \includegraphics[width=\textwidth]{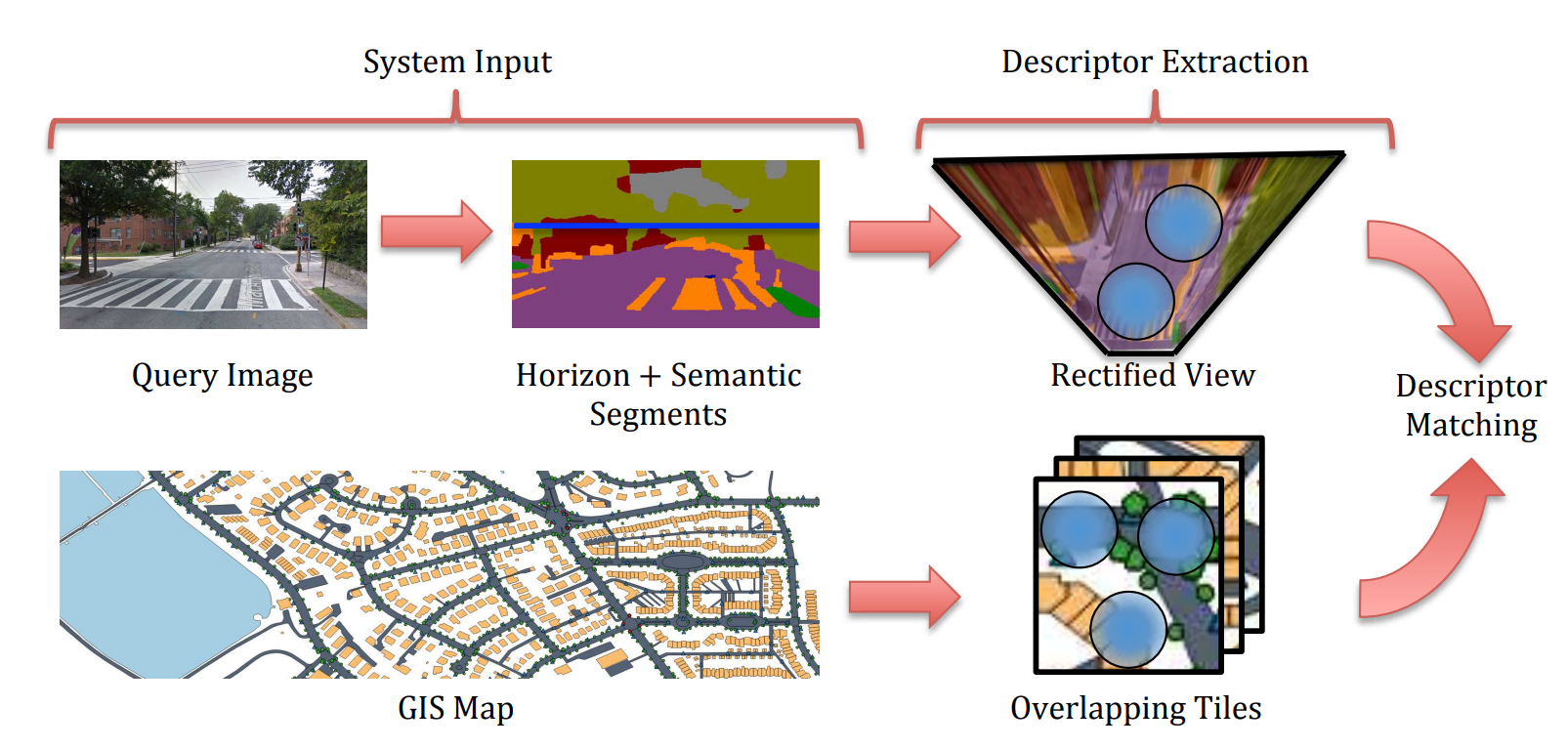}
    \caption{\textcolor{red}{Pipeline of Semantic cross-view matching. \xh{The descriptors extract features from both the query input stack (image and segments) and the tiled GIS map. \textcolor{red}{The estimated location is the GIS tile that is closest to the query image in $L2$ distance.}} Figure is taken from~\citep{castaldo2015semantic}.}}
    \label{Semantic}
\end{figure*}\\
\noindent\textbf{Semantic cross-view matching:}~\citet{castaldo2015semantic} leveraged a geographic information system (GIS) to perform cross-view geo-localization. Their key motivation was that the traditional feature descriptors such as SIFT~\citep{sift} are not useful because of the drastic appearance difference between the views. Unlike other approaches which directly compare features, they proposed to match the semantic segments to a GIS map to address the view angle difference. The ground semantic segments were obtained from an off-the-shelf algorithm~\citep{ren2012rgb}, and were transformed into a rectified view. For feature extraction, the authors proposed a Semantic Segment Layout (SSL) descriptor which is designed to simultaneously capture the semantic information and encode the rough geometric location. SSL is applied on both rectified semantic segments of the query image and the tiles of the GIS map. Finally, by matching the query features to the candidate features from the GIS map using the $L2$ distance, the query images are geo-localized. Their approach is illustrated in Figure~\ref{Semantic}.

\begin{figure*}
% \begin{mdframed}[backgroundcolor=red!50,linecolor=red!50]
    \centering
    \includegraphics[width=\textwidth]{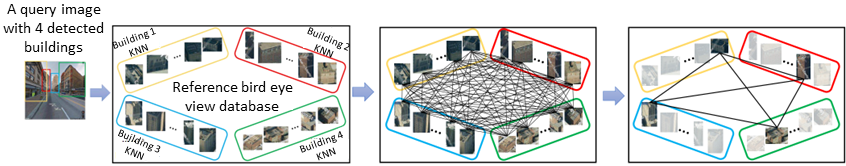}
    \caption{Visualization of the cross-view image geo-localization approach from \citep{tian2017cross}. \textcolor{red}{First all buildings are detected from the query image. Then a building matching model retrieves $K$ nearest neighbors' aerial-view buildings in latent space. Finally, it performs building matching using dominant sets and estimates the location of the query image.} The image is taken from~\citep{tian2017cross}.}
    \label{dominant_set}
% \end{mdframed}
\end{figure*}

% \begin{figure*}
%     \centering
%     \begin{subfigure}[b]{01\textwidth}
%     \centering
%     \includegraphics[width=0.85\textwidth]{cross-fig-1-Dominant_set.PNG}
%     \caption{Visualization of cross-view image geo-localization using building matching through dominant sets. The image is taken from \citep{tian2017cross}.}
%     \label{dominant_set}
%      \end{subfigure}
     
%      \begin{subfigure}[b]{01\textwidth}
%     \centering
%     \includegraphics[width=0.85\textwidth]{cross-fig-2-Landmark.png}
%     \caption{Paired landmarks from a satellite image (left panel) and its corresponding panoramic image (right panel). The image is taken from \citep{verde2020ground}}
%     \label{landmark}
%      \end{subfigure}
%      \caption{Figure (a) is visualization of dominant set based approach \citep{tian2017cross} and (b) shows landmarks based approach \citep{verde2020ground}. }
% \end{figure*}
\subsection{Graph-Based Methods}

% \begin{figure*}
%     \centering
%     \includegraphics[width=0.85\textwidth]{cross-fig-2-Landmark.png}
%     \caption{Paired landmarks from a satellite image (left panel) and its corresponding panoramic image (right panel). The image is taken from \citep{verde2020ground}}
%     \label{landmark}
% \end{figure*}

Graph-based cross-view image geo-localization explicitly constructs a graph according to nearby landmarks such as trees, buildings, and roads. Usually, each node in this graph represents a landmark and each edge represents the connectivity between landmarks. In this section, we present two methods that take the advantage of graphs to perform cross-view geo-localization.~\citep{geo_urban} leveraged dense buildings in urban areas and~\citep{verde2020ground} utilized the relative location of pre-annotated landmarks. Since landmarks are \textit{explicitly} modeled in the graph, the matching result can be easily explained by the correspondence between the two views. Thus, graph-based cross-view geo-localization naturally has better interpretability than other methods introduced in this section.\\
\newline
\noindent\textbf{Building matching:}~\citet{tian2017cross} proposed to match the surrounding buildings in two views to geo-localize in urban areas. The proposed pipeline has five stages: building detection from both overhead and ground images, building matching between cross-view images, retrieval of the nearest $k$ neighbors for each building, dominant set selection, and geo-localization. To detect buildings from ground images, a Fast R-CNN model~\citep{fastrcnn} is employed. For each detected building, a siamese~\citep{siamese} building matching network predicts the relevance between the building and other buildings detected from overhead images. This network is trained using a contrastive loss~\citep{contrastiveLoss} on a dataset customized for building matching. For each building, the top $k$ nearest neighbors are then selected from a reference database as a cluster. After that, a graph is constructed in which the nodes are buildings and the edges are connected between buildings from different clusters. The weight of each edge is determined by a combination of similarity scores from building matching and the physical distance. The goal is to select one reference building from each cluster, such that the total weight is maximized. The authors employ a replicator dynamics algorithm~\citep{DS1, DS2} to select a dominant set as shown in Figure~\ref{dominant_set}. The final prediction of the geo-localization is the average of each selected building's GPS in the dominant set. A diagram of their approach is shown in Figure~\ref{dominant_set}.\\

\noindent\textbf{Landmark matching:}~\citet{verde2020ground} proposed a graph matching method that assumed the location of the buildings, roads, and trees were pre-annotated on a map. The key idea was to construct co-visibility matrices from the locations of the pre-annotated landmarks to represent the adjacency from both satellite view and ground view. Candidate locations were extracted by identifying the regions in which the query and reference matrices overlapped. Then, a class adjacent matrix was built by calculating the edge type for each candidate matrix and query matrix. Finally, a Bayesian-based posterior maximum algorithm was adopted for selecting the closest matching satellite image for geo-localization. Compared with matching only buildings~\citep{tian2017cross}, landmark matching~\citep{verde2020ground} includes more classes of objects such as trees and roads. However, this paper assumed that the availability of pre-annotated landmarks which are not always available in real-world settings.

%\subsubsection{Contrastive Learning Approach}
\subsection{Deep Siamese-Like Methods}
With the development of CNNs~\citep{Alexnet, gradient} and siamese networks~\citep{siamese,siamese2, contrastiveLoss}, deep siamese-like networks have become mainstream in cross-view image geo-localization. Similar to hand-crafted feature representation methods, siamese-like deep learning methods frame the cross-view geo-localization problem as a retrieval task. Siamese networks traditionally contain two symmetric subnetworks which share the same weights in each layer. However, in cross-view geo-localization, applying the same feature extractor to both aerial and ground perspectives may not achieve good results~\citep {cvmnet}.% like methods discussed in Section~\ref{hand_crafted}.
Recent research has demonstrated that deep siamese-like methods which jointly train the two subnetworks with unique weights outperform the traditional siamese network in which the weights are shared. Thus, the model can learn a more domain-specific feature representation than the hand-crafted feature extractor and achieve more competitive results.
\newline\newline
\noindent\textbf{Where-CNN:} \citet{lin2015learning} proposed Where-CNN which is the first CNN model to address cross-view image geo-localization using a siamese network~\citep{siamese}. Unlike most of the methods described in this section which geo-localized the given street view images from aerial images, this paper proposed to use reference images from an oblique aerial view (bird's eye view). The key idea of this approach is that oblique aerial images share more common features with ground-level images than satellite images. To achieve this goal, the two branches of Where-CNN were modified from an AlexNet~\citep{Alexnet} which was pre-trained on the ImageNet~\citep{imagenet} and Places~\citep{placeCNN} datasets. Features extracted from the last fully connected layer were normalized to have a zero mean and unit standard deviation. A contrastive loss~\citep{contrastiveLoss} was applied for fine-tuning the model.
\newline\newline
\noindent\textbf{MCVPlaces:} \citet{CVUSA} proposed the MCVPlaces model which was the first Siamese-like CNN model that performed cross-view geo-localization on \textit{satellite} and ground-level images. MCVPlaces contained a multi-scale aerial feature extractor and a ground-level image feature extractor which was pre-trained on the Places~\citep{placeCNN} dataset. The authors proposed to only optimize the satellite feature extractor during the training phase and freeze the ground-level image feature extractor. A large-scale cross-view dataset, Crossview USA (CVUSA) was proposed and used to train the MCVPlaces model with a Euclidean distance loss on ground and satellite features. The CVUSA dataset was later refined in~\citep{zhai} and became one of the most popular datasets in the cross-view geo-localization field. \newline\newline

\begin{figure}
    \centering
    \includegraphics[width=0.47\textwidth]{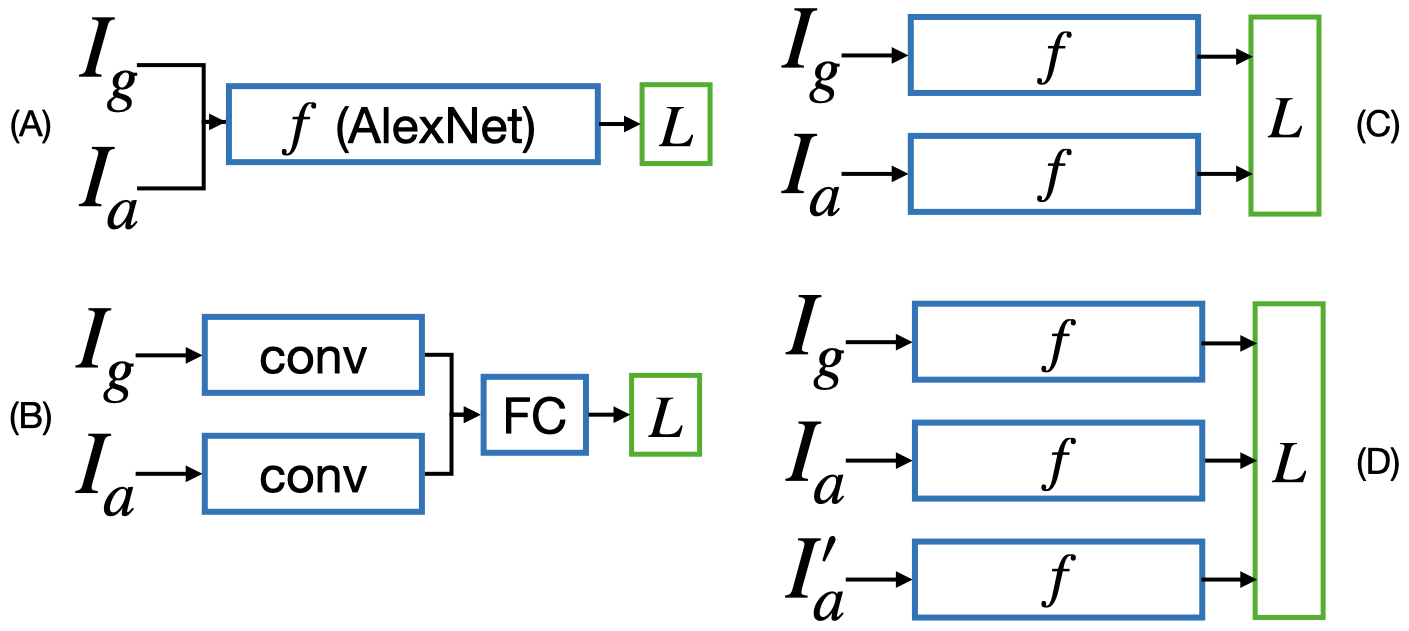}
    \caption{Four different structures are explored in~\citep{Vo}. \xh{$I_g$ represents the ground image. $I_a$ is the aerial image. $I^{\prime}_{a}$ is the negative aerial image from the triplet model. (a) indicates a classification model, (b) is a hybrid siamese-classification model, (c) is a siamese model, and (d) is a triplet model}}
    \label{Vo_4}
    % \end{mdframed}
\end{figure}

\begin{figure*}[!h]
    \centering
    \includegraphics[width=0.95\textwidth]{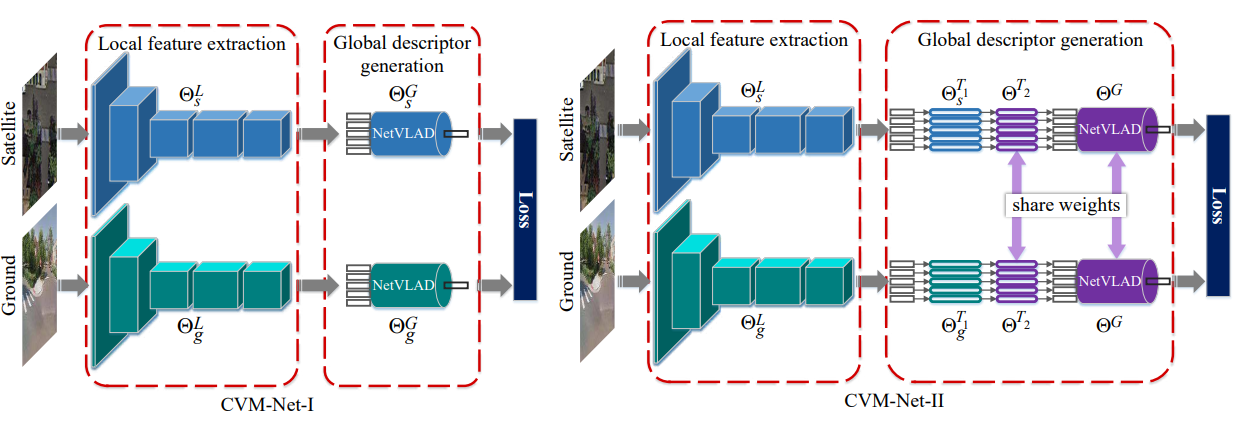}
    \caption{\xh{The proposed two CVM architecture in~\citep{cvmnet}. \textcolor{red}{CVM-Net-I and CVM-Net-II share the same backbone feature extractor ($\Theta^L_s$ and $\Theta^L_g$)} The critical difference between the two networks is the \textcolor{red}{architecture and the} weight sharing between the global descriptor generation module.} Figure is taken from~\citep{cvmnet}}
    \label{CVM_arch}
\end{figure*}

\noindent\textbf{DBL:} \citet{Vo} explored the effectiveness of four different architectures including classification architectures, hybrid Siamese-classification architectures, Siamese architectures, and triplet architectures, as shown in Figure~\ref{Vo_4}. A novel Distance-Based Logistic (DBL) loss layer was proposed for training the models. To evaluate the performance of the different architectures, the authors collected a large-scale dataset. Crucially, the goal of this paper was to localize the scenes depicted in the photo which is different compared to other methods in this section that geo-localize the location of the camera. The authors concluded that the triplet network achieved the best results on the testing set. Furthermore, an exhaustive mini-batch strategy was proposed in this paper to obtain maximum negative samples for a ground truth pair. In detail, assuming $M$ sampled locations in a mini-batch, each location has a ground-satellite pair. To construct triplet pairs for training, each query ground image has to have $1$ positive reference satellite image (ground truth) and $M-1$ negative reference satellite images. Similarly, each query satellite image has to have $1$ positive reference ground image (ground truth) and $M-1$ negative reference ground images. Thus, in this mini-batch, a total of $M \times 2(M-1)$ triplet pairs could be obtained. This exhaustive mini-batch strategy has become the standard operation for cross-view geo-localization methods trained with the triplet-based loss function.
\newline\newline
\noindent\textbf{CVM-Net:}
\citet{cvmnet} introduced NetVLAD~\citep{netvlad} into cross-view geo-localization. In detail, the authors combined a siamese network~\citep{siamese} with NetVLAD~\citep{netvlad} to jointly learn discriminative features from the ground and aerial views. The authors proposed two models named CVM-Net-I and CVM-Net-II in which \textcolor{red}{both share} the same backbone architecture for extracting features. CVM-Net-I optimized the parameters of the two NetVLAD layers (\textcolor{red}{the satellite} extractor and ground extractor) separately. However, the last convolutional layers and the NetVLAD layers of CVM-Net-II share the same parameters between the satellite extractor and ground extractor. To speed up the training and avoid manually choosing the margin, an improved weighted soft margin triplet loss was proposed. Because of the advanced architecture and the new loss function, CVM-Net achieved competitive results on \textcolor{red}{the} CVUSA~\citet{CVUSA} and Vo~\citep{Vo} datasets. Figure~\ref{CVM_arch} visualizes this approach. \\

\begin{figure}[!h]
% \begin{mdframed}[backgroundcolor=red!50,linecolor=red!50]
\centering
\includegraphics[width=\linewidth]{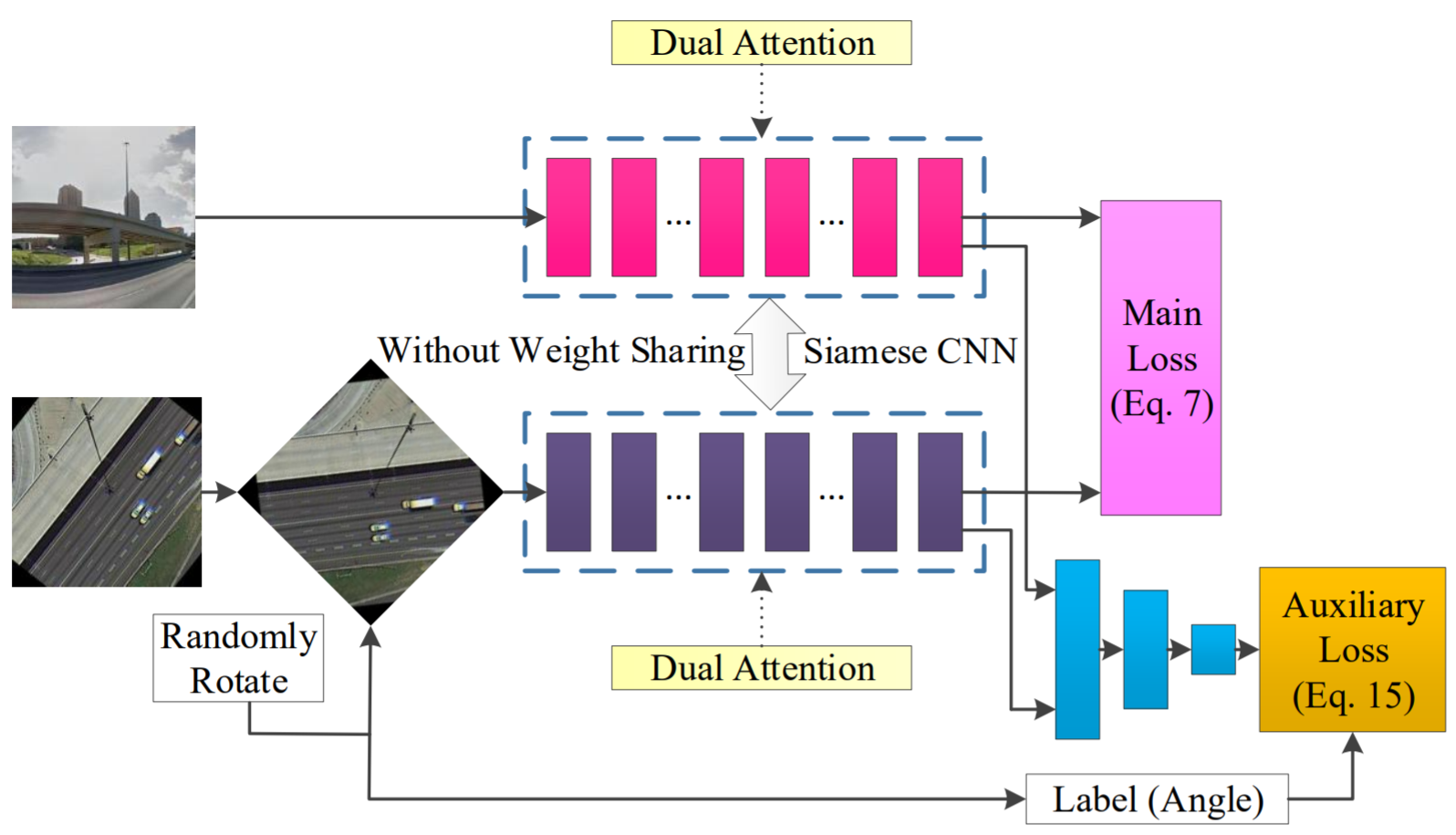}
\caption{The model proposed by~\citep{hardTriplet}. \xh{The model was composed of two dual attention networks for extracting features from aerial and ground perspectives respectively. An auxiliary subnetwork is adopted to predict the orientation of the randomly rotated input aerial image.} Figure is taken from~\citep{hardTriplet}.}
\label{fig:FCBAM}
% \end{mdframed}
\end{figure}

\noindent\textbf{FCBAM:} Deep siamese-based methods usually depend on metric learning where performance is largely affected by the hard samples in the training data. It is well-proven that hard samples in the training data decrease the training quality at an early stage~\citep{schroff2015facenet}. A similar pattern can also be observed in cross-view geo-localization methods which are trained with triplet loss. To alleviate this problem,~\citep{hardTriplet} proposed a Hard Exemplar Re-weighting (HER) triplet loss. The HER triplet loss is adopted from the soft margin triplet loss in~\citep{Vo}. The major difference is that the authors assign different weights for different triplets by a distance rectified logistic regression module. Another innovation in this paper is the Feature Context-Based Attention Module (FCBAM). FCBAM is inspired by the Convolutional Block Attention Module (CBAM)~\citep{cbam} and Contextual Reweighting Network (CRN)~\citep{feature_reweighting}. FCBAM is composed of a channel attention module that highlights the informative features for each channel, and a spatial attention module which captures the resulting features .%The output features are sequentially processed by these two modules. The channel attention module highlights the informative features for each channel, \textcolor{red}{which are captured by the spatial attention module.} 
 The overview of the proposed model is shown in Figure~\ref{fig:FCBAM}.\\

\begin{figure}[!h]
% \begin{mdframed}[backgroundcolor=red!50,linecolor=red!50]
\centering
\includegraphics[width=\linewidth]{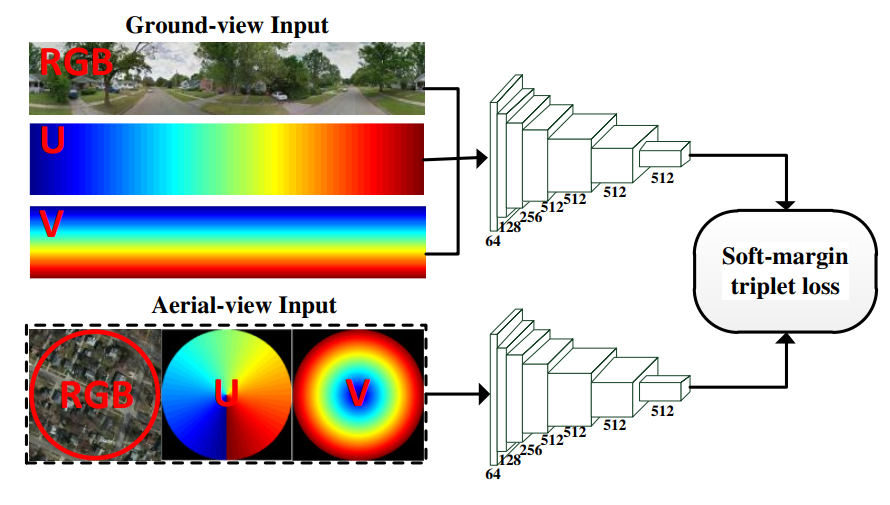}
\caption{Architecture of the OriCNN network proposed in~\citep{liu2019lending}. \xh{The $U$ channel represents the `azimuth' for both views. The $V$ channel contains the `altitude' for ground view images and the `range' for aerial view images. This figure is taken from~\citep{liu2019lending}.}}
\label{fig:oricnn}
% \end{mdframed}
\end{figure}

\noindent\textbf{OriCNN:} Orientation information has been found to be helpful for cross-view image geo-localization~\citep{Vo}. However, merging the orientation information with the CNN model is a challenging problem. \citet{liu2019lending} proposed an efficient method to fuse the heading direction of ground-level images into a network and learn more robust features. The authors borrowed ideas from the color-coded map and designed a similar mechanism to encode the pixel-wise orientation information. \xh{The author proposed to encode the azimuth of two views into the $U$ channel. The altitude of the ground view and the range of the aerial view is encoded in the $V$ channel.} A visualization is shown in Figure~\ref{fig:oricnn}. The siamese network takes a stack containing the $U$ channel and $V$ channel as input. A weighted soft-margin triplet loss~\citep{cvmnet} is adopted for training the model. To better evaluate the performance, the authors also proposed a large-scale dataset, CVACT, containing 10$x$ more samples than CVUSA in the testing set.
\newline\newline

\begin{figure}[t]
    \centering\includegraphics[width=0.47\textwidth]{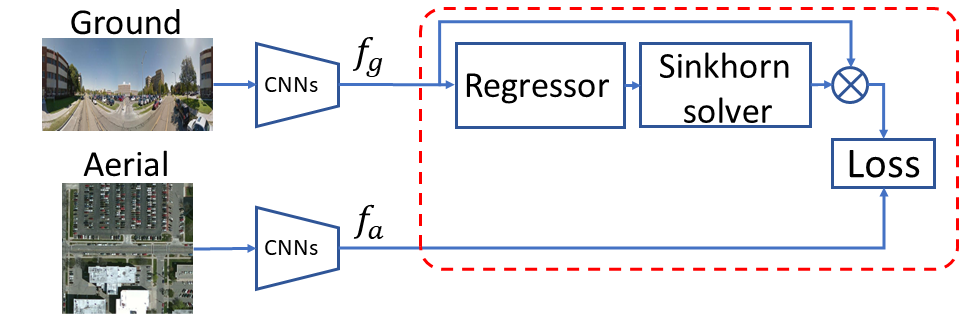}
    \caption{The proposed pipeline of CVFT~\citep{featureTransport}. For ground input images, the features extracted by the backbone network are projected to the aerial feature domain using a learned feature transportation module (denoted as a dashed red box) \textcolor{red}{which is composed of a regressor and a Sinkhorn solver. A soft-margin triplet loss is adopted to train the model.}}
    \label{fig_CVFT}
% \end{mdframed}
\end{figure}

\noindent\textbf{CVFT:} Inspired by Optimal Transport (OT) theory, \citet{featureTransport} proposed a Cross-View Feature Transport (CVFT) layer for domain feature transferring which facilitates cross-view feature matching. CVFT explicitly models the domain gap between \textcolor{red}{aerial-view imagery and ground-view} images by a transport matrix. Traditionally, OT is a linear programming problem that is computationally inefficient. \textcolor{red}{To solve} this problem, a different version~\citep{cuturi2013sinkhorn} of OT, called \textcolor{red}{a} Sinkhorn solver~\citep{sinkhorn1967concerning,knight2008sinkhorn}, which is based on entropy regularization was adopted. Finally, the feature transport problem was converted into a convex problem to be solved by a Sinkhorn solver. A diagram of their approach is provided in Figure~\ref{fig_CVFT}.\\

% \begin{figure*}
%     \centering
%     \begin{subfigure}[b]{0.49\textwidth}
%     \centering
%     \includegraphics[width=\textwidth]{cross-fig-8-FCBAM.PNG}
%     \caption{The proposed model in~\citep{hardTriplet}. Figure is taken from~\citep{hardTriplet}.}
%     \label{fig:FCBAM}
%     \end{subfigure}
%     \hfill
%     \begin{subfigure}[b]{0.49\textwidth}
%         \centering
%     \includegraphics[width=\textwidth]{cross-fig-6-OriCNN.PNG}
%     \caption{Architecture of OriCNN network proposed in~\citep{liu2019lending}.}
%     \label{oricnn}
%     \end{subfigure}
%     \hfill
%     \begin{subfigure}[b]{0.85\textwidth}
%     \centering
%     \includegraphics[width=\textwidth]{cross-fig-5-CVM.PNG}
%     \caption{CVM-Net architecture. Figure is taken from~\citep{cvmnet}}
%     \label{CVM_arch}
%     \end{subfigure}
%     \hfill
%     \begin{subfigure}[b]{0.85\textwidth}
%         \centering
%     \includegraphics[width=\textwidth]{cross-fig-7-Vigor.PNG}
%     \caption{Architecture of the method proposed in~\citep{ChenChen}. This architecture is the same as~\citep{Hongdong} except for the additional offset prediction branch.}
%     \label{vigor}
%     \end{subfigure}
%     \caption{(a) is taken from~\citep{liu2019lending} and (a) is taken from~\citep{ChenChen}}
% \end{figure*}

\noindent\textbf{SAFA:} The main challenge of cross-view geo-localization is the domain gap between the aerial and ground views. By leveraging the prior geometric knowledge between the two domains,~\citet{Hongdong} proposed a pre-processing technique that adopts a polar transformation on the \textcolor{red}{aerial-view} image. This technique bridges the visual domain gap between the \textcolor{red}{ground-view} images and the aerial images. The authors also proposed a spatial-aware feature aggregation network (SAFA) to capture global features from both aerial and ground images. Inspired by~\citep{spatial_pyramid}, the authors stacked multiple SAFAs in parallel to simultaneously aggregate features from different aspects. The proposed SAFA module and polar transformation largely improved the performance on the CVUSA~\citep{CVUSA} and CVACT~\citep{liu2019lending} datasets. However, the polar transformation proposed in this paper assumes the locations of the ground view images always align at the center of the aerial image, which is not always the case in real-world scenarios.\\
\\

\begin{figure*}[!h]
    \centering
    \includegraphics[width=0.85\textwidth]{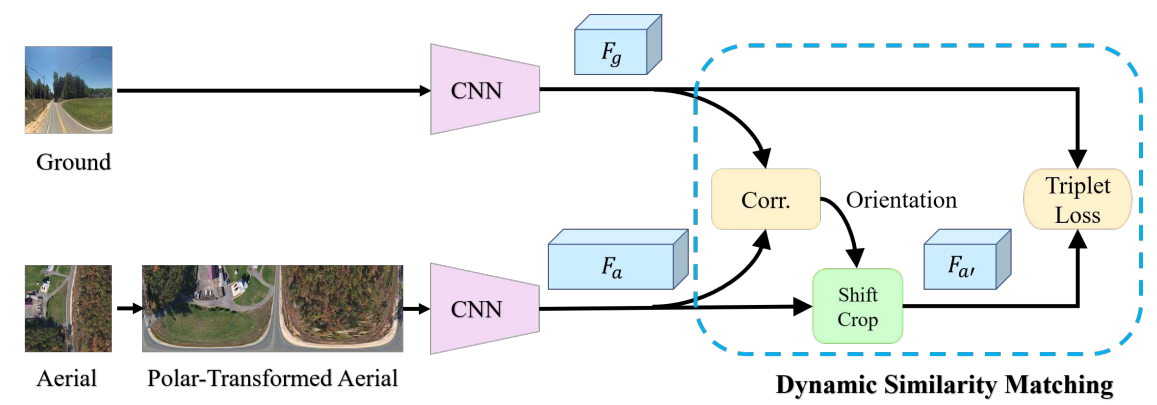}
    \caption{The proposed architecture for the Dynamic Similarity Matching (DSM)~\citep{DSM} network. \xh{The DSM module estimates the orientation of the ground image by sliding and matching between the feature maps of ground images and polar transformed aerial images.} Figure is taken from~\citep{DSM}}
    \label{fig:DSM}
\end{figure*}

\begin{figure*}[!h]
% \begin{mdframed}[backgroundcolor=red!50,linecolor=red!50]
    \centering
    \includegraphics[width=0.95\textwidth]{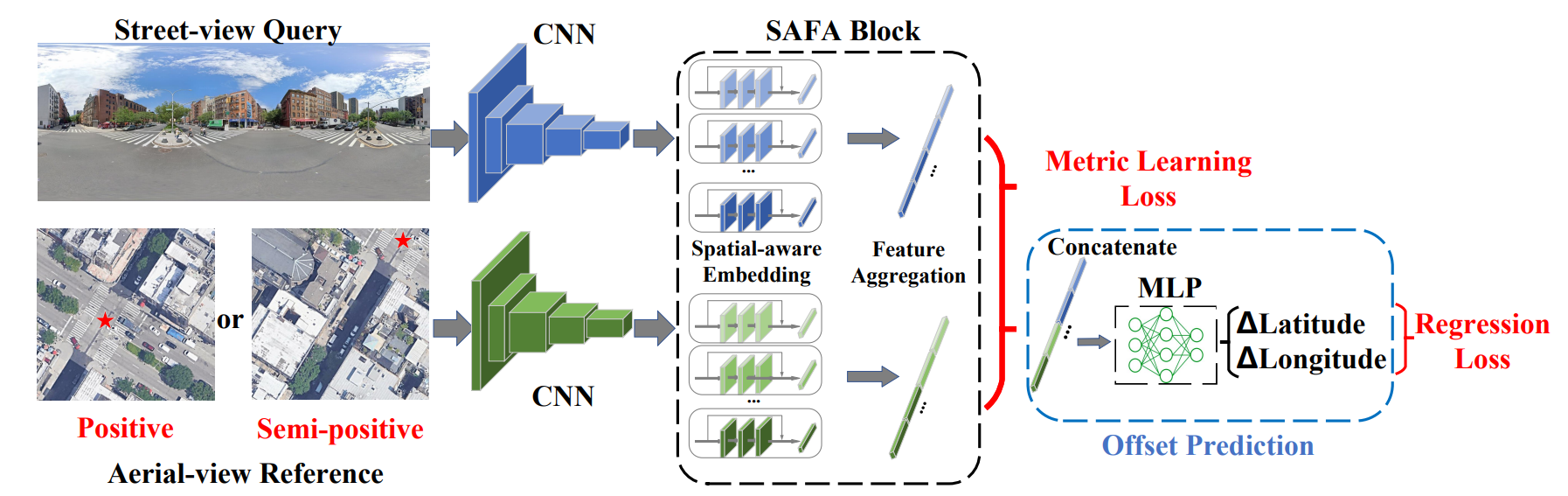}
    \caption{Architecture of the method proposed by~\citep{ChenChen}. \xh{The SAFA blocks aggregate the raw features extracted by backbone CNNs. A subnetwork predicts the offset latitude and longitude to determine the exact location of the ground image by taking the predicted features as input.} This architecture is the same as~\citep{Hongdong} except for the additional offset prediction branch.}
    \label{vigor}
% \end{mdframed}
\end{figure*}

\noindent\textbf{DSM:}
Most cross-view geolocalization methods~\citep{CVUSA, liu2019lending,cvmnet} assume that the query images are panoramic ground-level images.~\citep{DSM} focused on cross-view geo-localization on images with their heading orientation but a limited field of view. The authors proposed a Dynamic Similarity Matching (DSM) module to \textcolor{red}{both measure the feature similarity between two views and their orientation using a sliding window.} The proposed model is presented in Figure~\ref{fig:DSM}. Specifically, DSM takes the ground feature and the aerial feature extractors as input. Then a shifting window slides across the ground features to compute the inner product with the aerial features. The location of the highest inner product value indicated the orientation and similarity. Note that the polar transformation is one of the pre-processing steps in this paper.
\newline\newline
\noindent\textbf{VIGOR:} Previous existing datasets assume that there is a one-to-one correspondence between \textcolor{red}{ground-view images and aerial-view} images. To address this limitation, \citet{ChenChen} proposed a new dataset that has a many-to-one correspondence between the ground and aerial images. This dataset is densely sampled in four USA cities and grabbed panoramic images from Google Street View (GSV)~\citep{GSV}. The authors proposed three categories to define cross-view matching: positive, semi-positive, and negative. Positive and negative pairs are the same as in CVUSA and CVACT. In the semi-positive pairs, the ground-view location appears in \textcolor{red}{aerial-view} images but not in the center region. By doing so, cross-view geo-localization becomes a many-to-one retrieval problem rather than a one-to-one retrieval problem. This configuration is more realistic and closer to real-life deployment. Besides the many-to-one retrieval formation, VIGOR also achieved meter-level offset prediction of the camera location by utilizing an offset prediction subnetwork. Their experiments demonstrated that the noisy GPS coordinates can be refined using the offset prediction subnetwork. The proposed model in this paper adopted the same architecture as SAFA~\citep{Hongdong} with the proposed offset prediction subnetwork as shown in Figure~\ref{vigor}. The authors also proposed an IOU-based loss function to guide the network in learning features from semi-positive samples.

\noindent\textbf{AlignNet:}
\citet{revisiting} studied the effect of orientation alignment between ground images and aerial images. The authors found that some existing algorithms are unfair due to aligning the orientation between two views in both training and testing as prior. The authors proposed a novel method for alignment angle regression using the output from Grad-CAM~\citep{gradcam}. Moreover, a new loss function that can balance the gradient of positive pairs and negative pairs was proposed. Finally, the authors proposed a new global mining strategy for training the model with hard samples from full datasets rather than inside a mini-batch as was typical in previous works.
\newline\newline

% \begin{figure*}[!h]
%     \centering
%     \includegraphics[width=0.85\textwidth]{cross-fig-9-seeing.PNG}
%     \caption{The augmentation pipeline proposed in~\citep{rodrigues2021these}. Figure is taken from~\citep{rodrigues2021these}.}
%     \label{fig:seeing}
% \end{figure*}

\begin{figure*}
    \centering
    \includegraphics[width=0.99\textwidth]{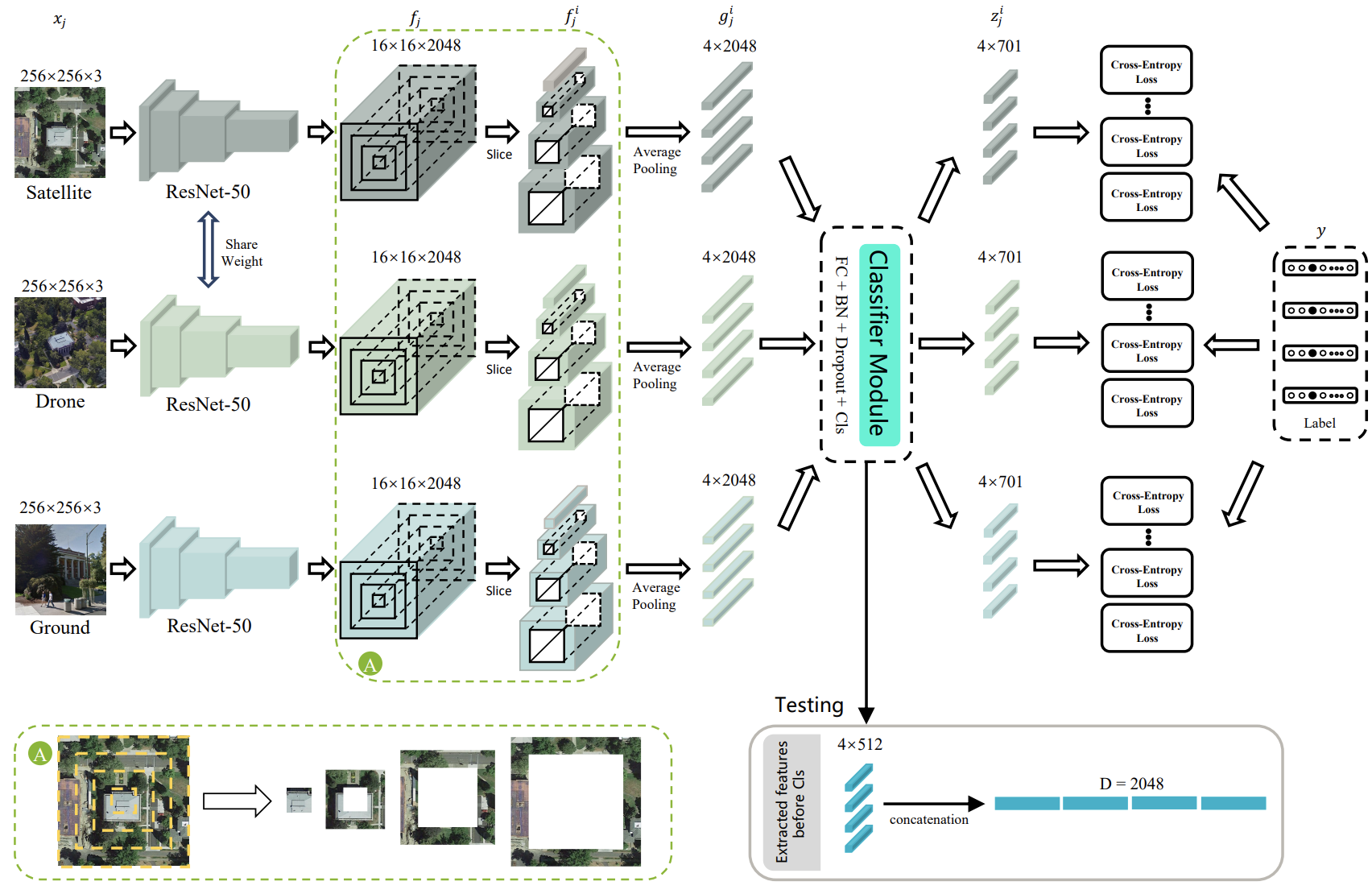}
    \caption{An overview of the Local Pattern Network (LPN) proposed by~\citep{wang2021each}. The proposed \textcolor{red}{square-ring} partitioning strategy generates latent features from the input images \textcolor{red}{at different scales of neighboring areas.} The LPN network utilizes a classification module to learn a latent space representation. Figure is taken from~\citep{wang2021each}}
    \label{fig:LPN}
\end{figure*}

%Are These from the Same Place?
%Seeing the Unseen in Cross-View Image %Geo-Localization~\citep{rodrigues2021these}\\
\noindent\textbf{Seeing the Unseen:}~\citet{rodrigues2021these} proposed a novel data augmentation pipeline for cross-view geo-localization. The authors utilized the segmentation map from existing models to cut out the objects (buildings, sidewalks, sky, etc.) in the \textcolor{red}{ground-view} images to force the network to learn from the unseen objects and perform unseen object matching. Apart from the novel data augmentation pipeline, a multi-scale attention module was proposed for the matching task.\\
\\
\noindent\textbf{LPN:}~\citet{wang2021each} proposed a Local Pattern Network (LPN) which is an end-to-end learnable model able to extract features globally from both \textcolor{red}{aerial and ground-view} images. Fig~\ref{fig:LPN} shows the structure of the LPN. The key idea of the LPN is to force the network to focus on the contextual information in the neighboring areas using the square-ring partitioning strategy shown in the green box of Fig~\ref{fig:LPN}. To be noticed, unlike any methods introduced in this section, LPN uses a multi-class cross-entropy loss for training. During the testing, features before the classification layer are used for estimating the similarity by cosine distance. This approach is visualized in Figure~\ref{fig:LPN}.\\

\begin{figure}
    \centering
    \includegraphics[width=0.5\textwidth]{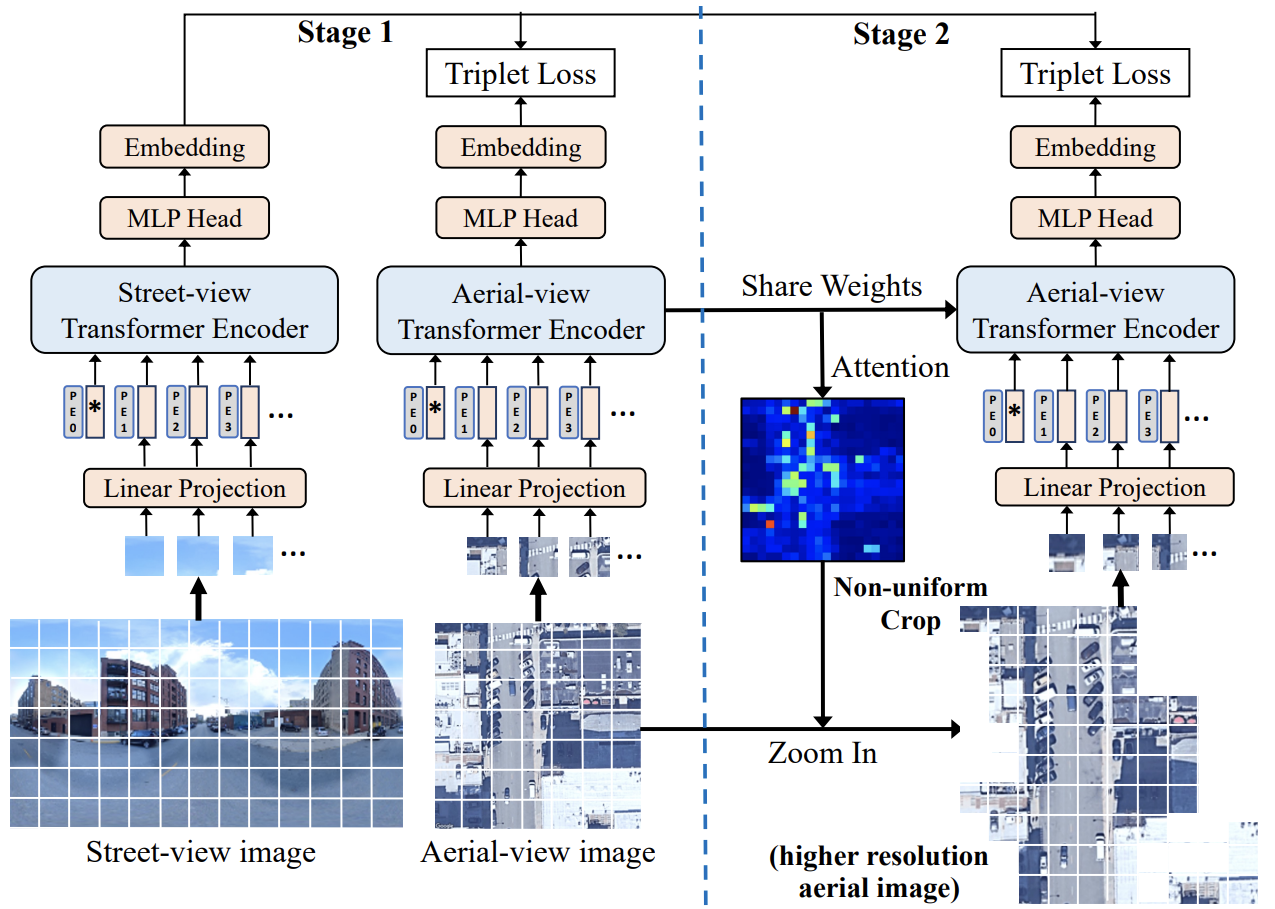}
    \caption{\textcolor{red}{An overview of the TransGeo~\citep{TransGeo} model. The model contains a street-view transformer encoder and an aerial-view transformer encoder. In the first stage, the model is trained by a soft-margin triplet loss. In the second stage, the model crops the important regions from the aerial view for fine-grained feature extraction. Figure is taken from~\citep{TransGeo}.}}
    \label{fig:trans_geo}
\end{figure}

\textcolor{red}{\noindent\textbf{TransGeo:} Transformers and multi-head attention mechanisms~\citep{transformer} have been developing rapidly in recent years. The capability of the transformer to explore the global correlations significantly boosts the performance of cross-view geo-localization research.~\citet{TransGeo} is one of the pioneers to apply transformers to this field. Their proposed TransGeo model is presented in Figure~\ref{fig:trans_geo}. TransGeo is composed of two main parts, namely a street-view transformer encoder and an aerial-view transformer encoder. Both transformer encoders are employed from the pre-trained DeiT~\citep{DeiT} model. A special two-stage training paradigm is proposed to train TransGeo. In the first stage, the model is trained by the normal soft-margin triplet loss. In the second stage, based on the saliency attention map from the aerial-view transformer encoder, non-uniform cropping is applied to the aerial image to extract more fine-grained features. Benefiting from the advanced transformer architecture and two-stage training paradigm, TransGeo~\citep{TransGeo} is considered one of the best models in cross-view geo-localization and achieves superior performance on popular benchmarks.}\\

\begin{figure}
    \centering
    \includegraphics[width=0.5\textwidth]{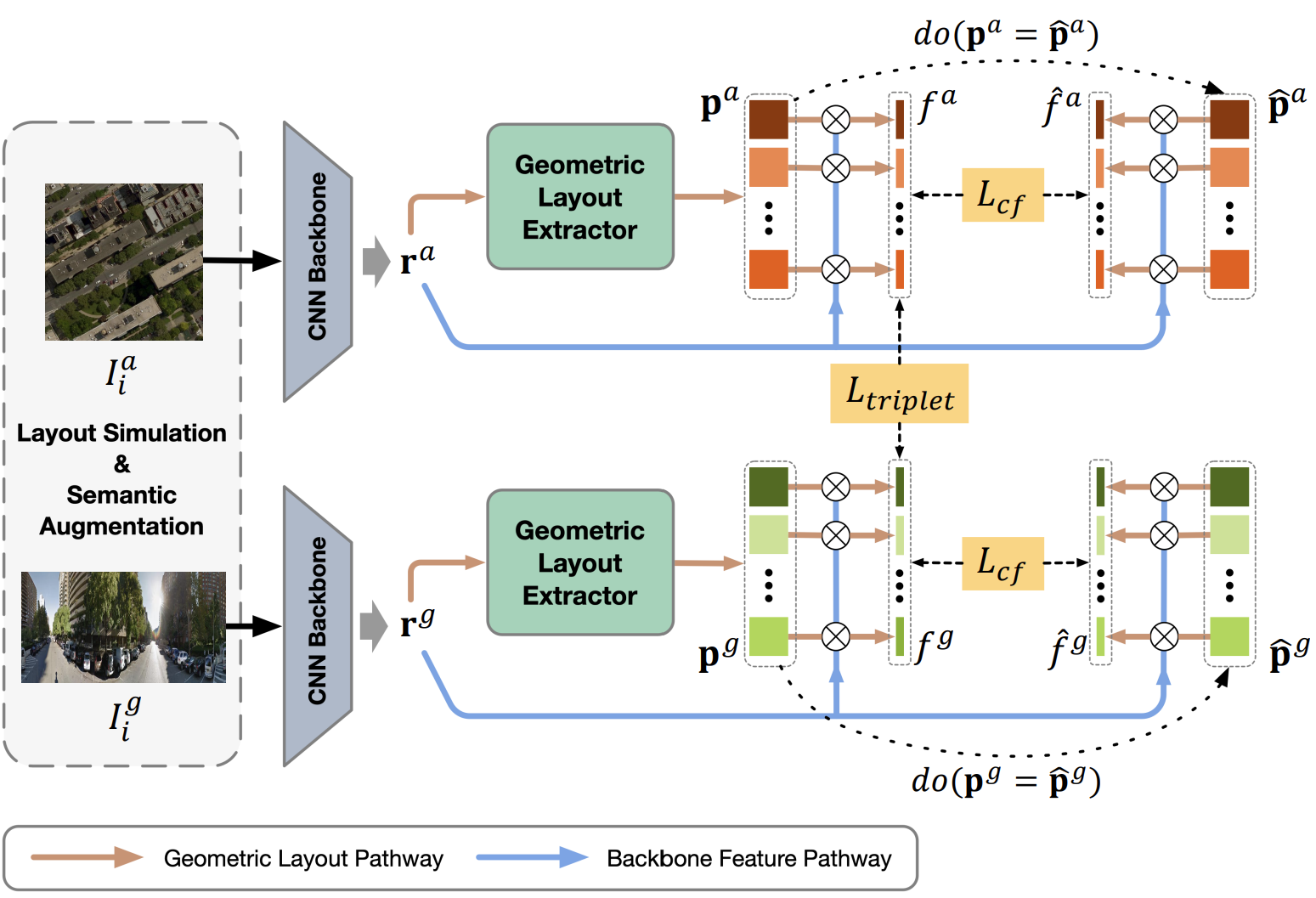}
    \caption{\textcolor{red}{An overview of the GeoDTR~\citep{geodtr} model. Each aerial and ground image pair is first augmented by the proposed layout simulation and semantic augmentation. The geometric layout extractor then extracts layout information from the raw features. The $do$ operation stands for the counterfactual learning schema. The predicted features are the Frobenius product between the layout features and the raw features. Figure is taken from~\citep{geodtr}.}}
    \label{fig:geo_dtr}
\end{figure}

\textcolor{red}{\noindent\textbf{GeoDTR:} Most cross-view geo-localization methods aim to implicitly learn the geometric layout similarity by matching the learned latent features. GeoDTR~\citep{geodtr} addresses this issue by explicitly capturing the geometric layout information from both aerial and ground images via a trainable geometric layout extractor which consists of 2 transformers, as presented in Figure~\ref{fig:geo_dtr}. To tackle the issue of lacking ground truths for the geometric layouts, a counterfactual learning scheme is proposed to provide weak supervision signals to train the model. To further alleviate the issue of overfitting to low-level details, the authors proposed layout simulation and semantic augmentation (LS) to augment training data. Unlike existing data augmentation methods, the authors argue LS does not break the correspondences between aerial and ground image pairs. LS maintains the correspondences while diversifying the geometric layout and low-level details of the training data.}

\subsection{Generative Methods}
Since the proposal of Generative Adversarial Networks (GAN)~\citep{gan} in 2014, image \textcolor{red}{synthesis} has increasingly been dominated by GANs. Due to \textcolor{red}{their} ability to generate realistic images, GANs have been applied in diverse applications such as image synthesis~\citep{bigGAN, stylegan, gan}, image-to-image translation~\citep{cycleGAN,dualgan,attentiongan, patchgan,ganimation, bicycle}, and super resolution~\citep{photoGAN,esrgan}. To this end, researchers also explored the relationship between satellite and ground images. For example, given an \textcolor{red}{aerial-view} image, a model learns to generate a \textcolor{red}{ground-view} image that keeps the visual features from the aerial image (building, road, tree, etc). \citet{zhai} first discovered that \textcolor{red}{ground-view} scene layouts can be generated from a segmentation map of an aerial image. \citet{crossCGAN} proposed two models named as X-Fork and X-Seq. These models generate street-view images and corresponding segmentation maps using only \textcolor{red}{aerial-view} images as input by leveraging a conditional GAN~\citep{mirza2014conditional}. \citet{tang2019multi} proposed SelectionGAN which utilized a multi-channel selection module to obtain better quality for the final generated \textcolor{red}{ground-view images.} More recently, predicting satellite depths and further taking advantage of geometric transformations to estimate \textcolor{red}{ground-view} panoramic images is becoming a new trend. \citet{shi2021geometry} proposed a satellite-to-street-view image projection (S2SP) module to achieve this goal. Whereas,~\citet{geometrySynthesis} achieved this by employing a panoramic projection via generating an occupancy grid from depth maps and the first encountering voxels in the generated occupancy grid. After geometric transformation, the result is fed into a generator which produces the final street-view image. For example, BiCycleGAN~\citep{bicycle} is employed after the panoramic projection in~\citep{geometrySynthesis}. Although the methods mentioned above are not cross-view geo-localization methods, they have inspired the following cross-view geo-localization methods which use GANs to boost the performance of cross-view geo-localization methods.
\newline \newline

\begin{figure*}
    \centering
    \includegraphics[width=0.85\textwidth]{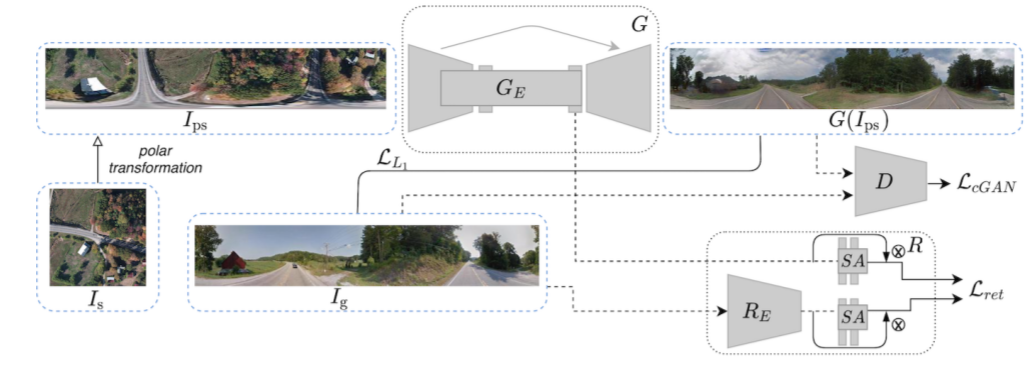}
    \caption{\textcolor{red}{An overview of SAFA-GAN proposed by~\citep{synthesis}. A GAN \textcolor{red}{$G$ and $D$} is adopted to perform ground image synthesis by taking the polar transformed aerial image as input. Then, a SAFA module ($SA$) is employed to train the model which performs cross-view geo-localization. \textcolor{red}{$R_E$ is a learnable submodule to map the ground image to the learned latent feature space.} This figure is taken from~\citep{synthesis}.}}
    \label{fig:safagan}
\end{figure*}

\noindent\textbf{SAFA-GAN:} Siamese-like CNN models easily neglect low-level detail because the triplet loss and its variants do not have constraints on them. However, low-level details are useful \textcolor{red}{when incorporated into the} cross-view geo-localization task. To accomplish this, \citet{synthesis} proposed to learn discriminative features between satellite and street images using a Generative Adversarial Network~\citep{gan}. The proposed network is composed of a GAN~\citep{gan} that maps a polar transformed \textcolor{red}{aerial-view} image to a synthetic street-view image, and a SAFA-based~\citep{Hongdong} \textcolor{red}{subnetwork} to perform cross-view geolocalization task. The proposed model is shown in Figure~\ref{fig:safagan}. The intermediate features from the generator are then reused in the retrieval \textcolor{red}{subnetwork} for cross-view geo-localization. In this manner, the two tasks, retrieval and generation, mutually learn from each other to create a feature representation informative for both tasks.\newline\newline

\begin{figure}
    \includegraphics[width=0.47\textwidth]{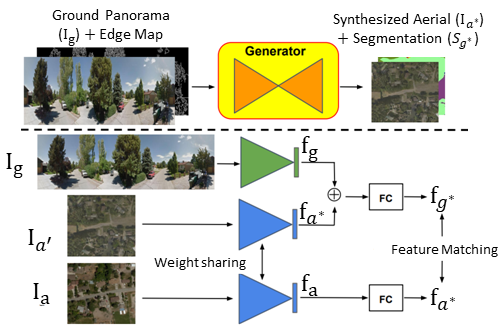}
    \caption{The method proposed in~\citep{bridging}. The upper panel \textcolor{red}{is the X-Fork~\citep{crossCGAN} generator which generates satellite images from ground images. The bottom panel illustrates the proposed feature fusion schema. The blue module extracts features from real satellite images $I_a$ and generated satellite images $I_{a'}$. The ground features $I_g$ are extracted from the green module. $I_g$ and $I_{a'}$ are fused by a fully connected layer. The fusion algorithm aims to push the fused features $f_{g^*}$ and aerial features $f_{a^*}$ closer in the latent space.} The figure is taken from~\citep{bridging}.}
    \label{fig:bridge}
\end{figure}
\noindent\textbf{Feature Fusion GAN:} ~\citet{bridging} proposed a novel feature fusion training strategy in which the features from synthesized satellite imagery are fused with corresponding \textcolor{red}{street-view} features. An overview of the proposed method is shown in Figure~\ref{fig:bridge}. To synthesize a satellite view from a street view, the authors adopted the generator from X-Fork~\citep{crossCGAN}. Two CNNs~\citep{Alexnet} are used to extract satellite features and street view features independently. After fusing the street view features and the generated satellite view features with a fully connected network, feature matching is performed between the output of the fully connected network and the ground truth satellite. This model pushed the performance of cross-view geo-localization by a large margin on the CVUSA~\citep{CVUSA} dataset.
\\

\begin{table*}[t]
    % \begin{mdframed}[backgroundcolor=red!50,linecolor=red!50]
    \footnotesize
    \centering
    \begin{tabular}{ccccc} \toprule \toprule
    Method & Type & Deep Learning vs. Traditional & Polar Transformation & Year  \\ \midrule
    AVG \& DT~\citep{lin2013cross} & Hand-Crafted Feature & Traditional & No & 2013 \\
    Semantic cross-view matching~\citep{castaldo2015semantic} & Hand-Crafted Feature & Traditional & No & 2015 \\
    Building matching~\citep{tian2017cross} & Graph-Based & Deep Learning$+$Dominant Set & No & 2017 \\
    Landmark matching~\citep{verde2020ground} & Graph-Based & Traditional & No & 2020 \\
    Where-CNN~\citep{lin2015learning} & Siamese-Like Network & Deep Learning & No & 2015 \\
    MCVPlaces~\citep{CVUSA} & Siamese-Like Network & Deep Learning & No & 2015 \\
    DBL~\citep{Vo} & Siamese-Like Network & Deep Learning & No & 2016 \\
    CVM-Net~\citep{cvmnet} & Siamese-Like Network & Deep Learning & No & 2018 \\
    FCBAM~\citep{hardTriplet} & Siamese-Like Network & Deep Learning & No & 2019 \\
    OriCNN~\citep{liu2019lending} & Siamese-Like Network & Deep Learning & No & 2019 \\

    Feature Fusion GAN~\citep{bridging} & Generative & Deep Learning & No & 2019 \\
    CVFT~\citep{featureTransport} & Siamese-Like Network & Deep Learning & No & 2020 \\
    SAFA~\citep{Hongdong} & Siamese-Like Network & Deep Learning & Yes & 2020 \\
    DSM~\citep{DSM} & Siamese-Like Network & Deep Learning & Yes & 2020 \\
    VIGOR~\citep{ChenChen} & Siamese-Like Network & Deep Learning & No & 2021 \\
    AlignNet~\citep{revisiting} & Siamese-Like Network & Deep Learning & No & 2021 \\
    Seeing the Unseen~\citep{rodrigues2021these} & Siamese-Like Network & Deep Learning & No & 2021 \\
    LPN~\citep{wang2021each} & Siamese-Like Network & Deep Learning & No & 2021 \\
    SAFA-GAN~\citep{synthesis} & Generative & Deep Learning & Yes & 2021 \\
    \textcolor{red}{TransGeo}~\citep{TransGeo} & Siamese-Like Network & Deep Learning & No & 2022 \\
    \textcolor{red}{GeoDTR}~\citep{geodtr} & Siamese-Like Network & Deep Learning & Yes & 2023 \\
    \end{tabular}
    \caption{Summary of cross-view geo-localization methods.}
    \label{tab:summary_crv_models}
    % \end{mdframed}
\end{table*}

% \textbf{Without retrieval}\\
% Cross-view image synthesis using conditional gans \citep{crossCGAN} \\
% Geometry-aware satellite-to-ground image synthesis for urban areas \citep{geometrySynthesis} \\
% Predicting \textcolor{red}{ground-view} scene layout from aerial imagery \citep{zhai} \\
% Multi-channel attention selection gan with cascaded semantic guidance for cross-view image translation\citep{tang2019multi}\\
% Local class-specific and global image-level generative adversarial networks for semantic-guided scene generation\citep{tang2020local}\\
% What is it like down there? Generating dense \textcolor{red}{ground-view} views and image features from overhead imagery using conditional generative adversarial networks\citep{deng2018like}\\
% Geometry-guided street-view panorama synthesis from satellite imagery\citep{shi2021geometry}\\

\subsection{Summary of cross-view geo-localization methods}
Cross-view geo-localization is a difficult task due to the extreme differences in views, varying photo-taking time, and inconsistent resolution between two views. Before the deep learning era, most works struggled with the accuracy of models. With the advancement of deep learning, the performance of cross-view geo-localization methods has increased drastically. However, deep learning models are analogous to a black box. The lack of explainability hinders the development of deep-learning-based cross-view geo-localization methods. Graph-based methods maintain better explainability but are either not scalable to large-scale datasets~\citep{verde2020ground} or only work on areas with highly distinguishable objects (i.e. skyscrapers)~\citep{tian2017cross}. Despite lacking explainability, several polar-transformation-based methods~\citep{Hongdong,synthesis,DSM,geodtr} have achieved state-of-the-art performance. Polar transformations, as a pre-processing technique, assume the camera location lies at the center of an \textcolor{red}{aerial-view} image. This strong requirement can rarely be achieved during deployment in real-world scenarios. More recent works such as VIGOR~\citep{ChenChen} address this problem by proposing to estimate the shift in the camera's location. However, the performance still has significant room for improvement. A brief summary of all the introduced cross-view geo-localization methods is presented in Table~\ref{tab:summary_crv_models}.

With the rapid development of cross-view image geo-localization, it is a natural idea to extend cross-view image geo-localization to cross-view video geo-localization. To tackle this problem, some existing methods studied different categories of temporal information aggregation methods. For example, SeqGeo~\citep{seqgeo} and CVLNet~\citep{cvlnet} proposed to use transformers to aggregate the temporal features from ground images. On the hand, \textcolor{red}{GAMA~\citep{gama}} utilized a 3D CNN to gather the ground sequential information but adopted a transformer to extract spatial information from an aerial view. Cross-view video geo-localization is still a new research area and is out of scope in this section. Thus, we will not benchmark the above-mentioned cross-view video geo-localization methods in the following sections.
\subsection{Cross-View Image Geo-localization Datasets}
\label{sec:cross_dataset}
\begin{figure}
% \begin{mdframed}[backgroundcolor=red!50,linecolor=red!50]
    \centering
    \includegraphics[width=0.47\textwidth]{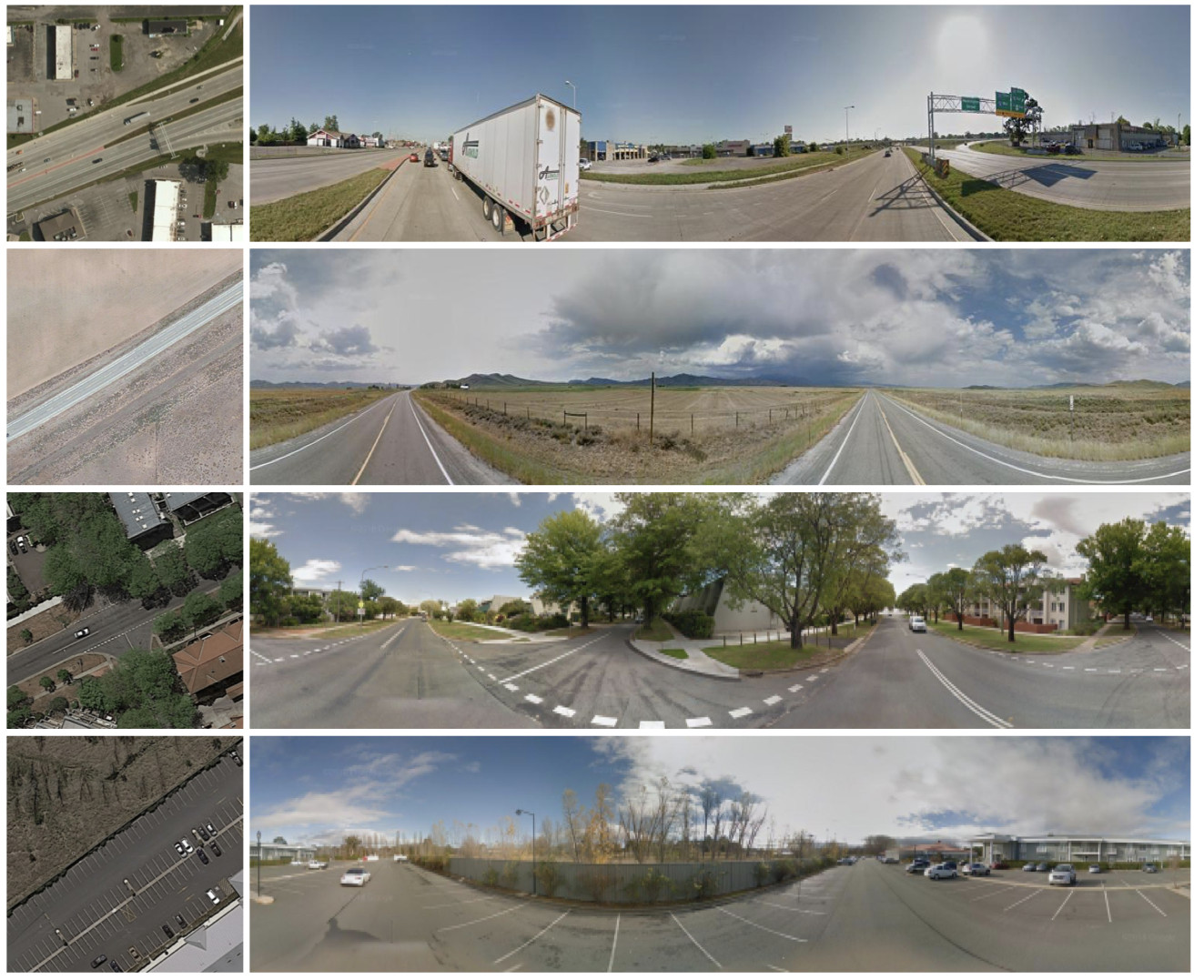}
    \caption{The top two rows display sample images from the CVUSA~\citep{CVUSA} dataset and the bottom two rows display sample images from the CVACT~\citep{liu2019lending} dataset. On the left is a satellite image and on the right side is its corresponding ground-level panorama.}
    \label{fig:sample_CVUSA_CVACT}
% \end{mdframed}
\end{figure}
\noindent\textbf{CVUSA:\footnote{http://mvrl.cs.uky.edu/datasets/cvusa/}} CVUSA~\citep{CVUSA} is the first large-scale cross-view geo-localization dataset containing street-view images downloaded from GSV~\citep{GSV} and Flickr\footnote{\label{foot:flickr}https://www.flickr.com/}. The satellite images were obtained from Bing Maps\footnote{\label{foot:bing}https://www.bing.com/maps/}. Specifically, GSV images were gathered from random locations in the USA, and the Flickr images were sampled and downloaded from the grid of 100 $\times$ 100 locations on a map of the USA. Finally, given the location of each ground image, corresponding satellite images with a resolution of $800\times800$ were downloaded from Bing Maps. In total, this dataset contains $879,318$ unique locations and a total of $1,588,655$ ground-satellite pairs. The well-known CVUSA benchmark is a refinement of the version from \citep{zhai}. It includes $35,532$ ground-satellite pairs for training and $8,884$ ground-satellite pairs for evaluation. Two sample images from the refined version are presented on the top two rows of Figure~\ref{fig:sample_CVUSA_CVACT}. In the rest of this paper, we refer `CVUSA' as the refined version unless otherwise specified.
\newline\newline

\noindent\textbf{CVACT:\footnote{https://github.com/Liumouliu/OriCNN}} CVACT~\citep{liu2019lending} covers 300 square miles of road in Canberra, Australia. To collect street and satellite images, the GSV API\footnote{\label{foot:google_street_api}https://developers.google.com/maps/documentation/streetview/overview} and Google Maps API was employed. All street view images (panoramas) were captured at zoom level 2 at a resolution of $1664 \times 832$ and satellite \textcolor{red}{images were} captured at zoom level 20 at a resolution of $1200 \times 1200$. Two sample images in this dataset are presented in the bottom two rows of Figure~\ref{fig:sample_CVUSA_CVACT}. In total, this dataset contains $128,334$ ground-satellite image pairs of which $35,532$ pairs are used for training, $8,884$ for validation, denoted as CVACT\_val, and $92,802$ for testing, denoted as CVACT\_test.
\newline\newline

\begin{figure}
    \centering
    \includegraphics[width=0.47\textwidth]{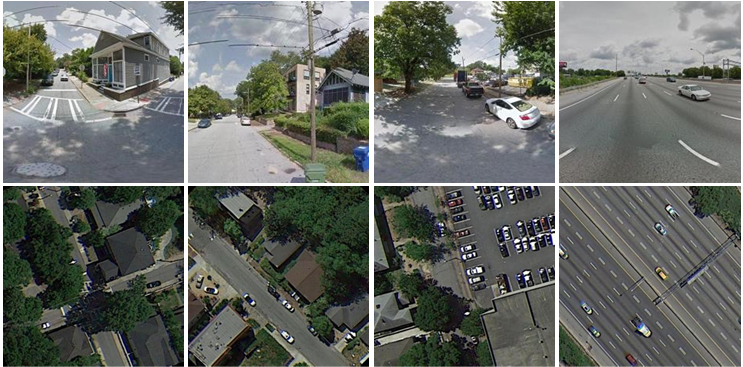}
    \caption{Four sample images from Vo~\citep{Vo} dataset. \xh{The top row shows the ground images and the bottom row shows the corresponding aerial images.}}
    \label{fig:sample_Vo}
\end{figure}

\noindent\textbf{Vo:\footnote{https://github.com/lugiavn/gt-crossview}} Vo~\citep{Vo} is a large-scale dataset containing more than 1 million ground-aerial image pairs. Four sample images are shown in Figure~\ref{fig:sample_Vo}. Different from other datasets mentioned in this section, this dataset focuses on localizing the scenes in the image (assuming the main object lies at the center of the satellite image) rather than the location of the camera. For example, in the first column of Figure~\ref{fig:sample_Vo}, the center area of the satellite image is the building which is the main object in the ground image. This is in contrast to the CVUSA~\citep{CVUSA} and CVACT~\citep{liu2019lending} datasets. To achieve this goal, the authors queried street-view panoramic images from GSV~\citep{GSV} and split them into several crops. For each crop, they obtained the depth estimation from GSV and downloaded the corresponding satellite image \textcolor{red}{using the} Google Maps API\footnote{\label{foot:google_map_api}https://developers.google.com/maps/documentation/maps-static/overview}. The ground panoramas were randomly collected from 11 different US cities by employing GSV~\citep{GSV}.
\newline\newline

\begin{figure}
    \centering
    \includegraphics[width=0.47\textwidth]{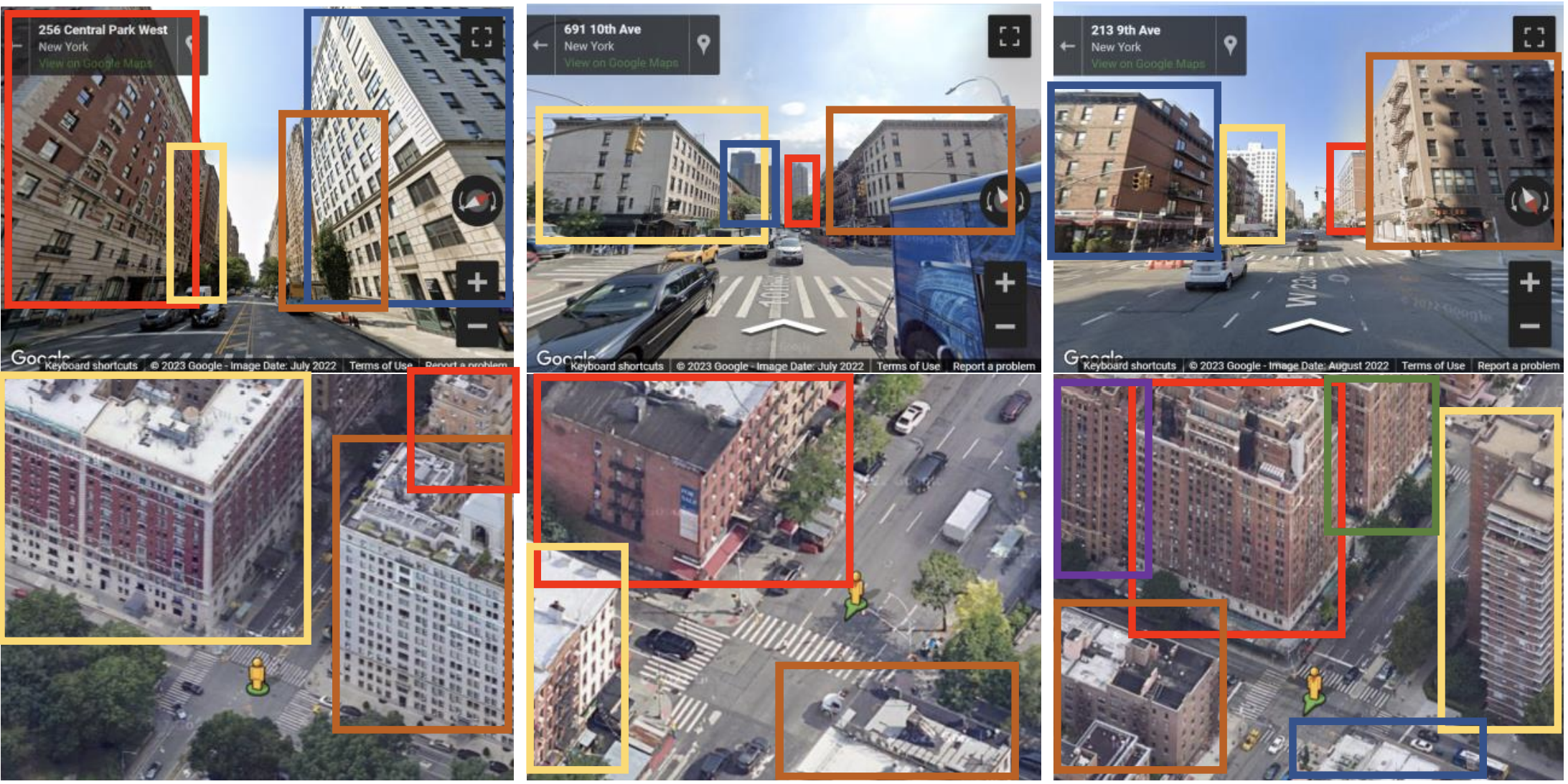}
    \caption{3 Demonstration images pairs from UrbanGeo~\citep{tian2017cross} dataset. The top rows are ground-level images and the bottom rows are corresponding bird's eye view images. The bounding boxes are detected buildings. The image is taken from \citep{tian2017cross}.}
    \label{fig:sample_urban}
\end{figure}

\noindent\textbf{UrbanGeo:\footnote{https://www.crcv.ucf.edu/research/cross-view-image-matching-for-geo-localization-in-urban-environments/}} \citet{tian2017cross} focused on cross-view geo-localization in urban areas and collected data from three cities in the U.S. including Pittsburgh,
Orlando and part of Manhattan. The author collected $8,851$ GPS points in total from these three cities. For each GPS location, 4 different bird's eye view images were captured with heading orientations of $0^{\circ}$, $90^{\circ}$, $180^{\circ}$, $270^{\circ}$, respectively. DualMaps\footnote{http://www.mapchannels.com/DualMaps.aspx} was employed to obtain corresponding street view images from GSV~\citep{GSV} by a given bird's eye view image. Furthermore, the author annotated each building's bounding boxes on street view images as well as on bird's eye view images if it co-exists in both views. Sample images are shown in Figure~\ref{fig:sample_urban}.
\newline\newline

\begin{figure}
    \centering
    \includegraphics[width=0.47\textwidth]{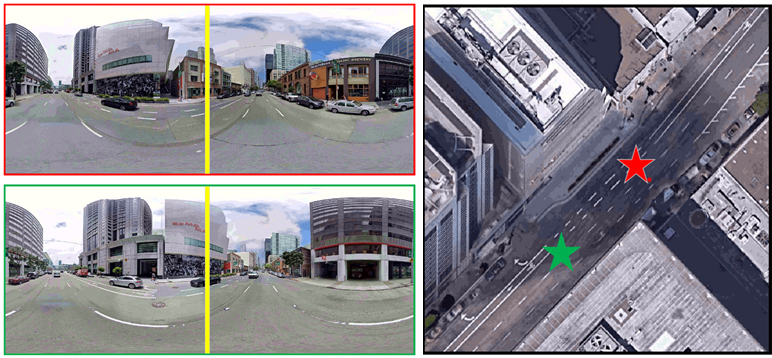}
    \caption{Sample images from VIGOR~\citep{ChenChen} dataset. On the left are two ground-level images covered by the satellite image on right. The yellow line in the ground-level images represents north direction. The color of the star in the satellite image represents the location of the ground-level image with same color border. This image is taken from \citep{ChenChen}.}
    \label{fig:sample_VIGOR}
\end{figure}

\noindent\textbf{VIGOR:\footnote{https://github.com/Jeff-Zilence/VIGOR}}  \citet{ChenChen} collected VIGOR dataset which contains densely sampled locations from four US cities namely:
New York City (Manhattan), San Francisco, Chicago, and
Seattle, using the Google Maps API and GSV API. As compared to the previously mentioned dataset~\citep{CVUSA, liu2019lending, Vo}, VIGOR is more challenging due to the presence of occlusion and shadows by tall buildings in metropolitan areas. Furthermore, this dataset is obtained in more realistic settings. First, the query images and the ground truth reference images do not exactly align with the center like previous datasets~\citep{CVUSA, liu2019lending}. Second, the location of a query ground-level image may be covered by several reference satellite images. VIGOR contains $90,618$ aerial images and $105,214$ ground panoramas. Satellite images were captured at a zoom level~\footnote{https://wiki.openstreetmap.org/wiki/Zoom\_levels} of $20$ (equivalent to a ground resolution of $0.149m$) with an image resolution of $640 \times 640$. The ground images were obtained with a resolution of $2048 \times 1024$. A sample image is shown in Figure~\ref{fig:sample_VIGOR}. VIGOR designed two evaluation settings, called same-area protocol and cross-area protocol. Same-area protocol includes data from all $4$ cities for training and evaluation. On the other hand, cross-area protocol includes images from New York and Seattle for training, and images from San Francisco and Chicago for evaluation.\\
\newline\newline

\begin{figure}
    \centering
    \includegraphics[width=0.47\textwidth]{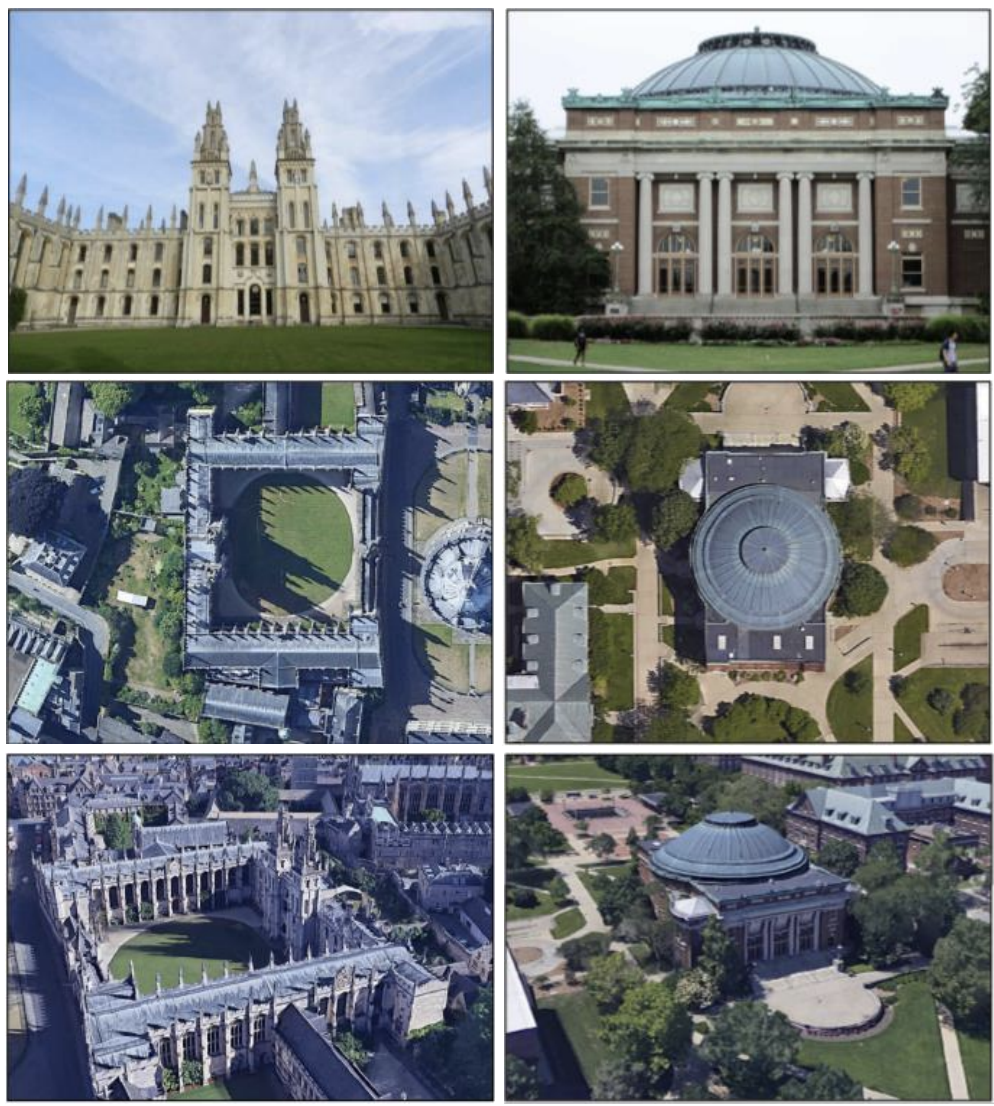}
    \caption{Two sample locations from University-1652~\citep{zheng2020university} dataset. The first row is the ground-level view, the second row is the aerial view, and the bottom row is the drone view. This image is taken from~\citep{zheng2020university}.}
    \label{fig:sample_university}
\end{figure}

\noindent\textbf{University-1652:\footnote{https://github.com/layumi/University1652-Baseline}} %A multi-view multi-source benchmark for drone-based geo-localization\citep{zheng2020university}\\
University-1652~\citep{zheng2020university} is a newly captured drone-based geo-localization dataset. Unlike the aerial-ground datasets~\citep{CVUSA,liu2019lending, Vo, ChenChen}, University-1652 provides three modalities, drone-view, ground-view, and satellite-view. To capture drone views, the authors synthesized images from Google Earth\footnote{https://earth.google.com/web/}. Two sample locations are shown in Figure~\ref{fig:sample_university}. Different from the other datasets in this section which focus on generic cross-view geo-localization, University-1652 highlights the building matching between the three modalities. To this end, University-1652 was built with $1652$ different buildings in which $1402$ buildings contain all $3$ modalities. The author split these $1402$ buildings into training and test sets equally. On average, each building appears in $54$ drone views, $3.38$ ground views, and $1$ satellite view. To be noticed, there is an extra ground view training dataset collected from search engines which results in $16.64$ ground view images per building.
\newline\newline

\begin{figure}
    \centering
    \includegraphics[width=0.47\textwidth]{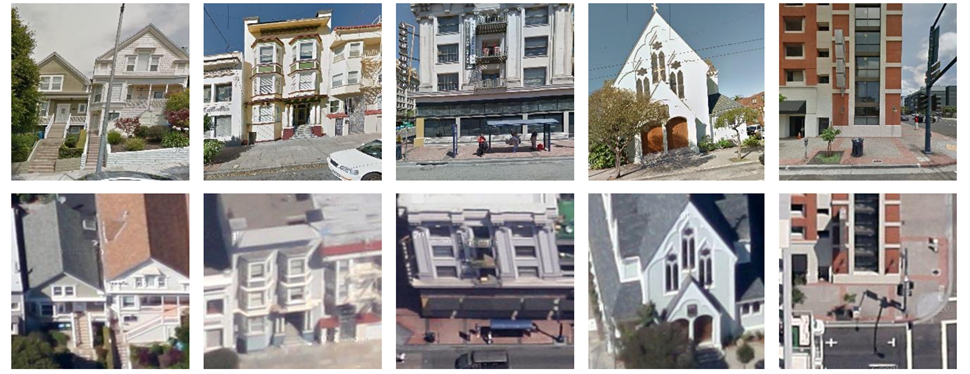}
    \caption{Sample images from DeepGeo~\citep{lin2015learning} dataset. The top row shows five ground-level images. The bottom row shows their corresponding oblique views. This image is taken from \citep{lin2015learning}.}
    \label{fig:sample_deepgeo}
\end{figure}

\noindent\textbf{DeepGeo:} \citet{lin2015learning} proposed DeepGeo which contains 1.1M ground and oblique aerial images. There are around 40K matched pairs and the remaining are unmatched images in the dataset. To capture the corresponding point of interest in oblique satellite imagery, they performed an automatic generation \textcolor{red}{process} which includes a coarse depth plan estimation~\citep{GSV} followed by an image re-projection. Sample images from this dataset are shown in Figure~\ref{fig:sample_deepgeo}. This dataset is not publicly available.

\begin{table*}[!t]
    \tiny
    \centering
    \begin{tabular}{c|ccccccc}
    \toprule \toprule
    & \makecell{CVUSA \\ ~\citep{CVUSA}} & \makecell{CVACT \\ ~\citep{liu2019lending}} & \makecell{Vo \\ ~\citep{Vo}} & \makecell{UrbanGeo \\ ~\citep{tian2017cross}} & \makecell{DeepGeo \\ ~\citep{lin2015learning}} & \makecell{VIGOR \\ ~\citep{ChenChen}} & \makecell{University-1652 \\ ~\citep{zheng2020university}} \\ \midrule
    \# of aerial images & 44,416 & 128,334 & $\approx$1M & $\approx$30K & $\approx$80K & 90,618 & 1652 \\
    \# of ground images & 44,416 & 128,334 & $\approx$1M & $\approx$30K & $\approx$80K & 90,618 & 11,664 \\
    \# of drone-view images & - & - & - & - & - & - & 89,210 \\
    Coverage & Urban,Suburb & Urban,Suburb & Urban,Suburb & Urban & Urban & Urban & Building only \\
    GPS information & Yes & Yes & Yes & Yes & Yes & Yes & Yes \\
    Evaluation metrics & R@K & R@K & R@K & AP, PR curves & AP, PR curves & R@K, Hit Rate & R@K, AP \\
    \end{tabular}
    \caption{Summary of cross-view geo-localization datasets. `R@K' is short for recall at top K.}
    \label{tab:summary_crv_dataset}
% \end{mdframed}
\end{table*}

\subsection{Evaluation Metrics}
The performance of cross-view image geo-localization approaches is usually evaluated by three measurements, recall accuracy at top-$K$(R@$K$), Average Precision (AP), and hit rate. R@$K$, which is similar to the `Reference Ranking' evaluation metric for single-view geo-localization mentioned in Section~\ref{sec:single-view}, is the most popular evaluation metric in cross-view geo-localization and has been used in every dataset as mentioned in Section~\ref{sec:cross_dataset}. $K$ stands for the top $K$ closest matching samples in reference images for a given query image. Euclidean distance is typically employed to measure the similarity~\citep{Hongdong,synthesis,ChenChen,cvmnet,liu2019lending}. For one-to-one retrieval tasks~\citep{CVUSA, liu2019lending, Vo}, it is sufficient to evaluate using a top-$K$ ranking. However, for many-to-one retrieval tasks~\citep{zheng2020university, ChenChen}, \textcolor{red}{top-$K$} ranking is not sufficient to evaluate the performance of the algorithms on multiple true-matched reference images. Thus, University-1652~\citep{zheng2020university} adopted the Average Precision (AP) metric which is the area under the \textcolor{red}{precision-recall} curve. In VIGOR~\citep{ChenChen}, the authors proposed hit rate to evaluate the performance in the many-to-one retrieval task. If the retrieved top-$1$ image covers the query image, it is considered to be a hit. The hit rate is calculated as the ratio of \textcolor{red}{queries} correctly hit to the total number of query images. \textcolor{red}{Table~\ref{tab:summary_crv_dataset} summarizes the} key aspects of all the introduced cross-view geo-localization datasets in section~\ref{sec:cross_dataset} as well as their evaluation metrics.

\begin{table}[!h]

    \footnotesize
    \setlength{\tabcolsep}{2pt}
    \centering
    \begin{tabular}{c r r r r}
    \multicolumn{5}{c}{\textbf{CVUSA~\citep{CVUSA}}} \\
    \toprule \toprule
         Method & R@1 & R@5 & R@10 & R@1\% \\
         \midrule
         MCVPlaces~\citep{CVUSA} & - & - & - & 34.30 \\
         DBL~\citep{Vo} & - & - & - & 63.70 \\
         CVM-Net-I~\citep{cvmnet} & 22.47 & 49.98 & 63.18 & 93.62 \\ OriCNN~\citep{liu2019lending} & 31.71 & 56.61 & 67.57 & 93.19 \\
         FCBAM~\citep{hardTriplet} & - & - & - & 98.30 \\
         Feature Fusion GAN~\citep{bridging} & 48.75 & - & 81.27 & 95.98 \\ CVFT~\citep{featureTransport} & 61.43 & 84.69 & 90.49 & 99.02\\ AlignNet~\citep{revisiting} & 54.50 & - & - & 97.70 \\
         Seeing the Unseen~\citep{rodrigues2021these} & 75.95 & 91.90 & 95.00 & 99.42 \\
         SAFA~\citep{Hongdong} & 89.84 & 96.93 & 98.14 & 99.64\\LPN~\citep{wang2021each} & 85.79 & 95.38 & 96.98 & 99.41 \\
         DSM~\citep{DSM} & 91.96 & 97.50 & 98.54 & 99.67 \\SAFA-GAN~\citep{synthesis} & 92.56 & 97.55 & 98.33 & 99.57 \\
         \textcolor{red}{TransGeo}~\citep{TransGeo} & 94.08 & 98.36 & 99.04 & 99.77 \\
         \textcolor{red}{GeoDTR}~\citep{geodtr} & \textbf{95.43} & \textbf{98.86} & \textbf{99.34} & \textbf{99.86}
    \end{tabular}
    \caption{Benchmark comparison on CVUSA~\citep{CVUSA} dataset.}
    \label{tab:cvusa}
\end{table}

\begin{table}[!h]
    \footnotesize
    \setlength{\tabcolsep}{2pt}
    \centering
    \begin{tabular}{c r r r r}
    \multicolumn{5}{c}{\textbf{CVACT\_val~\citep{liu2019lending}}} \\
    \toprule \toprule
         Method & R@1 & R@5 & R@10 & R@1\% \\
         \midrule
         CVM-Net-I~\citep{cvmnet} & 20.15 & 45.00 & 56.87 & 87.57 \\ OriCNN~\citep{liu2019lending} & 46.96 & 68.28 & 75.48 & 92.01 \\
         CVFT~\citep{featureTransport} & 61.05 & 81.33 & 86.52 & 95.93\\
         SAFA~\citep{Hongdong} & 81.03 & 92.8 & 94.84 & 98.1\\ LPN~\citep{wang2021each} & 79.99 & 90.63 & 92.56 & 97.03 \\
         Seeing the Unseen~\citep{rodrigues2021these} & 73.19 & 90.39 & 93.38 & 97.45 \\
         DSM~\citep{DSM} & 82.49 & 92.44 & 93.99 & 97.32  \\ SAFA-GAN~\citep{synthesis} & 83.28 & 93.57 & 95.42 & 98.22 \\
         \textcolor{red}{TransGeo}~\citep{TransGeo} & 84.95 & 94.14 & 95.78 & 98.37 \\
         \textcolor{red}{GeoDTR}~\citep{geodtr} & \textbf{86.21} & \textbf{95.44} & \textbf{96.72} & \textbf{98.77} \\
    \end{tabular}
    \caption{Benchmark comparison on CVACT~\citep{liu2019lending} validation set.}
    \label{tab:cvact_val}
\end{table}

\begin{table}[!h]
    \footnotesize
    \setlength{\tabcolsep}{3pt}
    \centering
    \begin{tabular}{c r r r r}
    \multicolumn{5}{c}{\textbf{CVACT\_test~\citep{liu2019lending}}} \\
    \toprule \toprule
         Method & R@1 & R@5 & R@10 & R@1\% \\
         \midrule
         CVM-Net-I~\citep{cvmnet} & 4.06 & 16.89 & 24.66 & 56.38 \\ OriCNN~\citep{liu2019lending} & 19.9 & 34.82 & 41.23 & 63.79 \\
         CVFT~\citep{featureTransport} & 34.39 & 58.83 & 66.78 & 95.99\\
         SAFA~\citep{Hongdong} & 55.5 & 79.94 & 85.08 & 94.49\\
         DSM~\citep{DSM} & 35.55 & 60.17 & 67.95 & 86.71  \\
         SAFA-GAN~\citep{synthesis} & 61.29 & 85.13 & 89.14 & 98.32 \\
         \textcolor{red}{GeoDTR}~\citep{geodtr} & \textbf{64.52} & \textbf{88.59} & \textbf{91.96} & \textbf{98.74} \\
    \end{tabular}
    \caption{Benchmark comparison on CVACT~\citep{liu2019lending} testing set.}
    \label{tab:cvact_test}
\end{table}

\begin{table}[!h]
    % \begin{mdframed}[backgroundcolor=red!50,linecolor=red!50]
    \footnotesize
    \centering
    \begin{tabular}{c r}
    \multicolumn{2}{c}{\textbf{Vo~\citep{Vo}}} \\
    \toprule \toprule
    Method & R@1\% \\ \midrule
    MCV~\citep{CVUSA} & 15.40 \\
    DBL~\citep{Vo} & 15.90 \\ 
    CVM-I~\citep{cvmnet} & 59.90 \\
    CVM-II~\citep{cvmnet} & 67.90 \\
    AlignNet~\citep{revisiting} & 88.30 \\
    \end{tabular}
    \caption{Benchmark comparison on Vo~\citep{Vo} testing set. MCV stands for MCVPlaces~\citep{CVUSA}. CVM-I and CVM-II stand for CVM-Net-I~\citep{cvmnet} and CVM-Net-II~\citep{cvmnet} respectively.}
    \label{tab:Vo}
    % \end{mdframed}
\end{table}

\begin{table*}[!h]
    \footnotesize
    \centering
    \begin{tabular}{c | r r r r | r r r r}
    \multicolumn{9}{c}{\textbf{VIGOR~\citep{ChenChen}}} \\
    \toprule \toprule  &
    \multicolumn{4}{c}{\textbf{Same-Area}} & \multicolumn{4}{c}{\textbf{Cross-Area}} \\
    \toprule
    Method & R@1 & R@5 & R@1\% & Hit Rate & R@1 & R@5 & R@1\% & Hit Rate \\
    \midrule SAFA~\citep{Hongdong} & 33.9 & 58.4 & 98.2 & 36.9 & 8.2 & 19.6 & 77.6 & 8.9 \\
    SAFA~\citep{Hongdong} + Mining~\citep{revisiting} & 38.0 & 62.9 & 97.6 & 41.8 & 9.2 & 21.1 & 77.8 & 9.9 \\
    VIGOR~\citep{ChenChen} & 41.1 & 65.8 & 98.4 & 44.7 & 11.0 & 23.6 & 80.2 & 11.6 \\
    \textcolor{red}{TransGeo}~\citep{TransGeo} & \textbf{61.48} & \textbf{87.54} & \textbf{99.56} & \textbf{73.09} & \textbf{18.99} & \textbf{38.24} & \textbf{88.94} & \textbf{21.21} \\
    \end{tabular}
    \caption{Benchmark comparison on VIGOR~\citep{ChenChen} testing set.}
    \label{tab:vigor_test}
\end{table*}

\begin{table}[!h]
    \footnotesize
    \centering
    \begin{tabular}{c r r }
    \multicolumn{3}{c}{\textbf{University-1652~\citep{zheng2020university}}} \\
    \toprule \toprule
    Method & R@1 & AP \\
    \midrule
    CVM-Net-I~\citep{cvmnet} & 53.21 & 58.03 \\
    Where-CNN~\citep{lin2015learning} & 52.39 & 57.44 \\
    Baseline~\citep{zheng2020university} & 58.49 & 63.31 \\
    LPN~\citep{wang2021each} & \textbf{75.93} & \textbf{79.14} \\
    \end{tabular}
    \caption{Benchmark comparison on University-1652~\citep{zheng2020university} testing set.}
    \label{tab:university_1652}
\end{table}

\subsection{Benchmarks}
In this section, we present experimental results on the cross-view geo-localization methods mentioned in previous sections.\newline
\noindent\textbf{Experiments on CVUSA~\citep{CVUSA}:} As compared to other datasets, most of the methods are benchmarked on the \textcolor{red}{CVUSA dataset~\citep{CVUSA}, due to being the} first large-scale cross-view geo-localization dataset. The experimental results are presented in Table~\ref{tab:cvusa}. As expected MCVPlaces~\citep{CVUSA} performed the worst in the benchmark because it is the first proposed method. CVM-Net~\citep{cvmnet} improved the accuracy by a large margin. The recall accuracy at top-$1$ was improved to around $22\%$ and the recall accuracy at top-$1\%$ was improved to above $90\%$. The second large improvement was brought by SAFA~\citep{Hongdong} \textcolor{red}{because of its feature aggregation techniques and the polar transformation}. The \textcolor{red}{recall accuracy} at top-$1$ was improved to nearly $90\%$. \\
\noindent\textbf{Experiments on CVACT~\citep{liu2019lending}:}
The validation set of CVACT~\citep{liu2019lending} includes the same number of ground-satellite pairs as that of the CVUSA~\citep{CVUSA} testing set. However, the testing set \textcolor{red}{for CVACT~\citep{liu2019lending} dataset has} $10$ times more ground-satellite pairs than the CVUSA~\citep{CVUSA} testing set. Thus we report the results on both of them. The results of the CVACT validation set (CVACT\_val) are reported in Table~\ref{tab:cvact_val}. We observe that every method has a performance drop on the CVACT validation set as compared to their performance on \textcolor{red}{the} CVUSA~\citep{CVUSA} dataset. One reason is that \textcolor{red}{CVACT~\citep{liu2019lending}} is densely captured in one single city. Therefore, the ground images in CVACT have less visual distinction than ground images in CVUSA~\citep{CVUSA} dataset. However, SAFA~\citep{Hongdong}, DSM~\citep{DSM}, SAFA-GAN~\citep{synthesis}, \textcolor{red}{TransGeo~\citep{TransGeo}, and GeoDTR~\citep{geodtr} }still achieved \textcolor{red}{greater} than $80\%$ accuracy on $R@1$ on CVACT validation (CVACT\_val) benchmark. The CVACT testing set (CVACT\_test) is even more challenging than the validation set \textcolor{red}{due to its large-scale ground and aerial images}. In Table~\ref{tab:cvact_test}, we observe that the performance of all methods drops further, where \textcolor{red}{only GeoDTR~\citep{geodtr} achieved $64.52\%$ accuracy on $R@1$}. This demonstrates that densely sampled data in a single area is harder than data sampled from random areas.\\ 
\noindent\textbf{Experiments on Vo~\citep{Vo}:} The performance on the Vo~\citep{Vo} dataset is presented in Table~\ref{tab:Vo}. Due to the difficulties of this dataset, such as the limited field of view and the large-scale testing data size, we only present the recall accuracy at top-$1\%$. We observe that the DBL~\citep{Vo} method achieved nearly $60\%$ on recall accuracy at top-$1\%$ and AlignNet~\citep{revisiting} boosted the performance to near $90\%$ because of the global hard example sampling strategy and its novel binomial loss function.

\noindent\textbf{Experiments on VIGOR~\citep{ChenChen}:} VIGOR~\citep{ChenChen} is a relatively new dataset. Only four methods reported results on this dataset which is shown in Table~\ref{tab:vigor_test}. The SAFA+Mining row refers to the SAFA~\citep{Hongdong} model with the global mining strategy proposed by~\citep{revisiting}. The performance of all methods on same-area settings is better than cross-area settings, \textcolor{red}{since the domain gap between training data and testing data under the same-area protocol is less than under the cross-area protocol. VIGOR~\citep{ChenChen} outperformed the two SAFA-based methods} due to its newly proposed IOU-based loss and offset prediction. \textcolor{red}{Benefiting from the advanced transformer architecture and two-stage training, TransGeo~\citep{TransGeo} brought a large improvement in both same-area and cross-area evaluation.}

\noindent\textbf{Experiments on University-1652~\citep{zheng2020university}:} Since University-1652 is a drone-view-focused dataset, we only show the benchmark result on drone-aerial pairs in Table~\ref{tab:university_1652}. The `baseline method' in Table~\ref{tab:university_1652} refers to the baseline model proposed by~\citep{zheng2020university}. LPN~\citep{wang2021each} achieved better results on both recall accuracy at top-$1$ and AP than all other methods because of its square-ring partitioning strategy and local pattern network.

%\textcolor{blue}{Building matching method does not provide exact value on Urban Geo dataset. Rather than they only shows a plot. So I did not report their result here.}

%\section{Conclusion and Future Work}\label{conclusion}
%More realistic datasets \\
%Consistent and standardized dataset formatting \\
%Universally accepted performance metrics \\
%Full end-to-end systems, current systems are %multi-stage and unwieldy \\
\subsection{Discussion and Future Work}

Cross-view image geo-localization \textcolor{red}{has been} developing rapidly in recent years because of \textcolor{red}{the accessibility} of geo-tagged aerial images. Many recent works take the advantage of these publicly available resources to build large-scale datasets. By leveraging large-scale datasets, deep learning models achieved remarkable results on cross-view image geo-localization. However, \textcolor{red}{future developments in} cross-view geo-localization are still achievable in several aspects: 1) Cross-view image geo-localization under low-light environments. 2) Cross-view cross-season image geo-localization.

Current techniques rely on ground-level images which are taken during the daytime. However, one major application area of cross-view image geo-localization is autonomous vehicles which should operate anytime. The objects in low-light images lose details of contextual information which makes \textcolor{red}{global representations hard to predict}. To address this problem, future works can employ an image-to-image translation model which converts low-light images into images taken during the daytime. Another solution is to directly use images taken by special sensors which are designed specifically for low illumination conditions, such as Near Infrared (NIR) sensors. However, the loss of color information from infrared sensors can be a problem.

Similar to cross-view image geo-localization under low-light environments, cross-season cross-view image geo-localization can be a practical problem. \textcolor{red}{Ground images taken in the summer vs the winter have large visual differences due to the heavy snow, ice, and the changes to the trees.} Currently, to the best of our knowledge, no existing dataset collected ground images from the northern area during winter. However, aerial images of such areas are normally captured during the summertime to maximize visibility (most objects are not covered by snow). Thus, there is a need to bridge the temporal domain gap between the ground and aerial images. Future works might solve this problem by using existing domain adaptation methods~\cite {domain1,domain2}. However, there is still a lack of a comprehensive \textcolor{red}{datasets} for benchmarking purposes.

% With the remarkable performance of cross-view image geo-localization on current datasets, it is a natural idea to extend cross-view image geo-localization to cross-view video geo-localization. To tackle this problem, \textcolor{red}{future research} may consider a new framework that takes a video as input and outputs a sequence of satellite images. More importantly, there is no dataset dedicated to cross-view video geo-localization. \textcolor{red}{Future works} may take benefit from the existing publicly available autonomous driving datasets and compose a new cross-view video geo-localization feasibly and efficiently.

\section{Object Geo-localization}\label{sec:stationary}
% Problem Overview
% What is the problem that needs to be solved?
% Applications of OG
% What are the motivations for research in this field?
% Construct map/database of assets
% Maintenance assessments and planning
% Surveying
% What is the difference between depth estimation and object geo-localization?
% Less robust, more noise
% Performed from single image view
% GPS ground truth accuracy can differ wildly due to
% Annotation technique (LIDAR, vs manual)
% Camera Movement Speed
% Hardware Quality (smartphones vs professional equipment)

The basic premise behind object geo-localization is to identify, classify, and determine the geolocations of objects visible in images. This technology enables the construction of automated systems that create maps of objects and their locations using only images as input, avoiding the need for hundreds or thousands of hours of human labor. This technology has applications in self-driving cars, automated mapping of roads, asset management and planning, and land surveying~\citep{uber, nassar1, nassar2}.

% Before providing a detailed discussion of this field, we must establish some clarifications. Object geo-localization differs from depth estimation in two key ways. First, depth estimation algorithms are performed from individual images using a single image view~\cite{depth_estimation_overview}. Object geo-localization spots objects in multiple views for greater reliability~\cite{nassar1, nassar2}. Second, an object geo-localization algorithm aggregate information from multiple images in which detected objects are visible to improve the robustness of geo-localization predictions~\cite{uber, arts}.

\subsection{Techniques}
% Technique Introduction
% Inputs
Object geo-localization can be more thoroughly defined as follows. As input, an algorithm receives a sequence of consecutive images in the order they were taken, often extracted from a video. Each image in the sequence is commonly referred to as a frame. Each frame is tagged with the GPS coordinates at which it was taken. Depending on the dataset, there may be large discrepancies in the frame rate, camera field of view, and the distance the camera moves between images. In some datasets, each frame may contain images from multiple camera perspectives. Some datasets may have additional metadata associated with each image, such as pose information indicating the 6D orientation of the camera.

% Outputs
As output, a stationary object geo-localization algorithm is expected to produce a single object prediction indicating the GPS position for each object that appears in at least one frame. The key challenge in this task is that each object does not appear in a fixed number of images, so algorithms must be designed to `merge' the repeated occurrences of objects across multiple frames into a single, geo-localized prediction for each ground-truth object. Sometimes, this task involves predicting additional outputs such as a class label or the object's condition.
% Most objects will appear in multiple consecutive frames, whereas some objects may appear only once. An algorithm must use some sort of technique to detect these multiple detections and merge them together such that a single final prediction is created for each individual object.

% Depth Estimation
We briefly note that there exists a related field,~\emph{Single Image Depth Estimation}, in which the goal is to predict a depth value for each pixel within an image. By contrast, object geo-localization algorithms are designed to receive additional camera metadata such as the camera's heading and GPS coordinates as inputs to the model, which enables global geo-localization by explicitly predicting GPS coordinates for each object. We direct the reader to~\citep{depth_estimation_overview} for additional information on depth estimation.

% Types of techniques
The most defining challenge in the object geo-localization task is the need to develop some sort of algorithm that merges repeated detections for each object into a single prediction per ground truth object. Therefore, we divide the proposed techniques into three categories based on how they handle repeated detections. \textbf{Tracker-based} approaches implement an object tracking algorithm to identify the same object across multiple frames to merge them together. \textbf{Triangulation-based} approaches implement a triangulation algorithm to find and merge nearby objects. \textbf{Re-identification} approaches use an object detector that detects objects by receiving multiple frames as input and implicitly merging repeated detections from the input frames. Next, we introduce each of the above-mentioned concepts in detail and then survey the popular implementations in the field.

% Approaches: 3D methods, tracker-based methods, non-tracker based methods
% The first fundamental form of object geo-localization involves geolocalizing the position of stationary objects, typically using sequential image frames captured as a camera moves past the object. This procedure can be performed from one or more camera views. Object may belong to one or multiple classes, meaning some models may perform a hybrid of classification an geo-localization. Due to the lack of standardized datasets in the field, most papers construct their own. We therefore provide both a survey of major contributions in addition to an overview of any datasets both authors have constructed. We take special care to note any limitations of built datasets, because small discrepancies in how geolocation datasets are constructed can have an unexpectedly large influence on its difficulty and applicability.

\subsubsection{Tracker-Based geo-localization}
% approach overview
All tracker-based approaches begin with an object detector that receives images as input and produces a detection for each object of interest. This leaves the need to merge repeated detections from objects appearing in multiple frames. Tracker-based approaches solve this problem by implementing an object tracker that performs object tracking by associating objects across frames. A list of assigned detections from a sequence of frames is referred to as a \textbf{tracklet}. It contains what is believed by the algorithm to be all the occurrences of the same object. These approaches will typically implement a heuristic approach to condense each tracklet into the final, geolocalized object predictions.

\begin{figure*}
    \centering
    \includegraphics[width=0.99\textwidth]{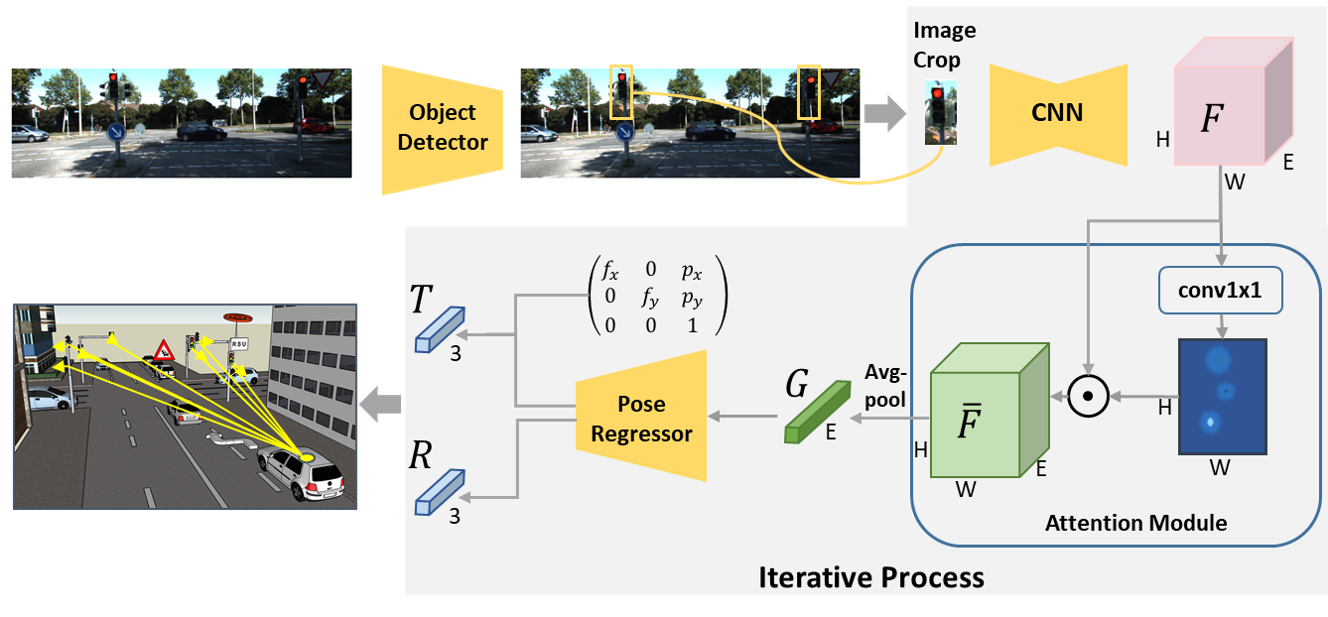}
    \caption{The object detector proposed by~\citep{uber} that regresses 5D poses for objects. The figure is taken from~\citep{uber}.}
    \label{fig:uber1}
\end{figure*}

% uber paper
~\citep{uber} constructed a three-stage system that was mostly end-to-end trainable. The first stage detected objects, regressed their bounding boxes, and predicted the 5D pose of visible objects as illustrated in Figure~\ref{fig:uber1}. An object's 5D pose contains its translation vector along the X, Y, and Z axis as well as its rotation along the two-axis vector orthogonal to the camera. GPS coordinates for an object were predicted by performing a coordinate transformation using the object's 5D pose relative to the camera's GPS coordinates. From each detection, the authors harvested geometric features predicted by the pose network in addition to visual information extracted from a CNN. These features were provided as input to the object matching network, which they trained to compute affinity scores between each possible pair of detected objects between frames. The output of the matching network was a matrix where each entry indicated the predicted similarity between a pair of detections. Finally, the Hungarian algorithm was employed to compute the optimal pairings between objects. The main advantage of this approach is that it is mostly end-to-end trainable. A noteworthy disadvantage is that it requires object 5D poses and the camera intrinsic matrix, which may be unavailable in many applications.

% \begin{figure*}
%     \centering
%     \includegraphics[width=1\textwidth]{object-fig-2-uber.png}
%     \caption{The affinity network and tracking system proposed by~\citep{uber}. Figure is taken from~\citep{uber}.}
%     \label{fig:uber2}
% \end{figure*}

\begin{figure}
% \begin{mdframed}[backgroundcolor=red!50,linecolor=red!50]
    \centering
    \includegraphics[width=.49\textwidth]{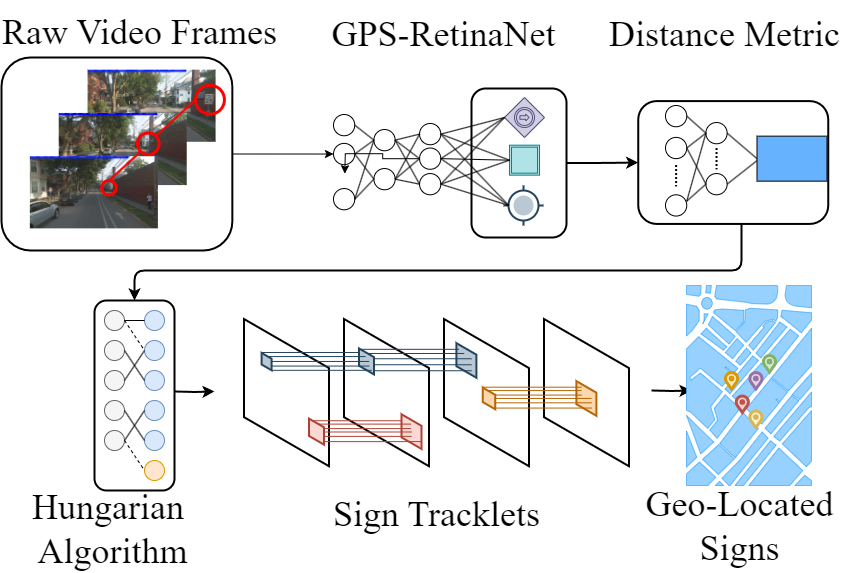}
    \caption{\textcolor{red}{The full end-to-end system including the object detector and tracker proposed by~\citep{arts}. Figure is taken from~\citep{arts}.}}
    \label{fig:arts}
% \end{mdframed}
\end{figure}

% arts paper
\citep{arts} modified the popular object detector RetinaNet~\citep{RetinaNet} with an additional subnet to predict GPS coordinates. They trained a second neural network which was designed to predict similarity scores between each pair of sign detections between consecutive frames. %This network received features from each pair of detections along with a tensor containing a summary of all other visible detections from each frame being compared as input. 
The authors used the Hungarian Algorithm to pair detections of objects across multiple frames. The algorithm was modified with a cutoff value to prevent traffic signs with low similarity from being paired. %By stepping through an image sequence and applying this pipeline at each step, multiple object tracking is preformed. 
Finally, each tracklet obtained through this procedure was condensed into a single geolocalized sign prediction with a simple weighted average. The full pipeline is displayed in Figure~\ref{fig:arts}. Compared to~\citep{uber}, this approach can be run with more commonly accessible hardware due to not requiring 5D poses, and has the additional capability to predict object classes as part of its geo-localization pipeline. However, the system is more complicated and not end-to-end trainable.

\subsubsection{Triangulation-Based Object Geo-localization}
% approach overview
Similar to tracker-based approaches, triangulation-based methods use an object detector to provide initial detections for objects in individual frames. However, triangulation approaches use an alternative method for condensing and geo-localizing repeated detections. Triangulation works by first constructing a triangle with three points. In the case of object geo-localization, there would be two points on the triangle corresponding to two camera locations, and then the third point would be the location of the object since it should remain constant between frames. The distance between images can be calculated from GPS metadata, and the angle between the camera and the object can be predicted by the object detection model. This allows the distance to the object to be predicted from this triangle using trigonometry. The calculated distance can serve as a known value in the next triangle, which is constructed by connecting the next frame in the sequence to this edge. This process can be repeated as many times as necessary to connect and merge object detections. In real applications, object positions may be very noisy and the rays drawn from object positions may not always intersect to form a perfect triangle.~\citep{triangulation} proposed a method robust to noise by formulating triangulation as a minimum least squares problem. Note that~\citep{geometry_computer_vision} provides a thorough description of various triangulation and geometric computer vision methods.

\begin{figure*}
% \begin{mdframed}[backgroundcolor=red!50,linecolor=red!50]
    \centering
    \includegraphics[width=1\textwidth]{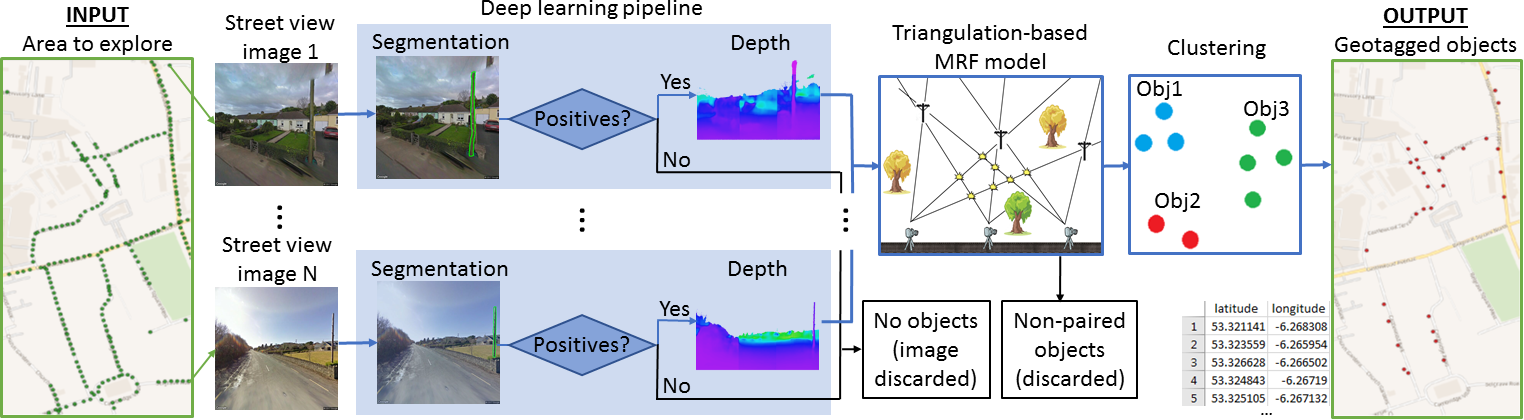}
    \caption{\textcolor{red}{The full pipeline including the detector, triangulation, and clustering proposed by~\cite{MRF}. Figure is taken from~\cite{MRF}.}}
    \label{fig:MRF}
% \end{mdframed}
\end{figure*}

% MRF paper
In practice, a hybrid approach is typically employed in which triangulation is one of components in a geo-localization pipeline.~\citet{MRF} built a complete geo-localization system involving two convolutional networks, the first of which segments objects visible in images, and the second estimates the depth of each detected object. After using these networks to detect objects and estimate their depths, objects are geolocated using a Markov Random Field model to perform triangulation. This model optimizes an energy function to yield refined predictions for objects and their geolocation from each pair of images. Since objects are still likely to appear in more than two frames (even after triangulation), the authors filtered the redundant object detections using a hierarchical clustering algorithm. The full pipeline is shown in Figure~\ref{fig:MRF}. This approach is conceptually simple and has strong explainability due to the use of \textcolor{red}{triangulation} to determine GPS coordinates, however, has less performance than deep learning-based GPS prediction methods.

\subsubsection{Re-Identification-Based Object Geolocalization}
% approach overview
Another approach to address the issue of repeated detections is to inherently detect the object from multiple views. Instead of receiving a single frame as input, a detector receives one or more nearby frames as simultaneous inputs, and learns to jointly predict and re-identify the same object between frames as a single output. By performing detection from multiple perspectives, the object detector implicitly merges repeated detections from multiple frames.

\begin{figure}
% \begin{mdframed}[backgroundcolor=red!50,linecolor=red!50]
    \centering
    \includegraphics[width=.49\textwidth]{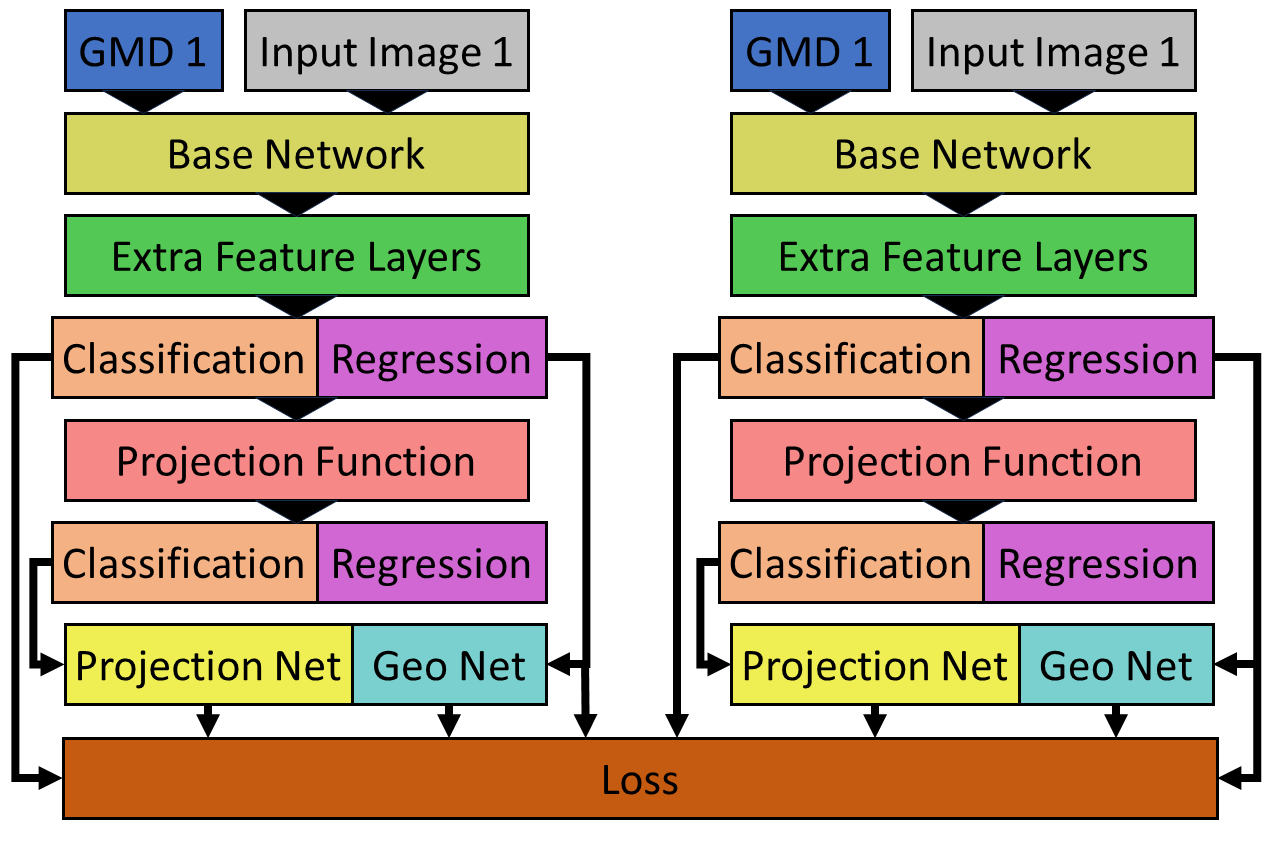}
    \caption{The siamese network that learns to re-identify objects between a pair of frames proposed by~\citep{nassar1}. Figure is taken from~\citep{nassar1}.}
    \label{fig:nassar1}
% \end{mdframed}
\end{figure}

% nassar 1st paper
\citep{nassar1} used an approach in which an object's geolocation is predicted using multi-view geometry. The system contained a siamese network that learned to detect, re-identify, and geolocalize objects using a pair of images as simultaneous inputs, as illustrated in Figure~\ref{fig:nassar1}. After regressing bounding boxes in one image, their model learned a transformation to project those bounding boxes to the perspective of the other image. This gave the model the capability to detect and predict the geolocation of objects by learning the joint distribution from both images. Since only a single prediction is generated for objects appearing in both images, their model implicitly merges detections from each pair of images. This means their model did not require a separate object tracker to merge repeated detections unlike~\citep{uber, arts}. The limitation of this approach is that it can not collapse redundant detections from objects visible in greater than two frames.

% nassar 2nd paper
\citep{nassar2} proposed an algorithm capable of geo-localizing objects using an arbitrary number of images as input. First, images were fed into an object detector, from which fused features across all input images were extracted and used to predict bounding boxes and classes for each visible object. To associate detections of the same objects, a graph neural network (GNN) was used. Each node in the graph represents a feature vector corresponding to a proposed object from the detector, and each edge weight encodes if two nodes are from the same object. The GNN was trained to learn weights on the edges, allowing geolocalized object detections to be produced by finding each connected component in the graph. The proposed architecture is fully end-to-end trainable. Since this approach can merge objects from greater than two images, this approach solves the previous pitfall present in~\citep{nassar1}.

% \begin{figure}
%     \centering
%     \includegraphics[width=0.5\textwidth]{object-fig-5-geograph.png}
%     \caption{The multi-view detection pipeline proposed by~\citep{nassar2}. The system is capable of detecting objects in multiple images through the use of a graph neural network. Figure is taken from~\citep{nassar2}.}
%     \label{fig:nassar2}
% \end{figure}
% \subsubsection{3D Object Geolocalization}
% % approach overview
% % TODO
% ~\citep{MonoDIS}

\subsubsection{Other Approaches}\label{Single Object Other Approaches}
Since detecting an object in a satellite image with known coordinates implicitly provides the approximate geolocation of the object, this task could be interpreted as an approximate form of geo-localization.~\citep{super_res_satellite} proposed to enhance satellite image quality using the Very Deep Super-Resolution Network~\citep{VSDR}. They trained object detection models on the xView dataset~\citep{xView} to detect common objects from satellite images such as planes and boats. To reduce the burden of human annotation, ~\citep{synthetic_satellite} proposed a two-step system. First, they generated synthetic satellite images of objects by employing CycleGAN~\citep{cycleGAN}. 
Second, they trained RetinaNet~\citep{RetinaNet} on the xView dataset. Then they used the synthetic images to determine a confidence score threshold to filter detections, which reduced false positives.~\citep{dota} provides a modern satellite image object recognition dataset that addresses several issues associated with previous datasets.
Specifically, they provide a large number of samples per class and propose a method of drawing bounding boxes to address the overlap typically associated with objects in satellite images. The key disadvantages are that smaller objects such as electric poles may not be visible, and different classes of objects such as traffic signs may not be distinguishable from an aerial perspective.
\subsection{Datasets}\label{sec:object_datasets}
% Overview
% unlike other fields where datasets same task comparable
% different problems solved
% different annotations methods
% different attributes associated class
% different distances from camera, camera speed, frame rates, number of camera views
% different hardware
% different camera perspectives
% most datasets contain camera metadata: GPS, potentially heading and orientation (pose)
% contain annotations indicating objets per frame
% object re-identification annotation
% no standard benchmarks in field
% most papers construct their own datasets and benchmark exclusively on that
Multiple datasets have been constructed to benchmark the performance of object geo-localization algorithms. Datasets in this field contain a few common characteristics. First, they all contain frames organized sequentially in the order in which they were taken. Each frame in these datasets has GPS metadata associated with them, which indicates the longitude and latitude at which the frame was captured. Second, all datasets contain per-frame annotations of objects. Each object annotation specifies a bounding box around the object along which either its GPS coordinates or a distance offset relative to the frame that can be used to recover the object's GPS coordinates using a coordinate transform. Third, each dataset needs to have some method of identifying repeated appearances of the same object in multiple frames, to provide the ground truth against which to compare the final geo-localization performance of the algorithm. Many datasets accomplish this with a unique label associated with each annotation, such that the same object will have the same label in each of its annotations. Some datasets provide a list of each of the alternate frames the object appears in as part of the object's annotation. Finally, a \textcolor{red}{few datasets} simply assume that all annotations within a certain distance of each other correspond to the same object.

% Dataset differ
Unlike other fields where datasets contain clearly defined inputs and outputs and existing models can be re-trained on other datasets, the field of object geo-localization is yet to adopt a standardized structure or consistent set of testing datasets. While all datasets contain bounding box and geolocation information associated with each annotation, datasets differ in terms of their annotation structure, available camera metadata, construction techniques, and hardware. Some datasets contain additional metadata such as the camera heading or orientation (also referred to as pose) associated with each image. This may serve as additional input to assist the object detector in identifying objects and predicting geolocation. Finally, datasets are built using different hardware. Many techniques mine publicly available images created with consumer equipment such as GPS-tagged cellphone images, whereas other datasets may be professionally constructed using enterprise cameras and LIDAR sensors. As a result of hardware differences, the resulting datasets have widely varying frame rates, movement speeds, and object sizes and distances from the camera. Some datasets may have multiple camera perspectives as part of a single frame.

Due to these differences in datasets, models in this emerging field are currently not cross-dataset-compatible. Models are typically built to use specific inputs associated with camera metadata or object annotations unique to that dataset, and would need to be restructured significantly to train on other datasets. Approaches therefore construct their own datasets and build their model specifically based on how those datasets are structured. This poses a significant challenge when comparing benchmarks across different datasets, as discussed in~\ref{sec:object_benchmarks}. Next, we will discuss each proposed dataset and provide a link to datasets that are publicly available. In Table~\ref{tab:dataset-stats}, we compare the popular single-object geo-localization datasets.
\newline
% TODO: Add nuscenes maybe
% https://www.nuscenes.org/

% Uber traffic light dataset
\noindent\textbf{Uber-TL:}\footnote{\url{https://github.com/MedChaabane/Static_Objects_Geolocalization}}~\citep{uber} constructed a dataset in which the objective is to geo-localize traffic lights from images captured from a roadside vehicle. The authors mined their data from nuScenes~\citep{nuScenes}, a pre-existing dataset commonly used for autonomous driving. The authors selected scenes from this dataset containing road intersections with traffic lights within 100 meters of the camera. Object coordinates are expressed relative to the camera's perspective, and each frame contains images from six separate camera angles as the vehicle moves down the road. The total dataset contains 96,960 distinct images.

This dataset contains very accurate GPS locations for traffic lights due to the use of LIDAR. There are many camera angles available providing a wide field-of-view and multiple camera angles for most objects. The images are captured at a high resolution and high frame rate. The limitation of this dataset is that many of these ideal criteria may not be met in real-world settings. For example, it is often unfeasible to capture data at six simultaneous camera angles, especially at a high frames rate due to storage constraints. Traffic signs which were not visible in five distinct keyframes were also removed from the dataset, artificially reducing the difficulty of the dataset relative to real-world settings. %It should also be noted that because GPS ground truth for the lights was annotated by calculating the distance offset relative to the image measured by LIDAR, inaccuracies in the GPS coordinates in the image would shift the object GPS by the corresponding amount, and thus the impact of camera GPS noise on the model may not be fully accounted for when benchmarking
% Explain "offset" problem dataset arguably TOO accurate
% Limitations:
% Dataset: Talk about traffic light dataset
% Limitations to mention: only one class of object, high frame rate, many camera angles (6), objects locations determined using LIDAR which may be more accurate than other systems, scenes only selected if traffic lights available in at least five different keyframes, don't remove blocked images
\newline

% ARTS v2 dataset
\noindent\textbf{ARTSv2:}\footnote{\url{https://drive.google.com/drive/folders/1u_nx38M0_owB0cR-qA6IOWgZhGpb9sWU?usp=sharing}}~\citep{arts} built a dataset of street-view images in which the objective is to classify and geo-localize traffic signs. The dataset contains 199 unique sign classes, 25,544 images containing at least one annotation, and 47,589 unique annotations. Each annotation specifies the sign's class, a bounding box around the sign, a unique integer identifier used to indicate the same sign in different images, and attributes specifying which side of the road the sign is placed on and whether it is a part of an assembly. Each image contains camera metadata indicating its heading and geolocation. The strengths of this dataset are that it has a broad and imbalanced distribution of classes and a variety of different environments, which is very representative of what a geo-localization model would encounter in the real world. Its limitations are noisy GPS coordinates and potentially inconsistent class labels due to the use of human annotators.
\newline

% MRF traffic lights dataset
\noindent\textbf{MRF-TLG:}~\citet{MRF} constructed two datasets. First, they built a traffic light dataset (different from~\citep{uber}) using data samples mined from Mapillary vistas~\citep{vistas} and Cityscapes~\citep{cityscapes}. In total, the dataset contains around 18,500 images scaled to 640$\times$640 resolution. The main limitation of this dataset is that objects within one meter are assumed to be the same object. This could be considered a fairly generous assumption, since many separate traffic lights are commonly placed on the same pole, and scenarios in which objects are close together are most challenging.
% Limitations: 
% Limitations: assume objects within one meter are same object, one sign class
\newline

% MRF telephone pole dataset
\noindent\textbf{MRF-TP:} The second dataset created by~\citep{MRF} contains 20,000 images of telephone poles and 15,000 Google Street View (GSV) images. The authors note that the GPS coordinates are very inaccurate in this dataset due to inconsistent coordinate labels for the GSV images combined with objects being frequently blocked.
\newline

% Nassar Pasadena Multi-View Dataset
\noindent\textbf{PMV:}~\citep{nassar1} constructed two datasets. First, they created the Pasadena Multi-View Re-Identification dataset containing streetside trees~\citep{nassar1}. This dataset contains 6,020 trees labeled in 6,141 panoramas taken from Google Street View (GSV). Each tree appears in exactly four panoramas, leaving a total of 25,061 bounding boxes indicating the location of trees. Each annotation provides the geo-location of the tree, and a unique identifier distinguishing it from other trees. Camera locations are also available in each image as metadata. The main limitation of this dataset is that each tree appears in exactly four images, and the images selected to be part of the dataset are those for which the tree is closest to the camera. This is a significant limitation since closer objects are far easier to geo-localize, artificially making the task less difficult than real-world applications.
% Limitations
% Each tree appears in exactly four panoramas
\newline

% Nassar Mapillary Dataset
\noindent\textbf{GeoSign:}~\citep{nassar1} built a second dataset sourced from Mapillary~\citep{vistas} involving traffic signs. This dataset contains 31,442 traffic signs and a total of 74,320 images. Each annotation specifies the sign's geolocation and altitude, a polygon surrounding the sign, and a list of images in which the sign appears. Camera location metadata is also available for each image. The main limitation of this dataset is inconsistencies in image quality and resolution, due to the crowd-sourced nature of the dataset.
% Limitations
% Non-standardized class labels
% Noisy GPS labeled by hand

\begin{table*}[] 
   \small
    \centering 
    \begin{tabu} to \linewidth {l C C cc C} 
        \toprule 
        & \textbf{PMV}~\citep{nassar1} 
        & \textbf{TLG}~\citep{uber}
        & \multicolumn{2}{c}{\textbf{ARTS v1.0}~\citep{artsv1}} 
        & \textbf{ARTS v2}~\citep{arts} \\ 
        & 
        &
        & \textsc{Easy} & \textsc{Challenging} 
        &  \\ \midrule \midrule 
        Number of classes       & 1 & 1 & 62 & 175 & 199 \\ \midrule
        Number of images        & 6141 & 96960 & 6807 & 16023 &  25544 \\ \midrule
        Number of annotations   & 25061 & N/A & 9006 & 27181  & 47589 \\ \midrule
        Side of the road        & & &  &  &  \checkmark \\ \midrule
        Assembly                & & &  &  &  \checkmark \\ \midrule
        Unique Object IDs       & \checkmark & \checkmark &  &  &  \checkmark \\ \midrule
        5D Poses                & & \checkmark & & & \\ \midrule
        GPS                     & \checkmark & \checkmark & \checkmark  & \checkmark &  \checkmark \\ \midrule
        Color Channels          & RGB & RGB & RGB & RGB  & RGB \\ \midrule
        Image Resolution        & $2048\times1024$ & $1600\times1900$ & $1920\times1080$ & $1920\times1080$ & $1920\times1080$ \\ \midrule
        Publicly Available      & & \checkmark & \checkmark & \checkmark & \checkmark \\ \bottomrule
    \end{tabu}
    \caption{A comparison of the characteristics of popular single-object geo-localization datasets. Table is taken from~\citep{arts}.
    } 
    \label{tab:dataset-stats} 
\end{table*}

\subsection{Performance Benchmarks}\label{sec:object_benchmarks}
We would like to preface this topic by emphasizing the caveat on performance metrics in this field. First, there are no standard performance metrics agreed upon to evaluate algorithms across this field. Second, the method to compute true positives, false positives, false negatives, and the distribution of distance errors differ between different approaches, mainly due to differences in dataset construction techniques. Some methods count all nearby objects within a certain distance threshold as a single object. Other methods ignore objects outside of a certain distance from the camera during evaluation. Due to the lack of standard test datasets and performance metrics, we can not perform a direct side-by-side comparison between the proposed methods. Therefore, we list each algorithm, the datasets it was benchmarked on, and provide details of how the evaluation was performed. A summary of the results is provided in Table~\ref{tab:object_performance}.

~\citet{uber} benchmarked on the traffic light dataset they constructed themselves. Since their system contains both a 5D pose regression network and an object detector, they provided separate benchmarks for the performance of each component. To benchmark the 5D pose network, the authors reported the translation and rotational error for both objects within 20 meters and for all objects at any distance. They reported average margins of error of $2.51$ meters and $14.21\deg$ for objects within 20 meters, and $4.43$ meters and $15.97\deg$ for objects at any distance. They analyzed the tracker's performance on their dataset using the Multiple Object Tracking Accuracy (MOTA) metric and reported a score of 85.52\%. To benchmark geo-localization performance, the authors defined a true positive as a traffic light prediction within 2 meters of Euclidean distance or 3 meters of Mahalanobis distance of the ground truth. They provided precision-recall curves for both units of measurement. The approximate precision-recall values from the `elbows' of the PR curves were 0.75 and 0.45, and 0.85 and 0.55 when using Euclidean and Mahalanobis distances respectively.

% arts evaluation method and reported performance
~\citep{arts} benchmarked a dataset they constructed using US traffic signs. Because the dataset contains hand-labeled annotations from low frame-rate videos, GPS noise is likely to be greater than other methods. They defined a true positive to be a scenario in which the geo-localization error plus a class mismatch penalty is less than 15 meters. They provided separate benchmarks for their object tracker and geo-localization performance. The authors reported an object detection mean average precision (mAP) of 0.701. When benchmarking geo-localization performance using the aforementioned definition of true positives, the authors reported a precision and recall of 0.81 and 0.708 respectively.

% nassar1 + 2 evaluation method and reported performance
~\citep{nassar1} benchmarked their performance on both the datasets they constructed, one of which contained trees and the other contained traffic signs. Note that this dataset does not involve sign classification. The authors reported multi-view re-identification results of 0.731 and 0.882 with mean average geo-localization errors of 3.13 and 4.36 on the respective datasets. The same two datasets were used to benchmark the multi-view re-identification performance of the graph neural network proposed in~\citep{nassar2}. When performing re-identification from six views, they reported a mean average precision of 0.763 and 0.924 and mean geo-localization errors in meters of 2.75 and 4.21 on the respective datasets.\\
% MRF evaluation method and reported performance
~\citep{MRF} tested their triangulation-based approach on the two datasets they constructed containing traffic lights and telegraph poles. They benchmarked their detector by first segmenting the pixels of objects in the image. They defined a true positive to be a scenario in which the segmentation network correctly labeled over 25\% of the object's pixels. Objects greater than 25 meters from the camera were ignored since they would cause a large performance drop. Using this method, they reported precision and recall values of 0.951 and 0.981 respectively on their traffic light dataset and 0.979 and 0.927 respectively on their telegraph pole dataset. To evaluate geo-localization performance, the authors defined a true positive as a situation in which a predicted object was within 2 meters of the ground truth. An important limitation to be noted is that annotated objects within 1 meter were assumed to be the same object. Using these definitions, the authors reported precision and recall values of 0.940 and 0.922 respectively on their traffic light dataset and 0.926 and 0.973 respectively on their telegraph pole dataset. They reported $95^{th}$ percentile geo-localization margins of error of 1.89 and 2.07 meters on the respective datasets.

\begin{table*}[]
    \footnotesize
    \centering
    \begin{tabular}{c|c|c|r|r|r|r}
         Method & Technique & Dataset & Mean GPS Error (m) & Precision & Recall & MAP \\
         \midrule \midrule
         UBER-NET~\citet{uber} & Tracker-Based & TLG~\citet{uber} & 4.43 & $0.75^1$ & $0.45^1$ & 0.928 \\
         GPS-RetinaNet~\citet{arts} & Tracker-Based & ARTSv2~\citet{arts} & 5.81 & 0.810 & 0.708 & 0.701 \\
         Multi-View~\citet{nassar1} & Re-Identification & PMV~\citet{nassar1} & 3.13 & N/A & N/A & 0.731 \\
         Multi-View~\citet{nassar1} & Re-Identification & Mapillary TLG~\citet{nassar1} & 4.36 & N/A & N/A & 0.882 \\
         GeoGraph~\citet{nassar2} & Re-Identification & PMV~\citet{nassar1} & 2.75 & N/A & N/A & 0.763 \\
         GeoGraph~\citet{nassar2} & Re-Identification & Mapillary TLG~\citet{nassar1} & 4.21 & N/A & N/A & 0.924 \\
         MRF~\citet{MRF} & Triangulation & Traffic Lights 2~\citet{MRF} & 1.89 (95th percentile) & 0.922 & 0.940 & N/A \\
         MRF~\citet{MRF} & Triangulation & Telegraph Poles~\citet{MRF} & 0.98 & 0.973 & 0.926 & N/A \\
    \end{tabular}
    \caption{A comparison of performance metrics reported by different stationary object geo-localization algorithms. Note that the structure of the datasets and the performance evaluation methods} differ drastically. Values marked $^1$ are approximate values taken from the "elbow" of the graph illustrating the precision-recall curve.
    \label{tab:object_performance}
\end{table*}
\subsection{Discussion and Future Work}
% broad applicability
% solves very practical problems
% much data can be mined from internet
% future Direction: needs more standardized datasets and performance metrics
% future Direction: collect more diverse datasets with different types of assets, different environments, more diverse regions of the world
% future Direction: improve performance on poor resolution images, since much data is mined from internet and consumer equipment this is what may be available in real-world especially developing countries
% future Direction: make algorithms robust to small datasets since annotations can be challenging and insconsistent
Object geo-localization is a rising field with broad applicability to a very practical set of problems. One advantage of this field is that there is an enormous amount of geotagged images that can be mined from the internet, especially since modern smartphones embed GPS coordinates in an image's EXIF data. Unfortunately, annotation can be challenging and inconsistent which is currently the largest shortcoming in the field. In the future, researchers may consider developing semi-supervised approaches to reduce the large annotation burden in this field. Similarly, the field needs to accept standardized datasets and performance metrics so that different methods can be properly compared. Researchers should also consider collecting datasets from more diverse environments, as most current datasets are limited to various road assets. Land surveying, for example, is a domain that is currently under-represented. These algorithms are particularly \textcolor{red}{applicable in} less developed countries where assets are less managed, creating demand for automated tools. Future research could address this demand by enhancing algorithm performance on low-resolution images, and making algorithms robust to work on smaller datasets and in cross-dataset settings.
%as would likely be available in these contexts.

% PAPERS TO ADD
% % Unified Vision‐Based Methodology for Simultaneous Concrete Defect Detection and Geolocalization

\section{Conclusion}
% What did we do in this paper?
% What are the categories we divided into

% What are applications of this field
% land survey
% self driving cars
% traffic sign asset management
% increasingly applicable in internet age with technological progress

% main idea single object geolocalization take image geolocalize it
% can be accomplished using (list each method)
% repeat this process for each of three types

% What are future areas of research
% potential in self-supervised learning
% incorporating transformer architectures
% low resolution images and less data for poorer areas
% larger datasets, better take advantage of internet for crowd sourced data sets, broader set of assets to geolocalize, greater image quality
% better find features and method invariant of perspective to translate between domains

In this paper, we surveyed noteworthy approaches within the broad field of geo-localization involving images and objects within these images. We divided geo-localization into three key sub-fields.

Single-view geo-localization approaches attempt to predict GPS coordinates taken from a single perspective. A simple approach to accomplish this task is to divide the surface of the earth into cells and which enables the formulation geo-localization as a classification problem in which the goal is to classify which cell an image originates from. The greatest strength of these approaches is their simplicity. Because classification is already a well understood and thoroughly studied problem in machine learning, existing classification models can be easily adapted to this solution. However, the accuracy of these predictions is limited based on the size of the divided cells, and division into smaller cells results in fewer training samples per class. An alternative approach is to match a query image against a similar image from a reference database with known coordinates. This allows for a direct prediction of GPS coordinates as opposed to limiting geospatial predictions to an approximate cell. The main disadvantage of this approach is that as the size of the reference database grows, the number of images against which the query must be compared increases which can become computationally prohibitive. It is also challenging to extract features that match the query to reference images regardless of illumination, camera angle, vegetation, etc. A less popular technique involves refining pre-existing GPS coordinates instead of predicting them from scratch. These approaches can achieve greater performance by using the pre-existing coordinates as a starting prediction, but the obvious disadvantage is these approaches require that GPS coordinates are already available. A final class of approaches involves geo-localizing rural images by aligning their skyline features with a digital elevation model. These approaches have the capability to geo-localize images in rural environments where fewer descriptive features are available, but their performance is limited due to the more difficult task formulation.

All single-view methods are inherently restricted in their capabilities due to only receiving a single ground image perspective as input. Cross-view approaches were designed to address these limitations by incorporating aerial images to increase performance. In cross-view geo-localization, a database of satellite imagery is incorporated to more robustly determine the location of ground-view imagery. Joint feature extraction methods aim to extract hand-crafted features that are similar for ground and satellite images from the same location. These methods are able to exploit a large database of satellite imagery, but their capabilities are limited by their use of hand-crafted features. To address this limitation, Siamese network approaches were developed to automatically learn features to associate ground and satellite images from the same location. These methods do not, however, explicitly model objects such as trees and buildings visible in an image. To accomplish this, graph-based approaches were proposed in which each landmark from an image was modeled with a node, and the connectivity between landmarks was modeled with an edge. This problem formulation yields greater interpretability than other approaches. All of these approaches, however, suffer from the issue of losing certain low-level features present in ground-view images. A final approach leverages generative models to create artificial ground-view images, which can then be directly matched against real ground-view images to perform geo-localization. This process trains a model to directly convert aerial to ground images thus maintaining as many features as the model is capable of. The primary disadvantage is the reliance on GANs, which are known to be difficult to work with.

A limitation of both single and cross-view approaches is that they are exclusively designed for the task of predicting image locations. Object geo-localization algorithms were designed to instead geo-localize objects visible within the images. These approaches are differentiated by the methods they use to condense repeated detections of the same object. Triangulation approaches use changes in camera and object depth to triangulate object positions. These approaches are conceptually simplistic, but they rely on precise camera parameters to be accurate, and typically require a separate algorithm (such as clustering) to completely condense repeated detections. Tracker-based approaches attempt to track the position of the same object between images. These approaches have high performance but require that the \textcolor{red}{input} images are sequential. Re-identification approaches build detection models that receive multiple images as input to produce a single output detection, which implicitly merges the repeated occurrences of objects across the input images. These approaches fail, however, if an object doesn't appear in enough input images for the detector. Furthermore, the detectors only merge objects from a fixed number of images.

Advancements in machine learning, computer vision, and available hardware provide continual opportunities to improve the state-of-the-art in image geo-localization. Future research can focus on the use of self-supervised learning to extract more powerful features from deep learning models. Transformer architectures~\citep{transformer} have recently been adapted for image processing tasks and in many cases are out performing convolutional networks. Applying modern vision transformers to image geo-localization problems has the potential to further improve the state-of-the-art in the field. Due to the amounts of data required by deep learning models, larger and higher quality datasets could further enhance results. Since applications for image geo-localization commonly \textcolor{red}{involve} user created imagery, continued research should be applied to building algorithms that are not reliant on specific illumination, vegetation, camera perspectives, or other consistent conditions. Instead, models must be trained to extract features invariant of the inconsistencies associated with crowd sourced datasets.

% What are future areas of research
% potential in self-supervised learning
% incorporating transformer architectures
% low resolution images and less data for poorer areas
% larger datasets, better take advantage of internet for crowd sourced data sets, broader set of assets to geolocalize, greater image quality
% better find features and method invariant of perspective to translate between domains
% less economically developed areas

\section*{Declarations}
The authors have no competing interests to declare that are relevant to the content of this article.


\bibliography{sn-bibliography}% common bib file


\end{document}